%% file: ContDPP.tex
\documentclass{article} % For LaTeX2e
\usepackage{nips13submit_e,times}
\usepackage{hyperref}
\usepackage{url}
\usepackage{geometry}
\usepackage{multirow}
\usepackage{amsmath, amsfonts, amssymb}
\usepackage{bm}
\usepackage{enumerate}
\usepackage{microtype}
\usepackage{hyperref}
\usepackage{verbatim}
\usepackage{color}

\usepackage[final]{graphicx} 
\usepackage{subfigure} 
\usepackage{epsfig,wrapfig}

\usepackage[numbers,sort&compress,comma]{natbib}
\usepackage{bibspacing}

\usepackage{algorithm}
\usepackage[noend]{algorithmic}
\usepackage{amsfonts}
\usepackage{amsthm}
\usepackage{amsmath}

\usepackage{authblk}

\newcommand{\erf}{\text{erf}}
\newcommand{\diag}{\text{diag}}

\newcommand{\argmax}{\operatornamewithlimits{argmax}}
\newcommand{\argmin}{\operatornamewithlimits{argmin}}
\newcommand{\tr}{\text{tr}}

\newcommand{\bx}{\bold{x}}
\newcommand{\by}{\bold{y}}

\newcommand{\bz}{\bold{z}}
\newcommand{\ba}{\bold{a}}
\newcommand{\bv}{\bold{v}}
\newcommand{\bw}{\bold{w}}
\newcommand{\bomega}{\boldsymbol{\omega}}
\newcommand{\Lapp}{\tilde{L}}
\newcommand{\LRFF}{\tilde{L}_{RFF}}
\newcommand{\LNys}{\tilde{L}_{Nys}}
\newcommand{\CRFF}{C^{RFF}}
\newcommand{\CNys}{C^{Nys}}
\newcommand{\FRFF}{F_{RFF}}
\newcommand{\FNys}{F_{Nys}}

\newcommand{\presec}{\vspace*{-10pt}} 
\newcommand{\postsec}{\vspace*{-6pt}} %TODO was 5
\newcommand{\pressec}{\vspace*{-10pt}} 
\newcommand{\postssec}{\vspace*{-5pt}} 
\newcommand{\prepar}{\vspace*{-8pt}}
\newcommand{\precap}{\vspace*{-0.1in}}
\newcommand{\postcap}{\vspace*{-0.275in}}

\title{Approximate Inference in Continuous\\Determinantal Point Processes}
\author[1]{\textbf{Raja Hafiz Affandi}}
\author[2]{\textbf{Emily B. Fox}}
\author[2]{\textbf{Ben Taskar}}
\affil[1]{University of Pennsylvania, \texttt{\small rajara@wharton.upenn.edu}}
\affil[2]{University of Washington, \texttt{\small \{ebfox@stat,taskar@cs\}.washington.edu}}

\newcommand{\out}[1]{}

\nipsfinalcopy % Uncomment for camera-ready version

\begin{document}

\maketitle

\begin{abstract}
Determinantal point processes (DPPs) are random point processes well-suited for modeling repulsion. %efficient modeling of repulsion. 
In machine learning, the focus of DPP-based models has been on diverse subset selection from a discrete and finite base set.  This discrete setting admits an efficient sampling algorithm based on the eigendecomposition of the defining kernel matrix. Recently, %in many areas of statistics and machine learning 
there has been growing interest in using DPPs defined on continuous spaces.  While the discrete-DPP sampler extends formally to the continuous case, computationally, the steps required are not tractable in general. %except in a few restricted cases.  
In this paper, we present two efficient DPP sampling schemes that apply to a wide range of kernel functions: one based on low rank approximations via Nystr{\"o}m and random Fourier feature techniques and another based on Gibbs sampling. %One approach is based on low rank approximations via Nystr{\"o}m and  random Fourier feature techniques while the other is based on Gibbs sampling. %In this paper, we present efficient approximate DPP sampling schemes based on Nystr{\"o}m and random Fourier feature approximations that apply to a wide range of kernel functions. 
We demonstrate the utility of continuous DPPs %in many applications where diversity is key, such as 
in repulsive mixture modeling and synthesizing human poses spanning activity spaces.
%DPPs on discrete sets has been shown to be useful in machine learning as a model of diverse subset selection. Furthermore, discrete DPPs offer efficient algorithms for sampling simply based on eigendecomposing the kernel matrix and choosing points based on probability distribution defined by the projection of these points on the subspace generated by selected eigenvectors. Recently, there has been a lot of interest in using DPPs defined on continuous spaces in many areas of statistics and machine learning. While in theory the concept of eigendecomposing the kernel and selecting points based on their projections extends to continuous space, in practice however, this cannot be easily done except in a few known cases. In this paper, we present an efficient approximate DPP sampling algorithms based on Nyst{\"o}m and random Fourier features methods. We will show that these sampling algorithms apply to a wide range of DPP kernel functions. Finally, we show that continuous DPPs are useful in modelling repulsive mixture models and generating diverse set of MoCap poses, among many other applications in statistics and machine learning.
\end{abstract}
\input{sections/introCR}

\input{sections/backgroundCR}

\input{sections/methodCR}
\input{sections/GibbsCR}

\input{sections/analysisCR}

\input{sections/applicationsCR}
\input{sections/conclusionCR}

\newpage
\bibliographystyle{plainnat}
\bibliography{ContDPP}

\appendix
\newgeometry{left=1in,right=1in,top=1.1in,bottom=0.8in}
\input{notation}
\title{Supplementary Material: \\Approximate Inference in Continuous Determinantal Processes}
\nipsfinalcopy % Uncomment for camera-ready version

%\counterwithin{figure}{section}
\author[1]{\textbf{Raja Hafiz Affandi}}
\author[2]{\textbf{Emily B. Fox}}
\author[2]{\textbf{Ben Taskar}}
\affil[1]{University of Pennsylvania, \texttt{\small rajara@wharton.upenn.edu}}
\affil[2]{University of Washington, \texttt{\small \{ebfox@stat,taskar@cs\}.washington.edu}}
 
 \maketitle
 
 \begin{abstract}
  We provide further details for the NIPS 2013 submission ``Approximate Inference in Continuous Determinantal Processes''. First, we elaborate upon the existing DPP samplers for the discrete and finite $\Omega$ case.  We then provide a list of standard cases when our (approximate) DPP sampling scheme can be performed. We derive the low-rank approximation and Gibbs sampling schemes for a few standard cases along with the details of empirical analysis of the low-rank approximations. For our mixture of Gaussian example application, we detail the model specification and Gibbs sampler and contrast with a standard (non-repulsive) mixture model.  Finally, we provide additional details on the settings used in our experiments and present some additional figures of results.
 \end{abstract}

\begin{appendix}
	
\section{DPP, $k$-DPP, and dual DPP sampling}

For $\Omega$ discrete and finite with cardinality $N$, we provide the algorithms for sampling from DPPs, $k$-DPPs, and DPPs via the dual representation in Algorithms~\ref{alg:dppsampling},~\ref{alg:kdppsampling},~\ref{alg:dualsampling}.  In the $k$-DPP sampler, $e_i$ denotes the $i$th elementary symmetric polynomial.  For $\Omega$ continuous, we provide the continuous $k$-DPP dual sampler in Algorithm~\ref{alg:kdualCont}.  Note that the only difference relative to the DPP dual sampler is in the for loop of Phase 1.  The revision exactly parallels the story for the discrete $\Omega$ case.

\begin{algorithm}[h!]
   \caption{DPP-Sample(L)}
   \label{alg:dppsampling}
	\begin{minipage}[l]{3in}
		\begin{algorithmic}
   \STATE {\bfseries Input:} kernel matrix $L$ of rank $D$\\
   \STATE{\bfseries PHASE 1}
   \STATE $\{(\v_n,\lambda_n)\}_{n=1}^D \leftarrow $
   eigendecomposition of $L$\\
   \STATE $J\leftarrow \emptyset$\\
   \FOR{$n=1,\ldots,D$} 
   \STATE  $J\leftarrow J\cup\{n\}$ with prob. $\frac{\lambda_n}{\lambda_n+1}$
   \ENDFOR
   \STATE $V\leftarrow \{\v_n\}_{n\in J}$
\end{algorithmic}
\end{minipage}
\begin{minipage}[r]{3.25in}
		\begin{algorithmic}
\STATE{\bfseries PHASE 2}
   \STATE $Y\leftarrow \emptyset$\\
   \WHILE{$|V|>0$}
   \STATE  Select $i$ from $\Omega$ with Pr($i)=\frac{1}{|V|}\sum_{\v\in V}(\v^{\top}e_i)^2$
   \STATE  $Y\leftarrow Y\cup\{i\}$ 
   \STATE  $V\leftarrow V_{\bot e_i}$, an orthonormal basis for the subspace of V orthogonal to $e_i$
   \ENDWHILE
   \STATE  {\bfseries Output:} $Y$
\end{algorithmic}
\end{minipage}

\end{algorithm}

\begin{algorithm}[h!]
  \caption{$k$-DPP-Sample(L)}
  \label{alg:kdppsampling}
	\begin{minipage}[l]{3in}
		\begin{algorithmic}
  \STATE {\bfseries Input:} kernel matrix $L$ of rank $D$, size $k$\\
  \STATE{\bfseries PHASE 1}
  \STATE $\{(v_n,\lambda_n)\}_{n=1}^D \leftarrow $
  eigendecomposition of $L$\\
   \STATE $J\leftarrow \emptyset$\\
  \FOR{$n=D,\ldots,1$} 
  \IF{$u\sim U[0,1] < \lambda_n\frac{e^{n-1}_{k-1}}{e^{n}_k}$} 
  \STATE $J\leftarrow J\cup\{n\}$\\
  \STATE $k\leftarrow k-1$\\
  \IF{$k=0$}
  \STATE $\mathbf{break}$ \\
  \ENDIF 
  \ENDIF
  \ENDFOR
   \STATE $V\leftarrow \{\v_n\}_{n\in J}$
\end{algorithmic}
\end{minipage}
\begin{minipage}[r]{3.25in}
	\begin{algorithmic}
\STATE{\bfseries PHASE 2}
  \COMMENT{same as Algorithm~\ref{alg:dppsampling}}\\
\end{algorithmic}
\end{minipage}

\end{algorithm}

\begin{algorithm}[h!]
	\begin{minipage}[l]{3in}
		\begin{algorithmic}
  \STATE {\bfseries Input:} $B \in \mathbb{C}^{D\times N}$ such that $L = B^* B$.
  \STATE{\bfseries PHASE 1}
\STATE $C \leftarrow BB^*$
  \STATE  $\{(\cv_n,\lambda_n)\}_{n=1}^D \leftarrow $
   eigendecompistion of $C$  \\
  \STATE $J \leftarrow \emptyset$
  \FOR{$n = 1,\dots,D$}
  \STATE $J \leftarrow J \cup \{n\}$ with prob. $\frac{\lambda_n}{\lambda_n+1}$
  \ENDFOR
  \STATE $\cV \leftarrow \left\{\frac{\cv_n}{\sqrt{\cv^* C\cv}}\right\}_{n\in J}$
\end{algorithmic}
\end{minipage}
\begin{minipage}[r]{3.25in}
	\begin{algorithmic}
\STATE{\bfseries PHASE 2}
  \STATE $Y \leftarrow \emptyset$
  \WHILE{$|\cV|>0$}
  \STATE Select $i$ from $\Omega$ with 
  $\Pr(i) = \frac{1}{|\cV|}\sum_{\cv\in \cV} (\cv^* B_i)^2$
  \STATE $Y \leftarrow Y \cup \{i\}$
  \STATE Let $\cv_0$ be a vector in $\cV$ with $B_i^* \cv_0 \neq 0$
  \STATE Update $\cV \leftarrow \left\{ \cv 
  - \frac{\cv^* B_i}{\cv_0^* B_i}\cv_0
  \ \vert\ \cv \in\cV - \{\cv_0\} \right\}$
  \STATE Orthonormalize $\cV$ w.r.t. %the dot product 
  $\langle \cv_1,\cv_2 \rangle = \cv_1^* C \cv_2$
  \ENDWHILE
  \STATE {\bfseries Output:} $Y$
\end{algorithmic}
\end{minipage}
\caption{Dual-DPP-Sample(B)}
\label{alg:dualsampling}
\end{algorithm}

\begin{algorithm}[h!]
 \caption{Dual sampler for a low-rank continuous $k$- DPP}
  \label{alg:kdualCont}
	\begin{minipage}[l]{3in}
		\begin{algorithmic}
 \STATE {\bfseries Input:} $\Lapp(\bx,\by)=B(\bx)^*B(\by)$,\\
  \STATE \hspace{0.4in} a rank-$D$ DPP kernel \\  
\STATE{\bfseries PHASE 1}
  \STATE Compute $C=\int_{\Omega}B(\bx)B(\bx)^* d\bx$  
  \STATE $\{(\bv_n,\lambda_n)\}_{n=1}^D \leftarrow $ eigendecomposition of $C$
   %eigendecomp. of  $CCompute eigendecomposition $C = \sum_{k=1}^D\lambda_k \bv_k \bv_k^*$\\ %$\{(\bv_k,\lambda_k)\}_{k=1}^D$ of $C$\\
  \FOR{$n=D,\ldots,1$} 
  \IF{$u\sim U[0,1] < \lambda_n\frac{e^{n-1}_{k-1}}{e^{n}_k}$} 
  \STATE $J\leftarrow J\cup\{n\}$\\
  \STATE $k\leftarrow k-1$\\
  \IF{$k=0$}
  \STATE $\mathbf{break}$ \\
  \ENDIF 
  \ENDIF
  \ENDFOR 
  \STATE $V\leftarrow\{\frac{v_k}{\sqrt{v_k^{*}C v_k}}\}_{k\in J}$
	\end{algorithmic}
\end{minipage}
\begin{minipage}[r]{3.25in}
		\begin{algorithmic}
\STATE{\bfseries PHASE 2}
  \STATE $X\leftarrow\emptyset$\\
  \WHILE{$|V|>0$}
  \STATE Sample $\hat{\bx}$ from density $f(\bx)=\frac{1}{|V|}\sum_{\bv\in V}|\bv^*B(\bx)|^2$ \\
  \STATE $X\leftarrow X\cup\{\hat{\bx}\}$\\
  \STATE Let $\bv_0$ be a vector in $V$ such that $\bv_0^*B(\hat{\bx})\neq 0$\\
  \STATE Update $V\leftarrow\{\bv-\frac{\bv^*B(\hat{\bx})}{\bv_0^*B(\hat{\bx})}\bv_0~|~v\in V-\{v_0\} \}$\\
  \STATE Orthonormalize $V$ w.r.t. $\langle \bv_1,\bv_2\rangle=\bv_1^{*}C\bv_2$\\
  \ENDWHILE
  \STATE  {\bfseries Output:} $X$
\end{algorithmic}
\end{minipage}
\end{algorithm}

\newpage
\section{Derivation of the Gibbs sampling scheme}
For a $k$-DPP, the probability of choosing a specific $k$ point configuration is given by
\begin{align}
p(\{\bx_j\}_{j=1}^k)\propto\det(L_{\{\bx_j\}_{j=1}^k}).
\end{align}
Denoting $J^{\backslash k}=\{\bx_j\}_{j\neq k}$ and $M^{\backslash k}=L_{J^{\backslash k}}^{-1}$, the Schur's determinantal identity formula yields
\begin{align}
\det(L_{\{\bx_j\}_{j=1}^k})=\det(L_{J^{\backslash k}})\left(L(\bx_k,\bx_k)-\sum_{i,j\neq k}M^{\backslash k}_{ij}L(\bx_i,\bx_k)L(\bx_j,\bx_k)\right).
\label{eq:Schur}
\end{align}
Conditioning on the inclusion of the other $k-1$ points, and suppressing constants not dependent on $\bx_k$ we can now write the conditional distribution as
%\begin{align}
%p(\bx_k|\{\bx_j\}_{j\neq k})= \frac{p(\{\bx_j\}_{j=1}^k)}{p(\{\bx_j\}_{j\neq k})}\propto p(\{\bx_j\}_{j=1}^k)\propto\det(L_{\{\bx_j\}_{j=1}^k}).
%\end{align}
%Using Eq.~\ref{eq:Schur}, we now conclude that
%\begin{align}
%p(\bx_k|\{\bx_j\}_{j\neq k})\propto\det(L_{J^{\backslash k}})\det(L(\bx_k,\bx_k)-\sum_{i,j\neq k}M^{\backslash k}_{ij}L(\bx_i,\bx_k)L(\bx_j,\bx_k)).
%\end{align}
%Since, conditioned on the other $k-1$ points, $\det(L_{J^{\backslash k}})$ is now a constant, we get
\begin{align}
p(\bx_k|\{\bx_j\}_{j\neq k})\propto L(\bx_k,\bx_k)-\sum_{i,j\neq k}M^{\backslash k}_{ij}L(\bx_i,\bx_k)L(\bx_j,\bx_k),
\end{align}
%which can be interpreted as a 1-DPP with a modified kernel $L(\bx_k,\bx_k)-\sum_{i,j\neq k}M^{\backslash k}_{ij}L(\bx_i,\bx_k)L(\bx_j,\bx_k)$. Note that conditioned on the other $k-1$ points, $L(\bx_k,\bx_k)-\sum_{i,j\neq k}M^{\backslash k}_{ij}L(\bx_i,\bx_k)L(\bx_j,\bx_k)$ is just a function of $\bx_k$, implying
%\begin{align}
%p(\bx_k|\{\bx_j\}_{j\neq k})\propto L(\bx_k,\bx_k)-\sum_{i,j\neq k}M^{\backslash k}_{ij}L(\bx_i,\bx_k)L(\bx_j,\bx_k).
%\end{align}
Normalizing and integrating this density yields a full conditional CDF given by

\begin{align}
F(\hat{\bx}_l|{\{\bx_j\}}_{j\neq k})= \frac{\int_{-\infty}^{\hat{\bx}_l}L(\bx_l,\bx_l)-\sum_{i,j\neq k}M^{\backslash k}_{ij}L(\bx_i,\bx_l)L(\bx_j,\bx_l)1_{\{\bx_l\in \Omega\}}d\bx_l}{\int_\Omega L(\bx,\bx)-\sum_{i,j\neq k}M^{\backslash k}_{ij}L(\bx_i,\bx)L(\bx_j,\bx)d\bx}.
\label{eq:fullCondCDF}
\end{align}
	
\section{Overview of analytically tractable kernel types  under RFF or Nystr{\"om}}

Sampling from a DPP with kernel $L$ using Algorithm~1 of the main paper %\ebf{maybe we can link to main paper} 
requires that (i) we can compute a low-rank decomposition $\Lapp$ of $L$ and (ii) the terms $C$ and $f(\bx)$ are computable.  In the main paper, we consider a decomposition of $L(\bx,\by) = q(\bx)k(\bx,\by)q(\by)$ where $q(\bx)$ is a quality function and $k(\bx,\by)$ a similarity kernel.  We then use either random Fourier features (RFF) or the Nystr{\"om} method to approximate $L$ with $\Lapp$.  In general, we can consider RFF approximations whenever the spectral density of $q(\bx)$ and characteristic function of $k(\bx,\by)$ are known.  For Nystr{\"om}, the statement is not quite as clear.  Instead, we provide a list of standard choices and their associated feasibilities for DPP sampling in Table~\ref{table:standard}.  The list is by no means exhaustive, but is simply to provide some insight. We also elaborate upon some standard kernels in the following sections.%the examples provided in the main paper in the following sections.

%\ebf{INCLUDE MORE EXAMPLES BELOW}
%
\begin{table}[h!]
	\caption{Examination of the feasibility of DPP sampling using Nystr{\"om} and RFF approximations for a few standard examples of quality functions $q$ and similarity kernels $k$.}
	\label{table:standard}
	\begin{center}
\begin{tabular}{|c|c|c l|}
	\hline
	$q(x)$ & $k(x,y)$ & Method &\\
	\hline
	\multirow{2}{*}{Gaussian, Laplacian} & \multirow{2}{*}{Gaussian, Laplacian} & Nystr{\"om} &\checkmark\\
	& & RFF &\checkmark \\& & Gibbs &\checkmark \\
	\hline
	\multirow{2}{*}{Gaussian, Laplacian} & \multirow{2}{*}{Cauchy} & Nystr{\"om} &?\\
	& & RFF & \checkmark \\ & & Gibbs &? \\
	\hline
          \multirow{2}{*}{Cauchy} & \multirow{2}{*}{Gaussian, Laplacian} & Nystr{\"om} &?\\
	& & RFF &\checkmark \\& & Gibbs &? \\
	\hline
	\multirow{2}{*}{Cauchy} & \multirow{2}{*}{Cauchy} & Nystr{\"om} &?\\
	& & RFF & \checkmark \\& & Gibbs & ? \\
	\hline
	\multirow{2}{*}{Gaussian, Laplacian} & \multirow{2}{*}{Linear, Polynomial} & Nystr{\"om} & \checkmark\\
	& & RFF & X\\ & & Gibbs & \checkmark \\
	\hline
\end{tabular}
\end{center}
\end{table}

\subsubsection*{Example: Sampling from RFF-approximated DPP with Gaussian quality}

Assuming $q(\bx)=\exp\left\{-\frac{1}{2}(\bx-\ba)^{\top}\Gamma^{-1}(\bx-\ba)\right\}$ and $k(\bx,\by)= k(\bx-\by)$ is given by a translation-invariant kernel with known characteristic function. We start by sampling $\bomega_1,\ldots,\bomega_D\sim\mathcal{F}(k(\bx-\by))$. Note, for example, that the Fourier transform of a Gaussian kernel is a Gaussian while that of the Laplacian is Cauchy and vice versa. The approximated kernel is given by
%\begin{equation}
%\LRFF=\frac{1}{D}\sum_{j=1}^D\exp\{i{\bomega_j}^{\top}(\bx-\by)\}
%\end{equation}
%The elements of dual matrix $\CRFF$ is then given by
%\begin{equation}
%\CRFF_{jk}=\frac{1}{D}\int_{\mathcal{A}^d}\exp\{i(\bomega_j-\bomega_k)^{\top}\bx\}d\bx
%\end{equation}
%Without loss of generality, let $\mathcal{A}^d=[-A,A]^d$, a hypercube in the standard basis.
%Then
%\begin{equation}
%\CRFF_{jk}=\frac{1}{D}\prod_{l=1}^d\Bigg[\frac{2\sin(A(w_{jl}-w_{kl}))}{w_{kl}-w_{jl}}\Bigg]
%\end{equation}
%where $\bomega_{jl}$ is the $l^th$ coordinate of vector $\bomega_j$.
%
%Finally, the CDF is given by
%\begin{equation}
%F(x)=\frac{1}{D|V|}\sum_{\bv\in V}\sum_{j=1}^D\sum_{k=1}^D\bv^{(j)}\bv^{(k)*}\prod_{l=1}^d [i(\frac{\exp\{i\bx_l(w_{jl}-w_{kl}\}-\exp\{iA(\bomega_{jl}-\bomega_{kl})\}}{\bomega_{jl}-\bomega_{kl}})]
%\end{equation}
%
%In case (2), the kernel is given by}
%
\begin{equation}
\LRFF=q(\bx)\Bigg[\frac{1}{D}\sum_{j=1}^D\exp{i\bomega_j}^{\top}(\bx-\by)\Bigg]q(\by)~~\text{where}~~q(\bx)=\exp\left\{-\frac{1}{2}(\bx-\ba)^{\top}\Gamma^{-1}(\bx-\ba)\right\}.
\end{equation}
%where
%\begin{equation}

%\end{equation}
The elements of the dual matrix $\CRFF$ are then given by
\begin{equation}
\CRFF_{jk}=\frac{1}{D}\int_{\mathbb{R}^d}\exp\{-(\bx-\ba)^{\top}\Gamma^{-1}(\bx-\ba)+i(\bomega_j-\bomega_k)^{\top}\bx\}d\bx.
\end{equation}

Letting $R\Delta R^{\top}$ be the spectral decompostition of $\Gamma^{-1}$ with $\Delta=\diag(\frac{1}{\delta_1^2},\ldots,\frac{1}{\delta_D^2})$, $\tilde{\bomega}_j=R^{\top}\bomega_j, \tilde{\ba}=R^{\top}\ba$ and $\by=R^{\top}\bx$, one can straightforwardly derive:
\begin{eqnarray}
\CRFF_{jk}&=&\frac{1}{D}\prod_{l=1}^d\Bigg[\sqrt{\pi\delta_l^2}\exp\left\{-\frac{\delta_l^2(\tilde{\bomega}_{jl}-\tilde{\bomega}_{jk})^2}{4}\right\}+i\tilde{a}_l(\tilde{\bomega}_{jl}-\tilde{\bomega}_{jk})\Bigg].
\end{eqnarray}  
Likewise,
\begin{equation}
\FRFF(\by)=\frac{1}{D|V|}\sum_{\bv\in V}\sum_{j=1}^D\sum_{k=1}^D\bv^{(j)}\bv^{(k)*}\prod_{l=1}^dg(\tilde{\bomega}_{jl},\tilde{\bomega}_{kl},\tilde{a}_l,\delta_l,y_l),
\end{equation}
where
\begin{equation*}
g(\tilde{\bomega}_{jl},\tilde{\bomega}_{kl},\tilde{\ba}_l,\delta_l,y_l)=\frac{1}{2}\sqrt{\pi\delta_l^2}\exp\left\{-\frac{\delta_l^2(\tilde{\bomega}_{jl}-\tilde{\bomega}_{kl})^2}{4}\right\}+i\tilde{a}_l\left(\tilde{\bomega}_{jl}-\tilde{\bomega}_{kl})(1-\erf\left(\frac{i\sqrt{\delta_l^2}(\tilde{\bomega}_{jl}-\tilde{\bomega}_{kl})}{2}-\frac{y_l-\tilde{a}_l}{2\sqrt{\delta_l^2}}\right)\right).
\end{equation*}
Once samples $\by$ are obtained, we transform back into our original coordinate system by letting $\bx=R\by$.  

\subsubsection*{Example: Sampling from Nystr{\"o}m-approximated DPP with Gaussian quality and similarity}
Assuming $q(\bx)=\exp\left\{-\frac{1}{2}(\bx-\ba)^{\top}\Gamma^{-1}(\bx-\ba)\right\}$ and $k(\bx,\by) = \exp\left\{-\frac{1}{2}(\bx-\by)^{\top}\Sigma^{-1}(\bx-\by)\right\}$, the approximated kernel is given by
\begin{equation}
\LNys(\bx,\by)=\sum_{j=1}^D\sum_{k=1}^DW_{jk}^2q(\bx)q(\bz_j)\exp\left\{-\frac{1}{2}(\bx-\bz_j)^{\top}\Sigma^{-1}(\bx-\bz_j)-\frac{1}{2}(\by-\bz_k)^{\top}\Sigma^{-1}(\by-\bz_k)\right\}q(\bz_k)q(\by).
\end{equation}
%where
%\begin{equation}
%q(\bx)=\exp\left\{-\frac{1}{2}(\bx-\ba)^{\top}\Gamma^{-1}(\bx-\ba)\right\}.
%\end{equation}

Let $\Sigma^{-1}=Q\Lambda Q^{\top}$ with $\Lambda=\diag(\frac{1}{\sigma_1^2},\ldots,\frac{1}{\sigma_D^2})$, $\Gamma^{-1}=R\Delta R^{\top}$ with $\Delta=\diag(\frac{1}{\delta_1^2},\ldots,\frac{1}{\delta_D^2})$ and
$(\Sigma^{-1}+\Gamma^{-1})=T\Theta T^{\top}$ with $\Theta=\diag(\frac{1}{\theta_1^2},\ldots,\frac{1}{\theta_D^2})$. Furthermore, let $\tilde{\bz}_j=T^{\top}(\Gamma^{-1}+\Sigma^{-1})\Sigma^{-1}\bz_j$, $\tilde{\ba}=T^{\top}(\Gamma^{-1}+\Sigma^{-1})\Gamma^{-1}\ba$ and $\by=T^{\top}\bx$. Then, the elements of the dual matrix $\CNys$ are then given by
\begin{equation}
\CNys_{jk}=\sum_{m-1}^D\sum_{n=1}^DW_{jn}W_{mk}A_{mn}\prod_{l=1}^d\sqrt{\pi\theta_l^2}.
\end{equation}
where
\begin{eqnarray*}
A_{mn}&=&\exp\bigg\{-\frac{1}{2}(\bz_n-\ba)^{\top}\Gamma^{-1}(\bz_n-\ba)-\frac{1}{2}(\bz_m-\ba)^{\top}\Gamma^{-1}(\bz_m-\ba)-\frac{1}{2}\bz_m^{\top}\Sigma^{-1}\bz_m-\frac{1}{2}\bz_n^{\top}\Sigma^{-1}\bz_n\\
&~&+(\Gamma^{-1}\ba+\Sigma^{-1}\frac{(\bz_m+\bz_n)}{2})^{\top}(\Sigma^{-1}+\Gamma^{-1})^{-1}(\Gamma^{-1}\ba+\Sigma^{-1}\frac{(\bz_m+\bz_n)}{2})-\ba^{\top}\Gamma^{-1}\ba\bigg\}.
\end{eqnarray*}
Finally, the CDF of $f(\by)$ is given by
\begin{equation}
\FNys(\by)=\frac{1}{|V|}\sum_{\bv\in V}\sum_{j,k=1}^D\bw_j(\bv)\bw_k(\bv) A_{jk}\prod_{l=1}^d\frac{\sqrt{\pi\theta_l^2}}{2}\Bigg[1-\erf\left(\frac{2\tilde{a}_l+\tilde{z}_{jl}+\tilde{z}_{kl}-2y_l}{2\sqrt{\theta_l^2}}\right)\Bigg].
\end{equation}
Once samples $\by$ are obtained, we transform back to our original coordinate system by letting $\bx=T\by$.  

\subsubsection*{Example: Sampling from Nystr{\"o}m-approximated DPP with Gaussian quality and polynomial similarity}

For simplicity of exposition, we consider a linear similarity kernel and $d=1$, although the result can straightforwardly be extended to higher order polynomials and dimensions $d$. Assuming $q(x)=\exp{\{-\frac{x^2}{2\rho^2}\}}$ and $k(x,y)=xy$, the approximated kernel is given by
\begin{equation}
\LNys(x,y)=\sum_{j=1}^D\sum_{k=1}^DW_{jk}^2\exp\left\{-\frac{(x^2+z_j^2+z_k^2+y^2)}{2\rho^2}\right\}(xz_j)(yz_k).
\end{equation}

The elements of the dual matrix $C^{Nys}$ are then given by
\begin{equation}
\CNys_{jk}=\sum_{m-1}^D\sum_{n=1}^DW_{jn}W_{mk}\frac{z_mz_n}{2}\exp\{-\frac{z_m^2+z_n^2}{2\rho^2}\}\sqrt{\pi}\rho^3.
\end{equation}

The CDF is given by
\begin{equation}
\FNys(y)=\frac{1}{|V|}\sum_{\bv\in V}\sum_{j,k=1}^D\bw_j(\bv)\bw_k(\bv)\frac{z_jz_k}{2}\exp\{-\frac{z_j^2+z_k^2}{2\rho^2}\}\Bigg[\frac{\sqrt{\pi}\rho^3}{4}\left[\erf\left(\frac{y}{\sqrt{r}}\right)+1\right]-2ye^{-\frac{y^2}{\rho^2}}\Bigg].
\end{equation}

\subsubsection*{Example: Gibbs sampling with Gaussian quality and similarity}

For generic kernels $L(\bx,\by)=q(\bx)k(\bx,\by)q(\by)$, we recall that the CDF of $\bx_k$ given $\{\bx_j\}_{j\neq k}$ for a $k$-DPP is given by 
\begin{align}
F(\hat{\bx}_k|{\{\bx_j\}}_{j\neq k})= \frac{\int_{-\infty}^{\hat{\bx}_k}q(\bx_k)^2(1-\sum_{i,j\neq k}M_{ij}q(\bx_i)q(\bx_j)k(\bx_k,\bx_i)k(\bx_j,\bx_k))1_{\{\bx_k\in \Omega\}}d\bx_k}{\int_\Omega q(\bx)^2(1-\sum_{i,j\neq k}M_{ij}q(\bx_i)q(\bx_j)k(\bx,\bx_i)k(\bx_j,\bx))d\bx}.
\end{align}

Assuming $q(\bx)=\exp\left\{-\frac{1}{2}(\bx-\ba)^{\top}\Gamma^{-1}(\bx-\ba)\right\}$ and $k(\bx,\by) = \exp\left\{-\frac{1}{2}(\bx-\by)^{\top}\Sigma^{-1}(\bx-\by)\right\}$, the integrals above can be solved to yield

\begin{align}
F(\hat{\bx}_k|{\{\bx_j\}}_{j\neq k})=\frac{\prod_{l=1}^d\Bigg[\frac{\sqrt{\pi\delta_l^2}}{2}\left[1-\erf\left(\frac{2\tilde{a}_l-2x_{kl}}{2\sqrt{\delta_l^2}}\right)\right]-\sum_{i,j\neq k}M_{ij}A_{ij}\frac{\sqrt{\pi\theta_l^2}}{2}\left[1-\erf\left(\frac{2\tilde{a}_l+\tilde{z}_{il}+\tilde{z}_{jl}-2x_{kl}}{2\sqrt{\theta_l^2}}\right)\right]\Bigg]}{\prod_{l=1}^d\left[\sqrt{\pi\delta_l^2}-\sum_{i,j\neq k}W_{ij}A_{ij}\sqrt{\pi\theta_l^2}\right]}.
\end{align}
where $\tilde{\ba},\tilde{\bz},\delta_l,A_{ij}$ and $\theta_l$ are as given in the previous examples.

\section{Details of the empirical analysis}

To evaluate the performance of the RFF and Nystr{\"o}m approximations, we compute the total variational distance
\begin{equation}
\|\mathcal{P}_L-\mathcal{P}_{\tilde{L}}\|_1=\frac{1}{2}\sum_{X}|\mathcal{P}_L(X)-\mathcal{P}_{\tilde{L}}(X)|~,
\end{equation}

where $\mathcal{P}_L(X)$ denotes the probability of set $X$ under a DPP with kernel $L$, as given by Eq.~(1). One can show that the normalized density is $\mathcal{P}_L(X) =\frac{\det(L_X)}{\prod_{n=1}^\infty(1+\lambda_n(L))}$, which requires the eigenvalues of the kernel $L$. Thus, we restrict our analysis to the case where the quality function and similarity kernel are Gaussians with isotropic covariances $\Gamma=\diag(\rho^2,\ldots,\rho^2)$ and $\Sigma=\diag(\sigma^2,\ldots,\sigma^2)$, respectively, since the eigenvalues of the kernel is easily computable in this setting ~\cite{fasshauer2012stable}. In this case, letting $n=(n_1,\dots,n_d)$ with $n_j \in \mathbb{Z}_+$, the eigenvalues (indexed by multi-index $n$) are given by:%\ben{need to define $n_j$ below (is n a multindex?)}:
%\begin{equation}
%\phi_n(x)=\left(\frac{\rho^2}{\pi}\right)^{\frac{d}{4}}\exp\left\{-\frac{\|x\|^2}{2\rho^2}\right\}\prod_{j=1}^d\sqrt{\frac{\beta}{2^{n-1}\Gamma(n)}}\exp\bigg(-\frac{(\beta^2-1)x_j^2}{2\rho^2}\bigg)H_{n_j-1}\left(\frac{\beta x_j}{\sqrt{\rho^2}}\right)
%\end{equation}
%and 
\begin{equation}
%\lambda_n=\prod_{j=1}^d \sqrt{\frac{2\pi}{\rho^2(\beta^2+1)+4(\frac{\rho^4}{\sigma^2})}}\bigg(\frac{4}{(\frac{\sigma^2}{\rho^2})(\beta^2+1)+4}\bigg)^{n_j-1}~,
\lambda_n=\prod_{j=1}^d \sqrt{\frac{\pi\rho^2}{\frac{c_1}{2}+c_2}}\bigg(\frac{1}{\frac{c_1}{c_2}+1}\bigg)^{n_j-1} \hspace{0.25in} c_1 = (\beta^2+1) \hspace{0.25in} c_2 = \frac{\rho^2}{\sigma^2}~.
\end{equation}
where $\beta=(1+\frac{2\rho^2}{\sigma^2})^{\frac{1}{4}}$. Since the eigenvalues are known in closed form, we can estimate the total variation distance by sampling sets $X$ from the approximated DPP and calculating the absolute difference between $\mathcal{P}_L(X)$ and $\mathcal{P}_{\tilde{L}}(X)$. %and $H_n$ denotes the $n$th Hermite polyomial.
%Our analyses also focus on sampling from $k$-DPPs for which the size of the set $X$ is always $k$.
\section{Empirical analysis of Gibbs sampling}
To assess the mixing rate of the Gibbs sampling scheme, we run the Gibbs sampler to sample points from a 1-dimensional $15$-DPP with uniform quality and Gaussian similarity kernels in the space $\Omega=[-\frac{1}{2},\frac{1}{2}]$.  We perform this sampling under two values of repulsion parameter, $\sigma^2=0.01$ (high repulsion) and $\sigma^2=0.001$ (low repulsion). We run 100 Gibbs chains, each of length 3000, discard the first 1500 samples as burn-in and thin every 15 iterations which we call cycles. Each cycle represents a full resampling of the set, having cycled through the past 15 points.  We compare the results to i.i.d. sampling of Nystr{\"o}m-approximated DPP as a baseline. 

\begin{figure}[htb!]
	\begin{center}
		\begin{tabular}{cc}
\includegraphics[scale=0.5]{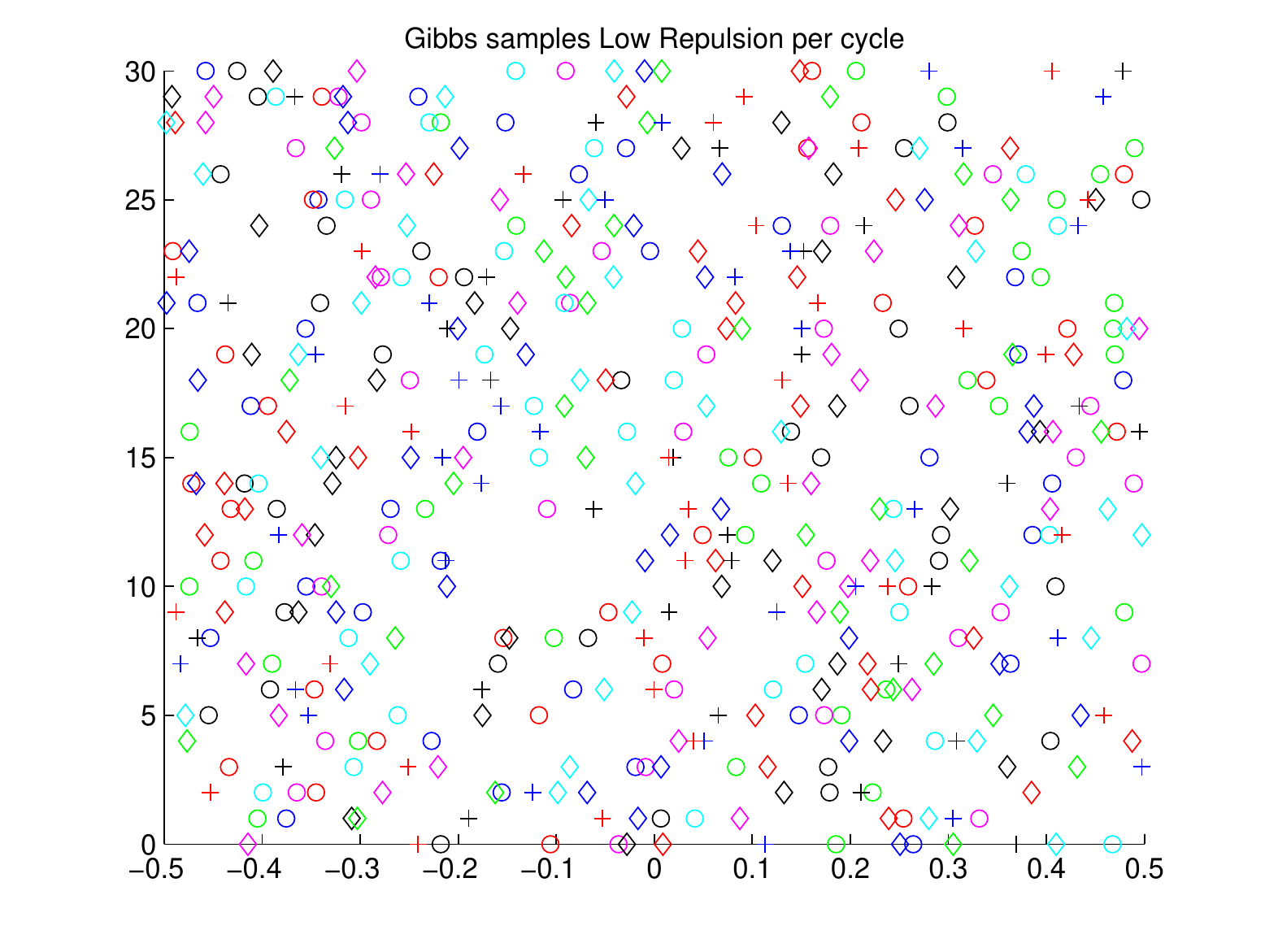} & 
\includegraphics[scale=0.5]{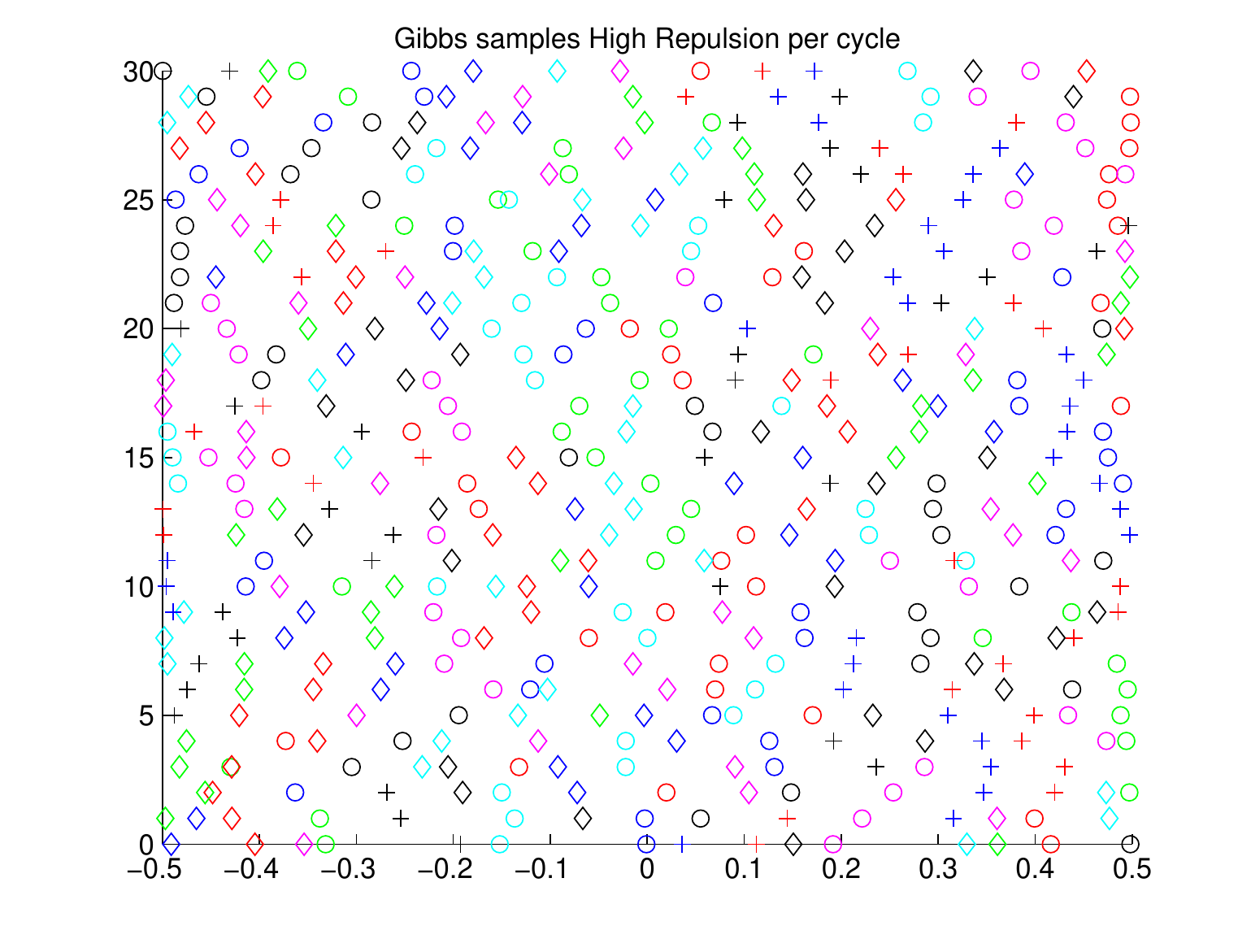}\\
(a) & (b)\\
\includegraphics[scale=0.5]{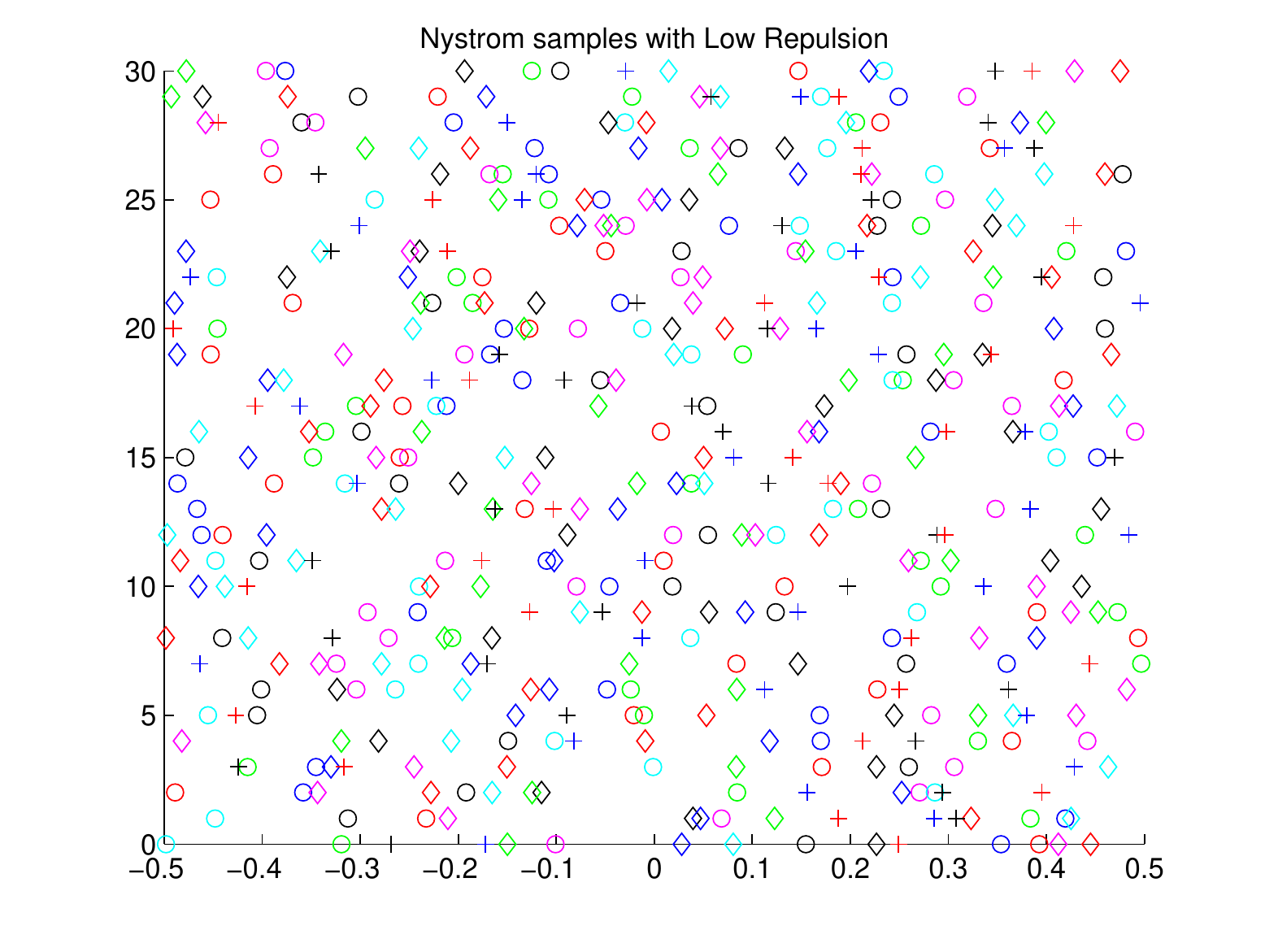} & 
\includegraphics[scale=0.5]{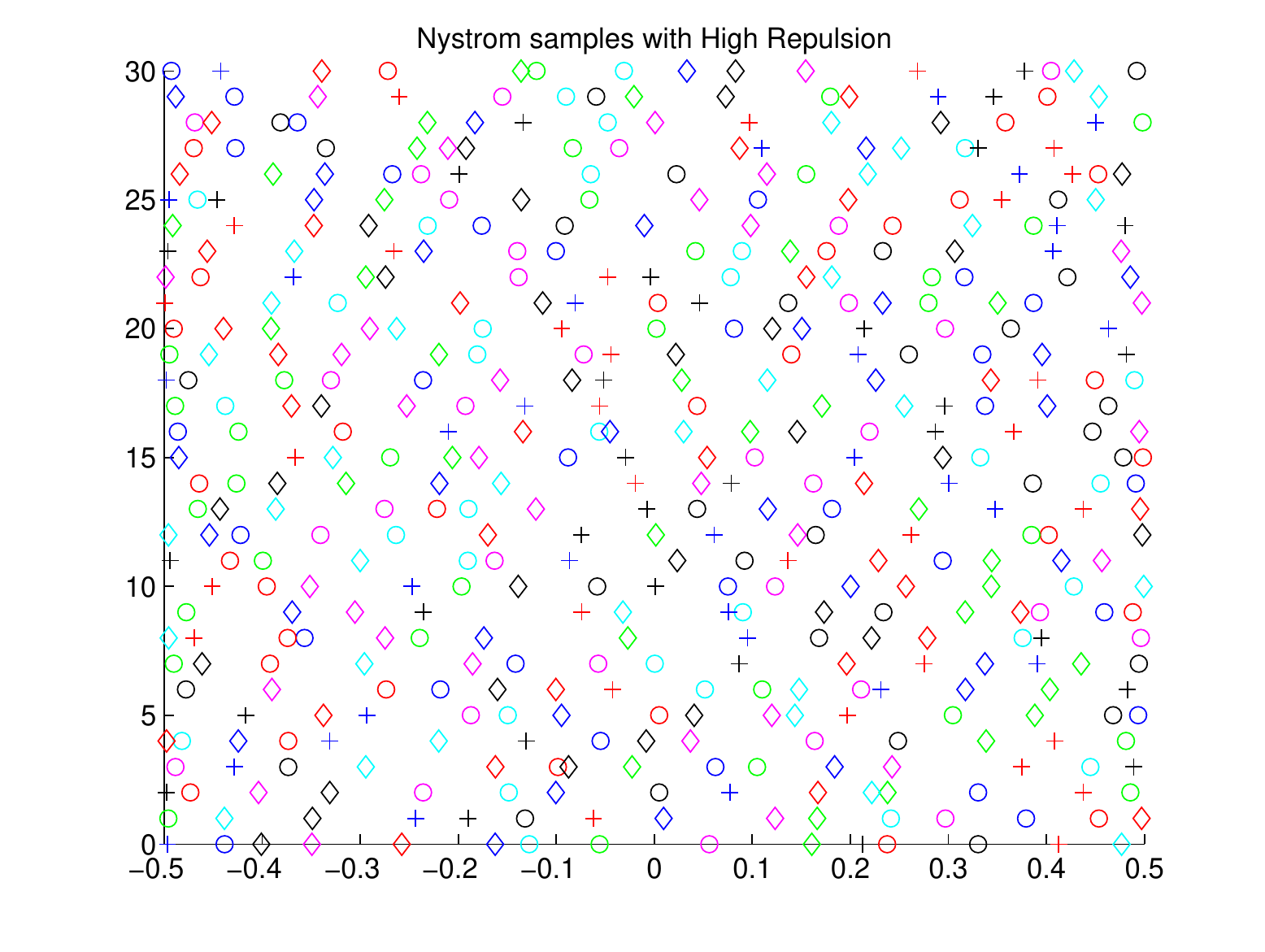}\\
(c) & (d)\\
\end{tabular}
\end{center}
\precap
\caption{Visualization plots of location of 1-dimension DPP samples: (a)-(b) are samples from Gibbs scheme in low repulsion and high repulsion setting, respectively, (c)-(d) are i.i.d. samples from the Nystr{\"o}m-approximated DPP. }
\label{fig:scatterGibbs}
\end{figure}

Figure \ref{fig:scatterGibbs} (a)-(b) shows a visualization of the 15 points of the 15-DPPs. Figure \ref{fig:scatterGibbs} (c)-(d) shows the plots of the Nystr{\"o}m-approximated DPP samples. As an ordered set, we see qualitatively that the locations of the points are highly correlated from cycle to cycle in the high repulsion Gibbs samples while less correlation is observed in the low-repulsion counterpart. In the Nystr{\"o}m approximated case, there are no correlations as the samples are generated i.i.d.. 

Quantitatively, we use two measures as a proxy to the mixing rate: the average movement of point from cycle to cycle and the effective sample size. The average movement, $m$, is simply defined as the average difference in distance between points from one cycle to another averaged over the cycles:
\begin{align}
m=\frac{1}{T-1}\frac{1}{k}\sum_{t=1}^{T-1}\sum_{i=1}^k (x_i^{t+1}-x_i^t)^2,
\end{align}
where $T$ is the length of the chain after burn-in and thinning, $k$ is the number of points and $x_i^t$ is the coordinate of point $x_i$ at cycle $t$. In our experiment, $T$ and $k$ are 100 and 15, respectively. When the Gibbs chain is mixing well, we expect the average movement to be high as this signals that the points are less correlated across cycles.

The effective sample size is a standard measure in assessing the mixing of a Gibbs chain. To compute this, we first compute the lag-$s$ autocorrelation function of each point in the sampled sets. We then average the autocorrelation function at lag-$s$ across the $k$ points and denote this quantity $\bar{\rho}_s$. The effective sample size is then given by: $\alpha T$, where
\begin{align}
\alpha= \frac{1}{1+2\sum_{s=1}^{2\delta+1}\bar{\rho}_s},
\end{align}
where $\delta$ is the smallest positive integer satisfying $\bar{\rho}_{2\delta}+\bar{\rho}_{2\delta+1}>0$. In the case of i.i.d. samples, we expect $\alpha$ to be close to 1 while in cases where the mixing is bad, $\alpha$ will be much lower. 

Table \ref{table:Gibbsmix} shows the average values of $m$ and $\alpha$ for our Gibbs samples with i.i.d. Nystr{\"o}m-approximated DPP samples serving as a benchmark. We see that in the low repulsion setting, the Gibbs chain mixes well with values close to the benchmarks while for the Gibbs sampler in the high repulsion setting, the values of $m$ and $\alpha$ are much lower, indicating slow mixing.  
  
\begin{table}[h!]
	\begin{center}
\begin{tabular}{|r|c|c|c|c|}
           \hline
	 & Gibbs High Repulsion & Gibbs Low Repulsion & Nystr{\"o}m High Repulsion& Nystr{\"o}m Low Repulsion\\
	\hline
	$m$ & 0.08 (0.07,0.08) & 0.1 (0.10,0.11)& 0.11 (0.1,0.11) & 0.11 (0.11,0.12)\\
          \hline
          $\alpha$ & 0.39 (0.31,0.45) & 0.92 (0.80,1) & 0.98 (0.82, 1) & 0.98 (0.90, 1)\\
	\hline
\end{tabular}
\end{center}
	\caption{The mean and 95$\%$ confidence interval for average movement, $m$ and the effective sample size coefficient, $\alpha$ for Gibbs samples and i.i.d. Nystr{\"o}m samples in high and low repulsion settings.}
	\label{table:Gibbsmix}
\end{table}

\section{Gibbs sampling for repulsive mixtures of Gaussians}
\begin{figure}[htb!]
	\begin{center}
		\begin{tabular}{cc}
\includegraphics[scale=0.4]{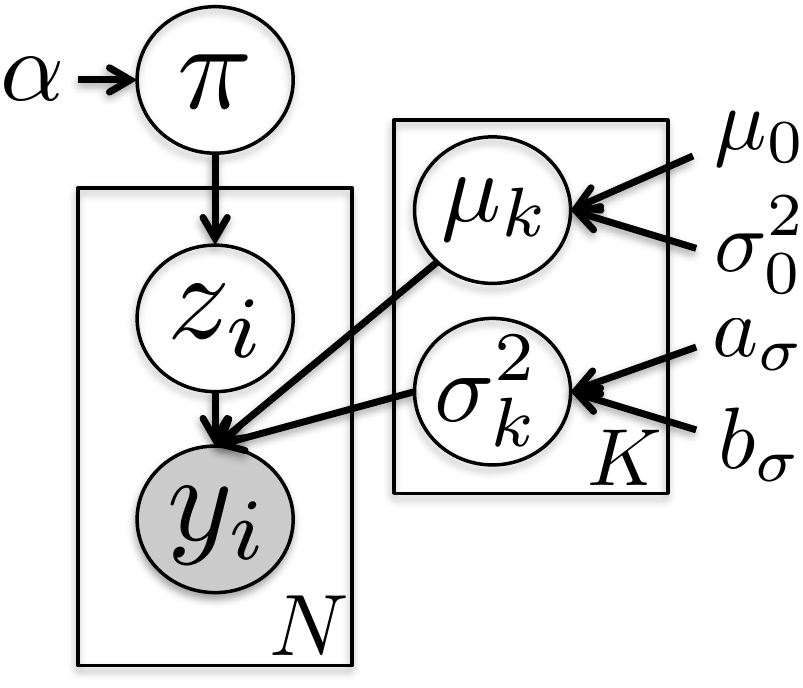} & 
\includegraphics[scale=0.4]{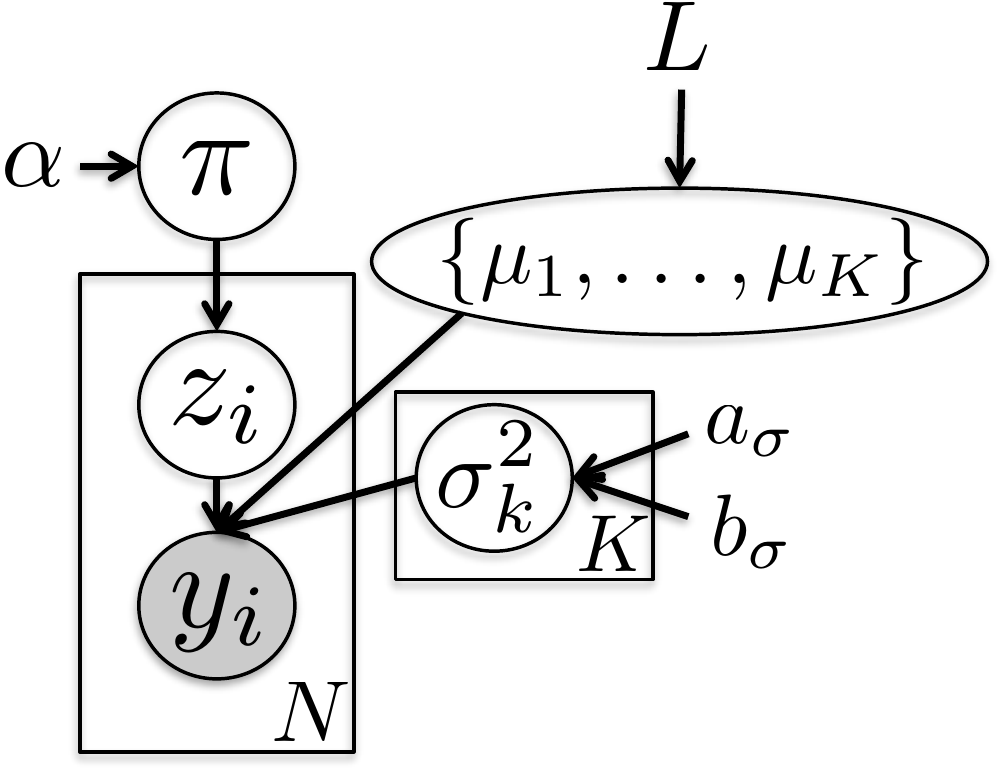}\\
\texttt{IID} & \texttt{DPP}
\end{tabular}
\end{center}
\precap
\caption{Graphical models for mixtures of Gaussians using \texttt{IID} and \texttt{DPP} priors on the location parameters.}
\label{fig:DPPgraph}
\end{figure}

\paragraph{Generative Model} We consider a Bayesian mixture of Gaussians with either an independent normal (\texttt{IID}) or $K$-DPP (\texttt{DPP}) prior on the location parameters. In both cases, the $K$-component model with $N$ observations is specified as:
\begin{align}
	\begin{aligned}
	\pi \mid \alpha &\sim \mbox{Dir}(\alpha,\dots,\alpha)\\
	\sigma_k^2 \mid a_{\sigma},b_\sigma &\sim \mbox{IG}(a_\sigma,b_\sigma), \quad k=1,\dots,K\\
	\{\mu_1,\dots,\mu_K\} &\sim F\\
	z_i \mid \pi &\sim \pi, \quad i=1,\dots, N\\
	y_i \mid \pi, \{\mu_k,\sigma_k^2\} &\sim N(\mu_{z_i},\sigma_{z_i}^2), \quad i=1,\dots,N.
\end{aligned}
\end{align}
Here, $\mbox{IG}$ denotes the inverse gamma distribution and $\mbox{Dir}$ a $K$-dimensional Dirichlet. For simplicity, we consider the univariate case here, though the multivariate case follows directly by considering an inverse Wishart prior in place of the inverse gamma and likewise modifying $F$ accordingly.  Such a multivariate case is examined in the \emph{iris} classification example in the main paper.  

The difference between the models is in how the location parameters are specified. For the \texttt{IID} case, we simply have:
\begin{align}
	\mu_k \mid \mu_0,\sigma_0^2 \sim N(\mu_0,\sigma_0^2)
\end{align}
For the \texttt{DPP} case, we jointly sample:
\begin{align}
	\{\mu_1,\dots,\mu_K\}\mid L &\sim \mbox{$K$-DPP}(L).
\end{align}
We consider $L$ decomposed into Gaussian quality and similarity terms:
\begin{align}
L(\mu_m,\mu_n)=q(\mu_m)k(\mu_m,\mu_n)q(\mu_n),
\end{align}
with
\begin{align}
k(\mu_m,\mu_n)=\exp\left\{-\frac{(\mu_m-\mu_n)^2}{\gamma_0^2}\right\}, \quad q(\mu_m)= N(\mu_0,2\sigma_0^2).
\end{align}
\begin{algorithm}[t]
\begin{algorithmic}
  \STATE {\bfseries Input:} Previous mixture weights $\pi$, emission parameters $\{\mu_k,\sigma_k\}^2$.\\
  \FOR{$i=1,\dots,N$}
  \STATE Sample cluster indicators $z_i \mid y_i,\{\mu_k,\sigma_k^2\},\pi_k \propto \frac{1}{C_i} \sum_{k=1}^K \pi_k N(y_i; \mu_k,\sigma_k^2) \delta(z_i,k)$\\
  \ENDFOR
  \STATE Sample mixture weights $ \pi \mid \{z_i\},\alpha \sim \mbox{Dir}(\alpha+N_1,\dots,\alpha+N_K)$\\
  %\STATE Sample permutation $\omega\sim\frac{1}{k!}\prod_{i=1}^Np(y_i|z_i,\{\mu_k,\sigma_k\},\omega)$\\
  \FOR{$k=1,\dots,K$}
  \STATE Sample scale parameters $\sigma_k^2 \mid \{y_i : z_i=k\},\mu_k, a_\sigma,b_\sigma \sim \mbox{IG}\left(a_\sigma + \frac{N_k}{2},b_\sigma + \frac{1}{2} \sum_{i:z_i=1}(y_i - \mu_k)^2\right)$\\
  \ENDFOR
  \STATE Sample location parameters $\{\mu_1,\dots,\mu_K\} \mid \{y_i\},\{z_i\},\{\sigma_k^2\} \sim F_{post}$\\
  \STATE {\bfseries Output:} New mixture weights $\pi$, emission parameters $\{\mu_k,\sigma_k^2\}$.
\end{algorithmic}
\caption{Mixture of Gaussians sampler}
\label{alg:MoG}
\end{algorithm}
\paragraph{Gibbs sampling} For the uncollapsed setting, where mixture weights $\pi$ and emission parameters $\{\mu_k,\sigma_k^2\}$ are sampled, Algorithm~\ref{alg:MoG} summarizes the Gibbs sampler for the finite mixture of Gaussians.  We write the algorithm generically so that the overlap between \texttt{IID} and \texttt{DPP} is clear.  In particular, the locations are sampled from $F_{post}$, which generically refers to the full conditional of the cluster means.  For the \texttt{IID} case, we sample i.i.d. for each $k$ from
\begin{align}
	\mu_k \mid \{y_i : z_i=k\},\sigma_k^2,\mu_0,\sigma_0^2 \sim N\left(\hat{\mu}_k,  \hat{\sigma}_k^2\right),
\end{align}
where
$\hat{\mu}_k=\left(\frac{1}{\sigma_0^2} + \frac{N_k}{\sigma_k^2}\right)^{-1}\left(\frac{\mu_0}{\sigma_0^2} + \frac{1}{\sigma_k^2} \sum_{i:z_i=k} y_i\right)$ and
$\hat{\sigma}_k^2=\left(\frac{1}{\sigma_0^2} + \frac{N_k}{\sigma_k^2}\right)^{-1}$.
Here, $N_k = |\{y_i : z_i=k\}|$, i.e., the cardinality of the set of observations assigned to cluster $k$.

For \texttt{DPP}, note that $p(\{\mu_j\}_{j=1}^k|\{y_i\},\{z_i\},\{\mu_k,\sigma_k^2\})\propto \det(L_{\mu_1,\ldots,\mu_k})\prod_{j=1}^k \prod_{i:z_i=j} N(y_i;\mu_j,\sigma_j^2)$.
Unfortunately, this posterior distribution is not a $k$-DPP. However, fixing the rest of $k-1$ centroids, the full conditional of $\mu_k$ is (dropping constant terms that do not depend on $\mu_k$)
\begin{align}
p(\mu_k|\{y_i\},\{z_i\},\{\mu_j,\sigma_j^2\}_{j \neq k},\sigma_k^2)\propto \det(L_{\mu_1,\ldots,\mu_k})\prod_{i:z_i=k} N(y_i;\mu_k,\sigma_k^2).
\end{align}

As before, we can use Schur's determinantal equality \cite{schur1917potenzreihen} to get
\begin{align}
\det(L_{\mu_1,\ldots,\mu_k})&\propto L(\mu_k,\mu_k)-\sum_{i,j\neq k}M^{\backslash k}_{ij}L(\mu_i,\mu_k)L(\mu_j,\mu_k)\\
& = q^2(\mu_k)\left(1-\sum_{i,j\neq k}M^{\backslash k}_{ij}q(\mu_i)k(\mu_i,\mu_k)k(\mu_j,\mu_k)q(\mu_j)\right).
\end{align}
Combining the previous two equations, we get the full conditional
\begin{align}  
p(\mu_k|\{y_i\},\{z_i\},\{\mu_j,\sigma_j^2\}_{j \neq k},\sigma_k^2)\propto q^2(\mu_k)\left(1-\sum_{i,j\neq k}M^{\backslash k}_{ij}q(\mu_i)k(\mu_i,\mu_k)k(\mu_j,\mu_k)q(\mu_j)\right)\prod_{i:z_i=k} N(y_i;\mu_k,\sigma_k^2).
\end{align}

The CDF of the distribution above can be computed easily, since it only involves exponential quadratic forms. The inverse CDF method can then be used to obtain a sample from the above distribution. Note once again that $q^2(\mu_k)\prod_{i:z_i=k} N(y_i;\mu_k,\sigma_k^2)$ is defined to be exactly the same as the Gaussian distribution where $\mu_k$ would have been sampled from in the \texttt{IID} case. Thus the equation above gives a nice intuition on the conditional density of $\mu_k$ in the \texttt{DPP} setting: it is an exponentially tilted distribution in which $q^2(\mu_k)\prod_{i:z_i=k} N(y_i;\mu_k,\sigma_k^2)$ is corrected by a factor that depends on the location of the other centroids. In the case where all of the other centroids are far away from the cluster center $\hat{\mu}_k$, the correction factor is close to one and we would recover the density for the \texttt{IID} case.  

\begin{figure}
\begin{center}
\includegraphics[scale=0.5]{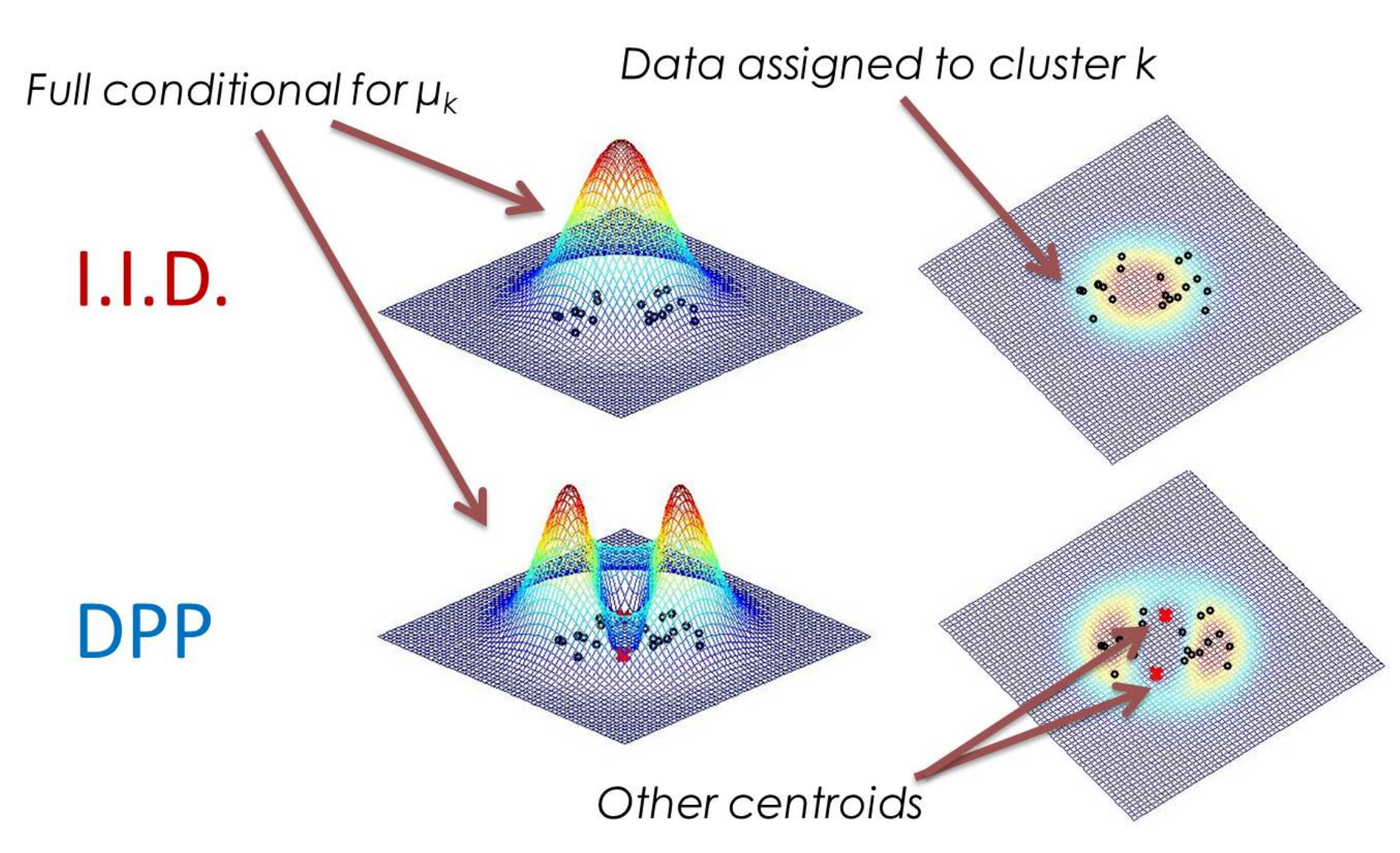}
\end{center}
\caption{Comparison between the full conditional for $\mu_k$ using the \texttt{IID} and \texttt{DPP} models at a given iteration $m$ of the sampler.}
\label{fig:CondDist}
\end{figure}

To get a sense of why the \texttt{DPP} leads to more diverse cluster centers than \texttt{IID}, consider the full conditional for $\mu_k$ at some iteration $m$ of our sampler, as visualized in Fig.~\ref{fig:CondDist}.  We have some data points currently assigned to cluster $k$ via cluster indicators $z_i=k$.  The \texttt{IID} model assumes that $\mu_k$ is independent of the other $\mu_j$'s whereas the \texttt{DPP} conditions on the other cluster centers leading to a conditional distribution for $\mu_k$ that puts more mass on uncovered regions.  In subsequent iterations, the data that had been assigned to cluster $k$ but are not well covered by the sampled (and repulsed) $\mu_k$ will instead be assigned to one of the existing cluster centers that have mass near that data item.  Such an alternative cluster exists, and is why $\mu_k$ was repulsed from that region, or will likely exist in future draws.

One attractive aspect of our \texttt{DPP} formulation is the fact that the sampling strategy maintains nearly the same simplicity as the standard \texttt{IID} sampler.  This is in contrast, for example, to the repulsive mixture formulation of~\cite{petralia2012repulsive} which relied on slice sampling and draws from truncated normals, where the truncating region could only be computed in closed form for a restricted set of repulsive functions.

\section{Additional details on experiments}

\subsection{Hyperparameter settings}

For our mixture of Gaussian experiments, we used an inverse Wishart $\mbox{IW}(\nu,\Psi)$ with $\nu=d+1$ and $\Psi=I$, which corresponds to $a_\sigma =2$ and $b_\sigma=1$ for the inverse Gamma in 1-dimension.  Here, we use an inverse Wishart specification such that $\Sigma \sim \mbox{IW}(\nu,\Psi)$ has mean $E[\Sigma] = \frac{\Psi}{\nu-d+1}$.  The Dirichlet hyperparameters were set to $\alpha=\frac{1}{3}$, just as in~\cite{petralia2012repulsive}.  For the location hyperparameters, in the \texttt{IID} case we set $\mu_0=0$ and $\sigma_0^2=1$. In the \texttt{DPP} case, we use $\mu_0$ and $\sigma_0^2$ as in the \texttt{IID} case and set the repulsion parameter $\rho_0^2 = 1$.

For the MoCap experiment, we computed the covariance estimate from the training data, and set the similarity covariance parameter $\Sigma$ equal to this estimate. We then take the quality covariance parameter to be $\Gamma=\frac{1}{2}\Sigma$.  

\subsection{Additional figures for MoCap experiments}

\begin{figure}
\centering
	\begin{tabular}{ccc}
\hspace{-0.3in}\includegraphics[scale=0.23]{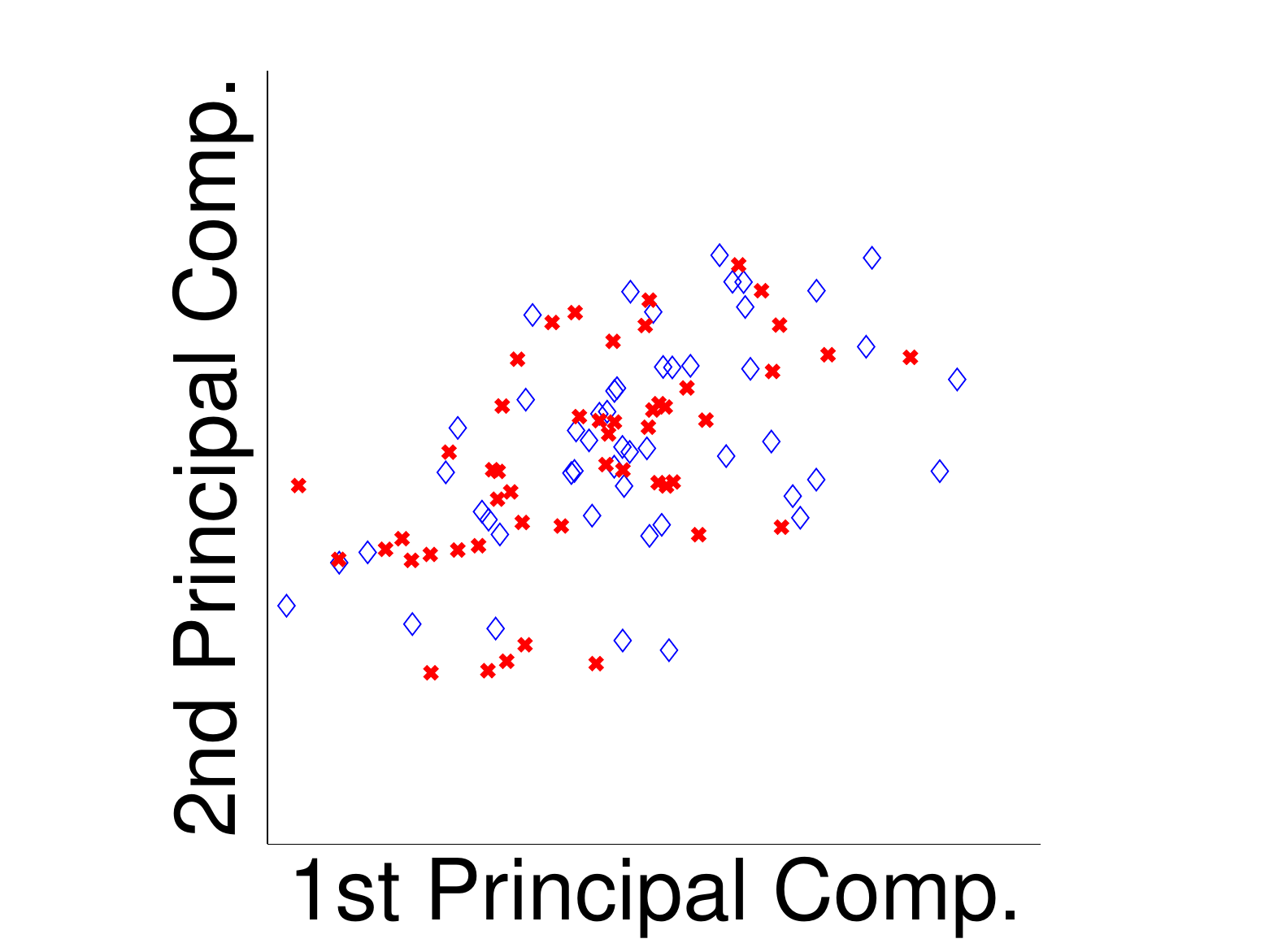}&\hspace{-0.4in}
\includegraphics[scale=0.23]{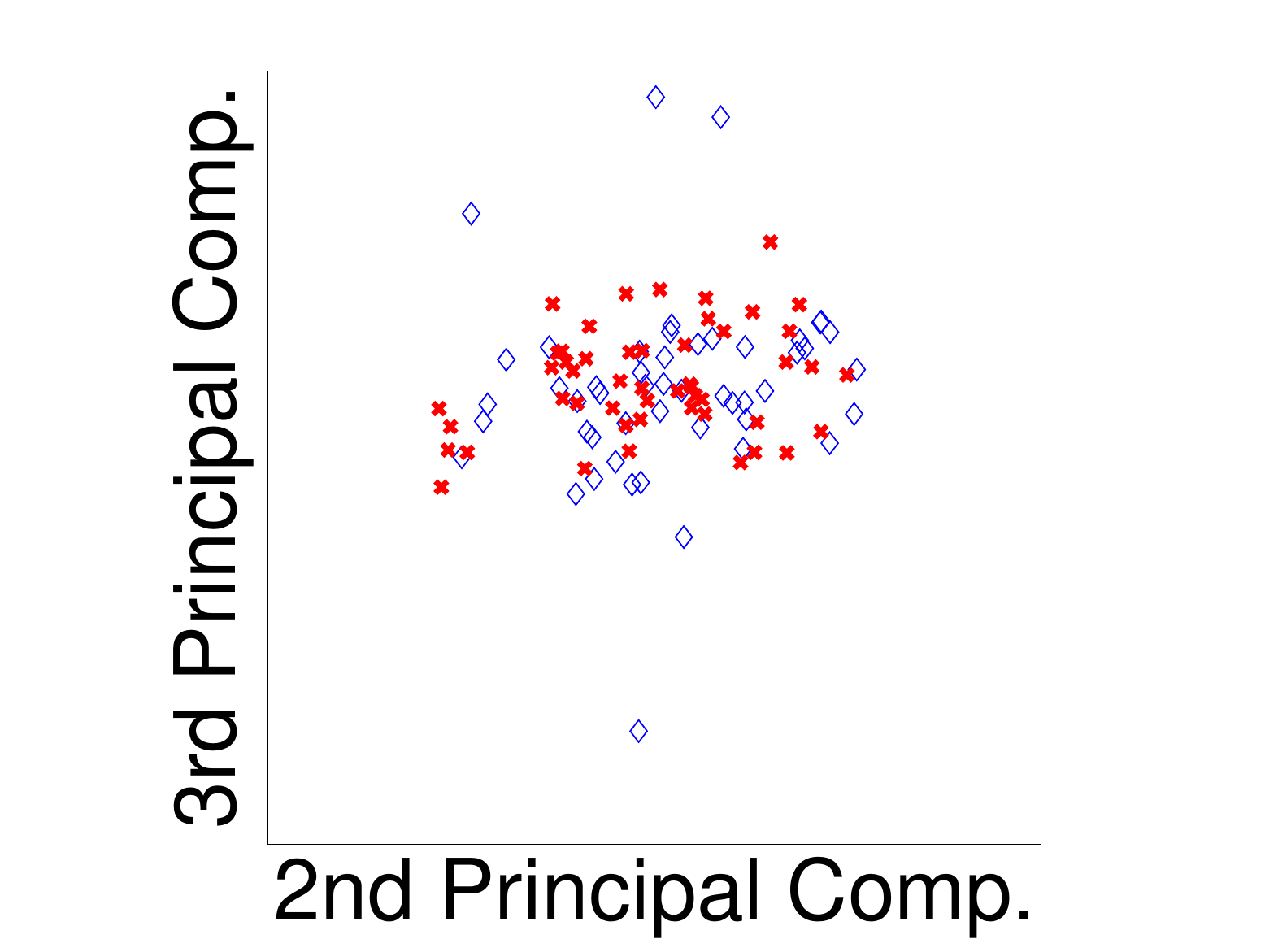}&\hspace{-0.4in}
\includegraphics[scale=0.23]{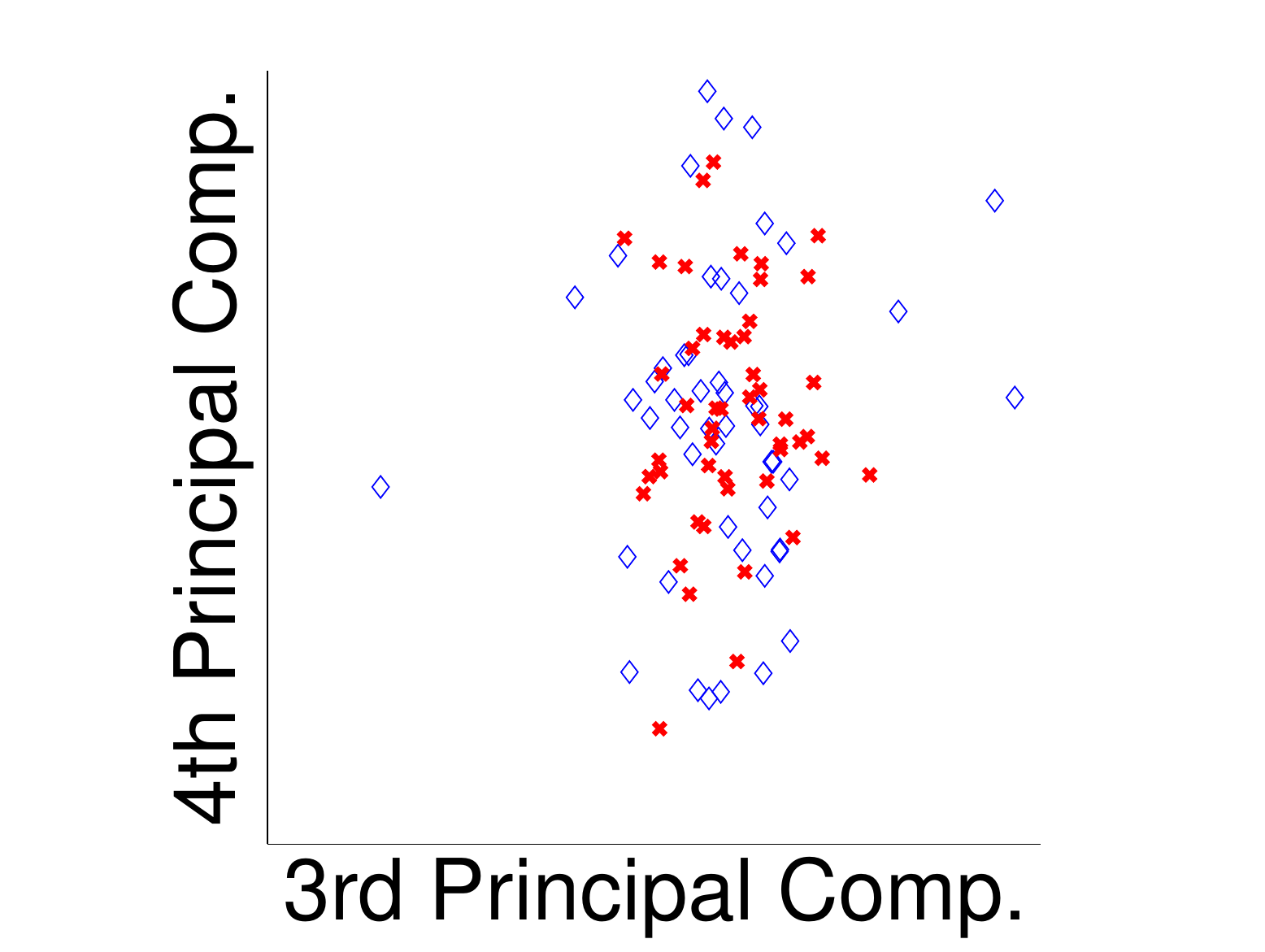} \\
{\small (a)} & {\small (b)} & {\small (c)}
\end{tabular}
\precap
\caption{\small (a)-(c) DPP (blue) and i.i.d. multivariate Gaussian (red) samples projected onto the top 4 principal components of the \emph{dance} data.}
\label{fig:MoCapPCA}
\vspace{-0.05in}
\end{figure}

In Fig.~\ref{fig:MoCapPCA}, we provide a visualization of poses sampled from the DPP relative to i.i.d. sampling of poses from a multivariate Gaussian.  From these plots, we see how the sample of poses from the DPP covers a broader space, even when the covariance of the multivariate Gaussian is inflated to match that of the DPP.  The reason for this broader coverage is the fact that the under the DPP, sampled poses repulse from regions already covered by other sampled poses.

\begin{figure}[htb]
	\begin{center}
	\begin{tabular}{ccccc}
\multicolumn{5}{c}{\includegraphics[scale=0.3]{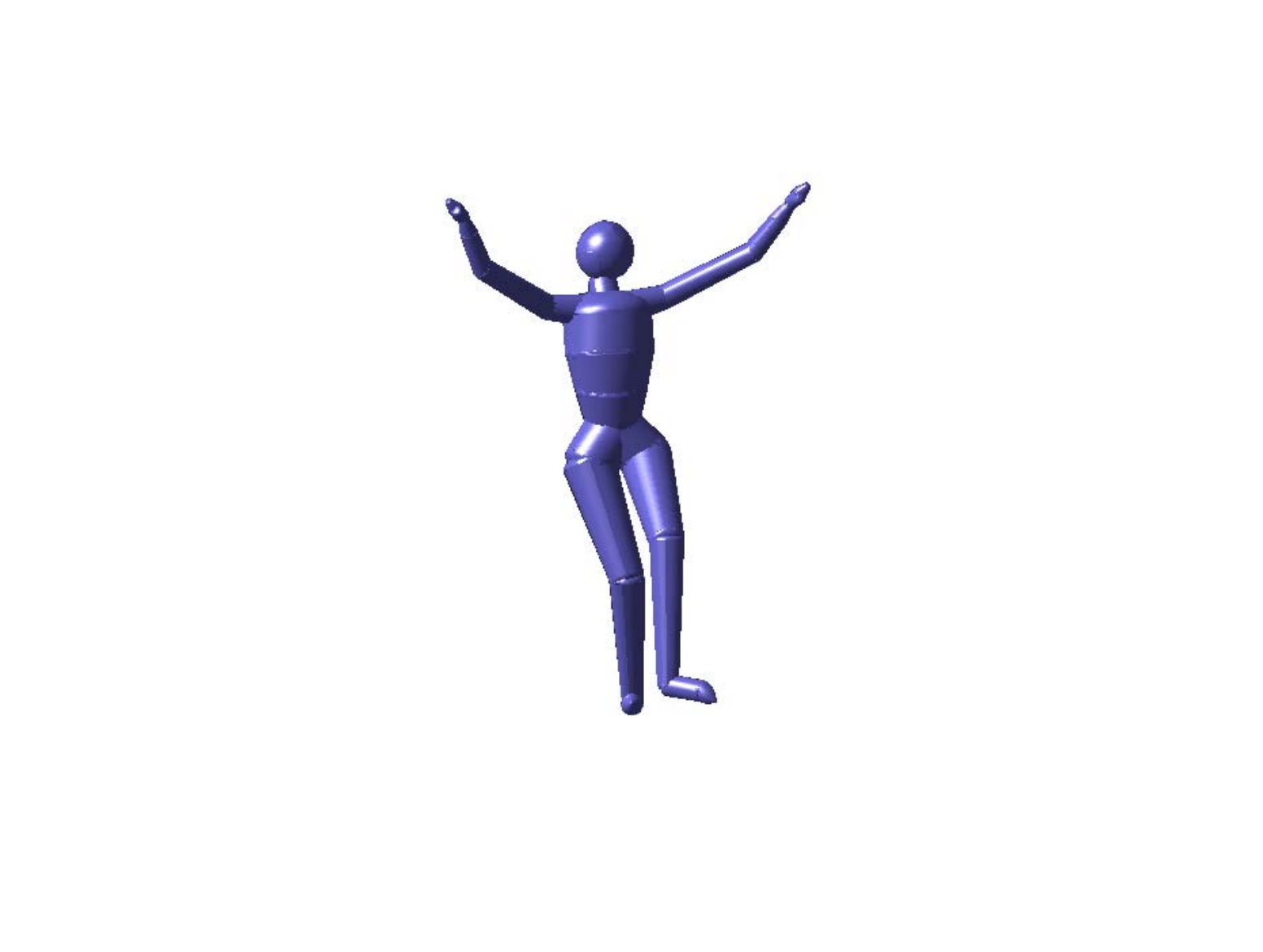}}\\
\multicolumn{5}{c}{Original Pose}\\
\includegraphics[scale=0.2]{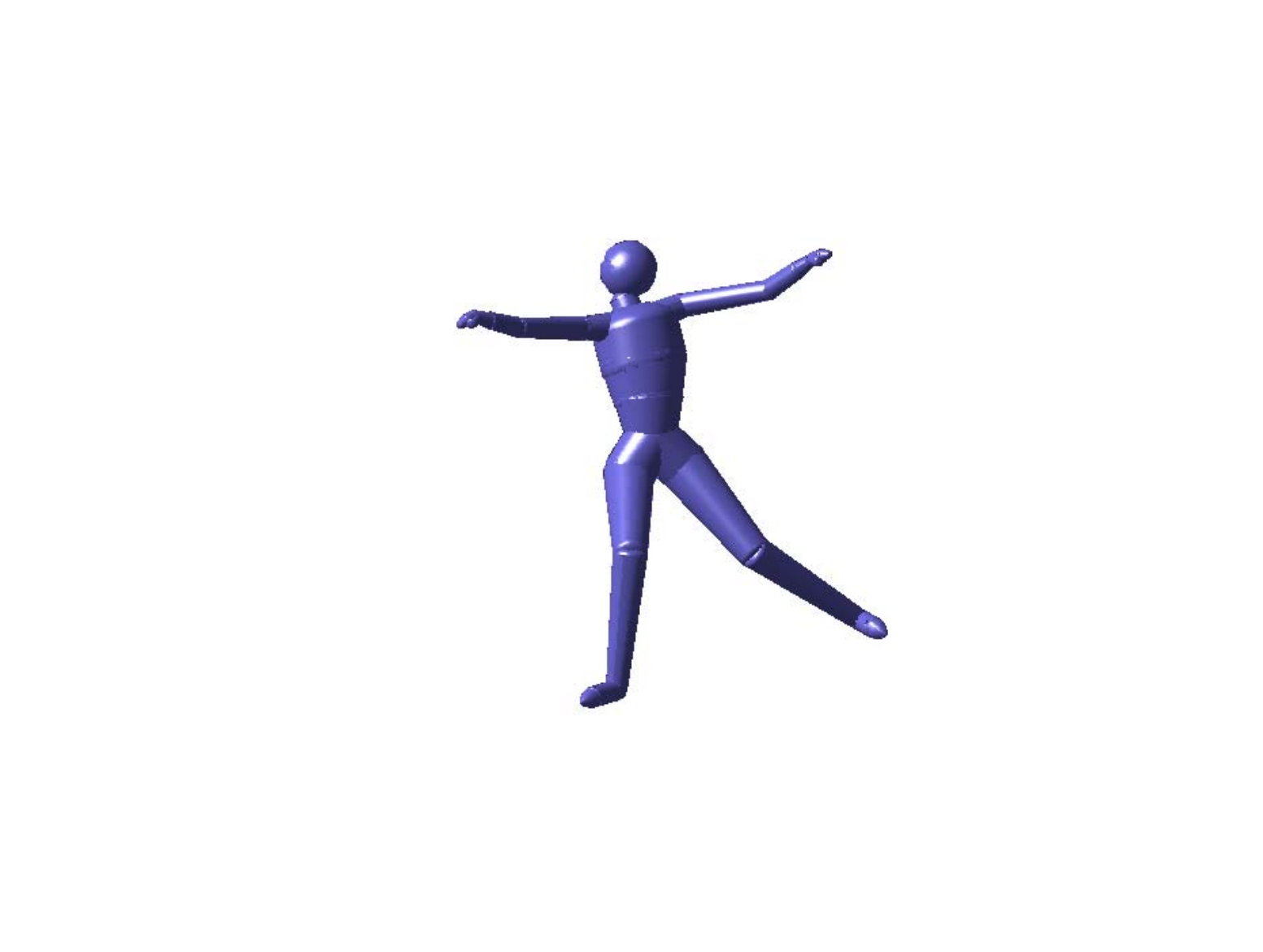} &
\includegraphics[scale=0.2]{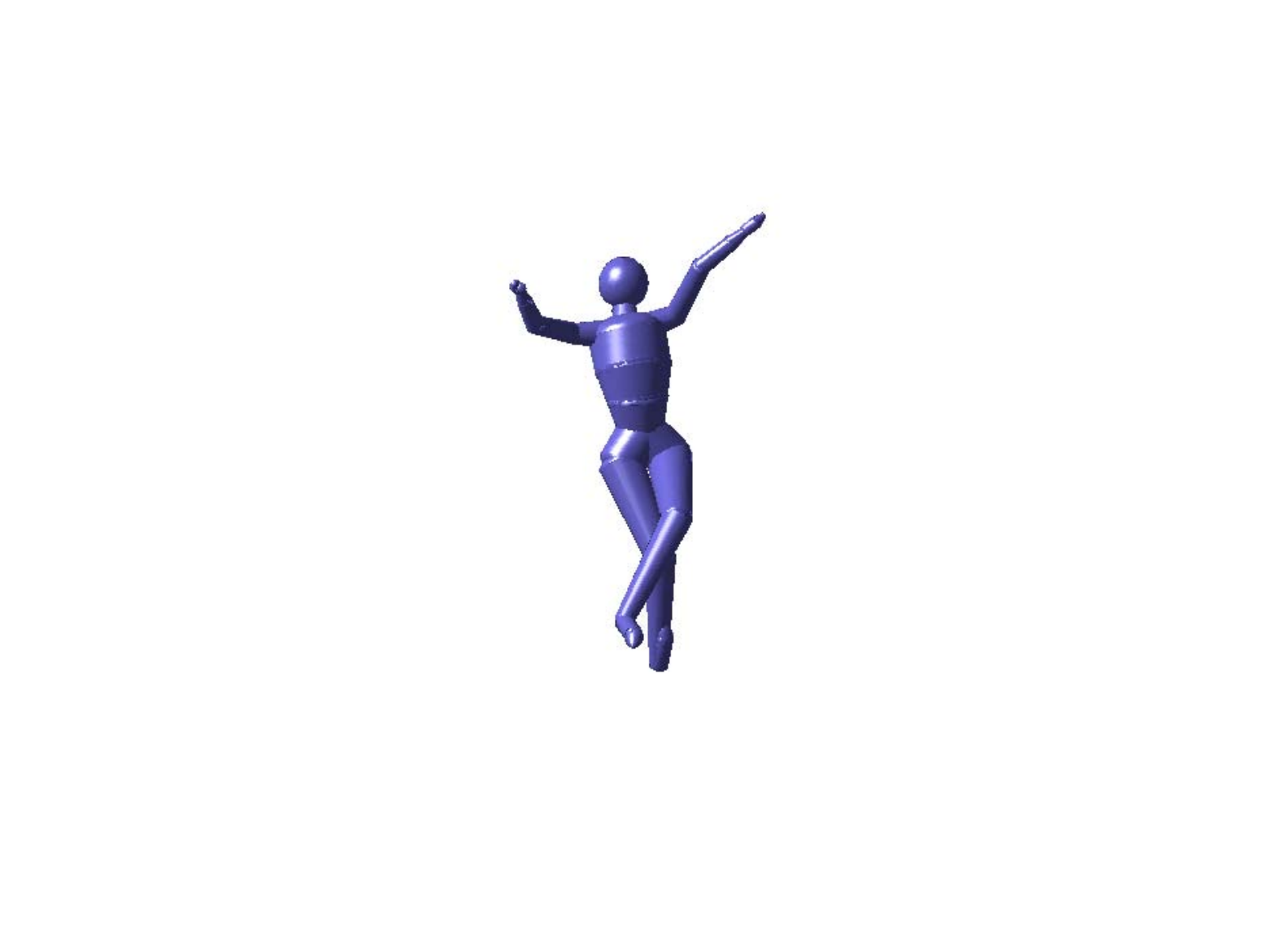} &
\includegraphics[scale=0.2]{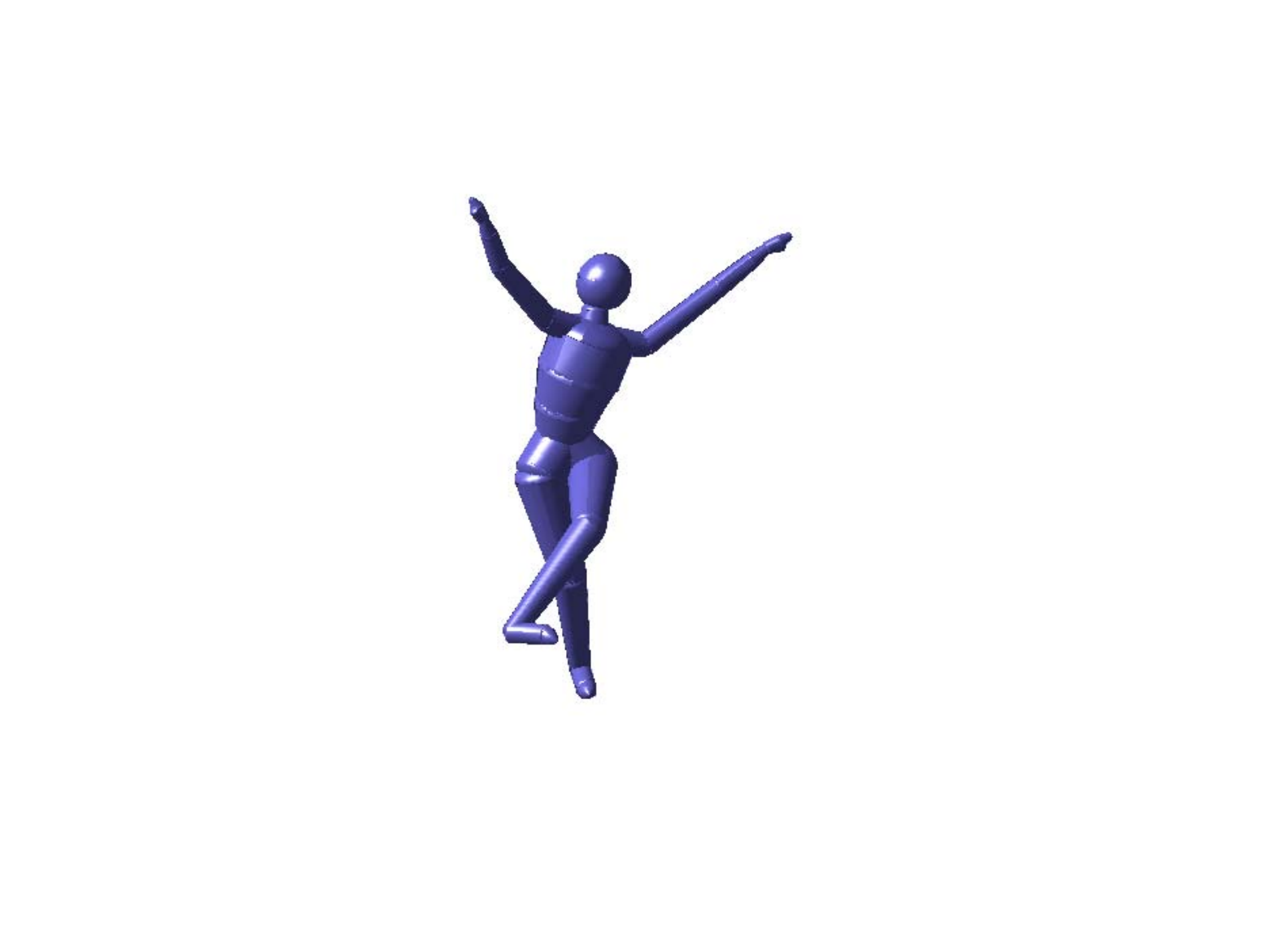} &
\includegraphics[scale=0.2]{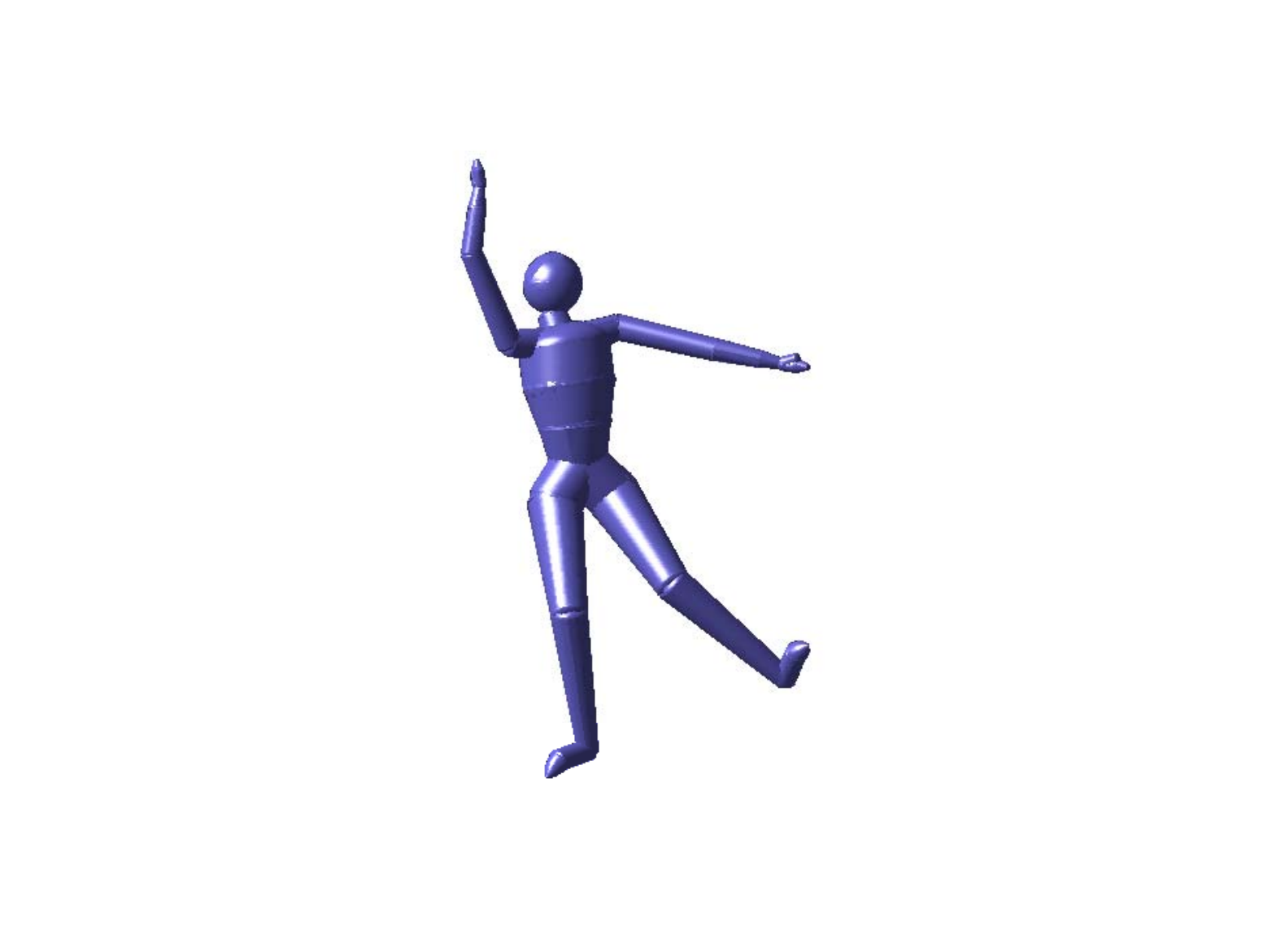} &
\includegraphics[scale=0.2]{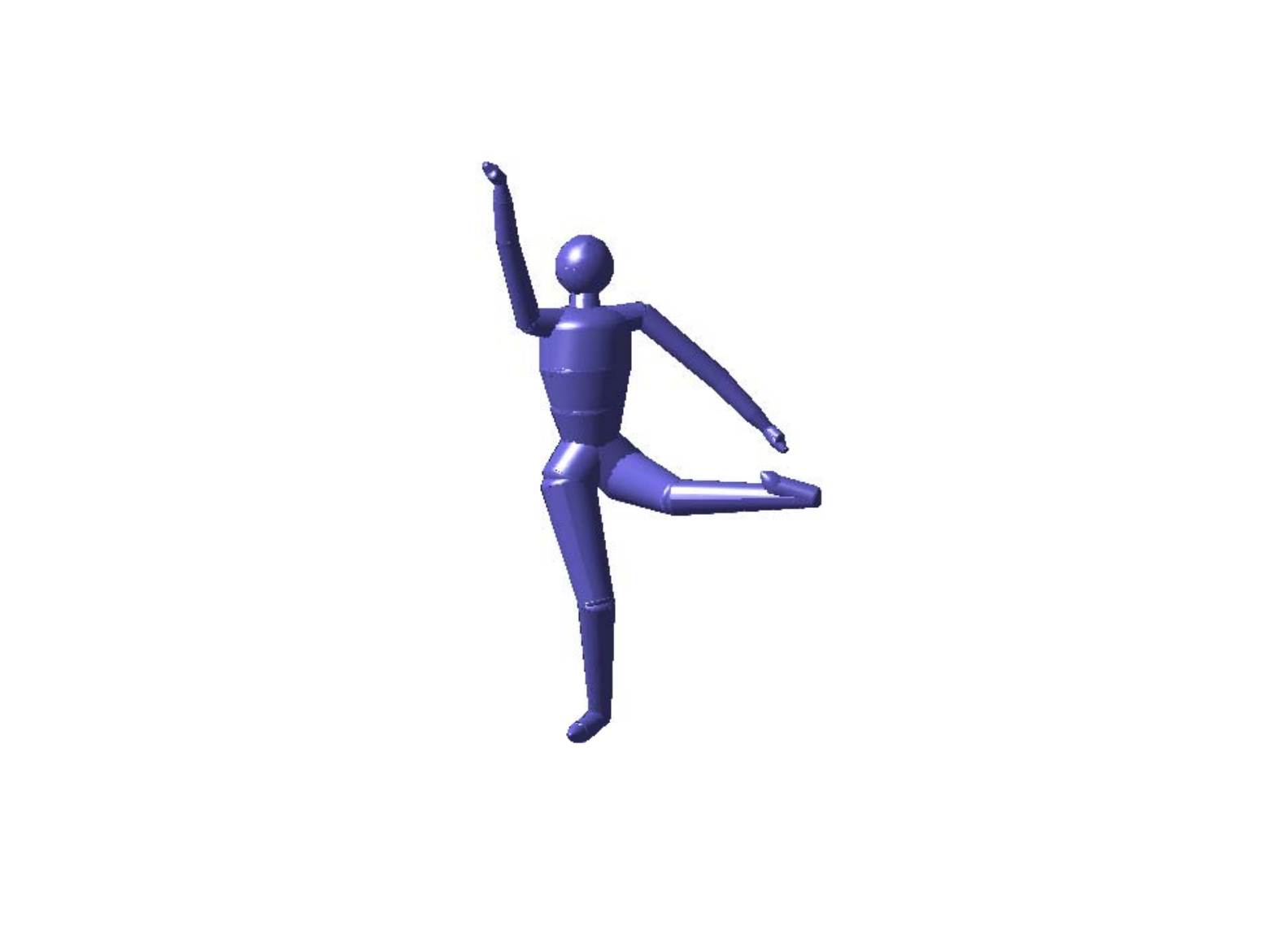}\\
\multicolumn{5}{c}{Poses synthesized from i.i.d. draws from a multivariate Gaussian}\\
\includegraphics[scale=0.2]{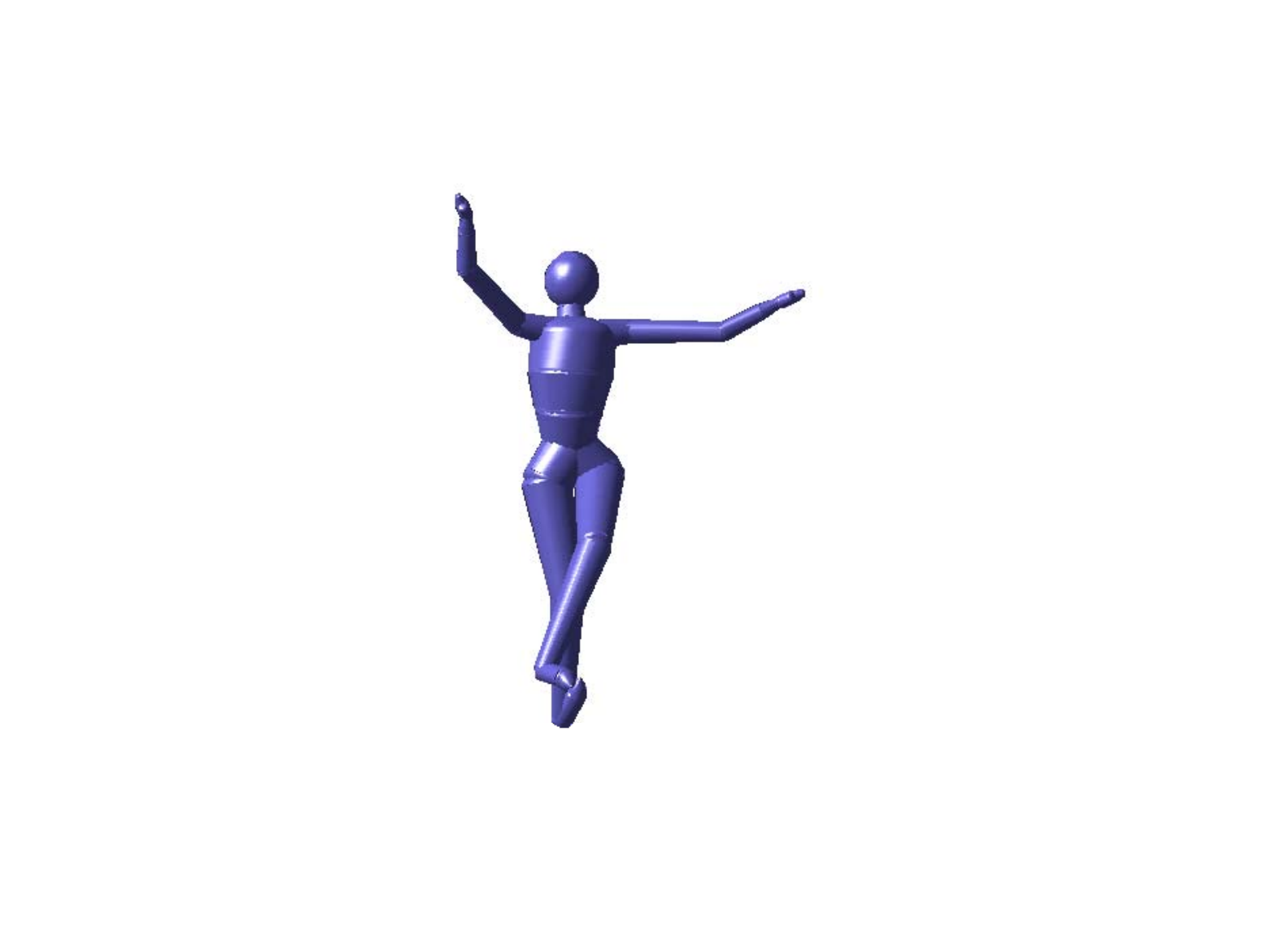} &
\includegraphics[scale=0.2]{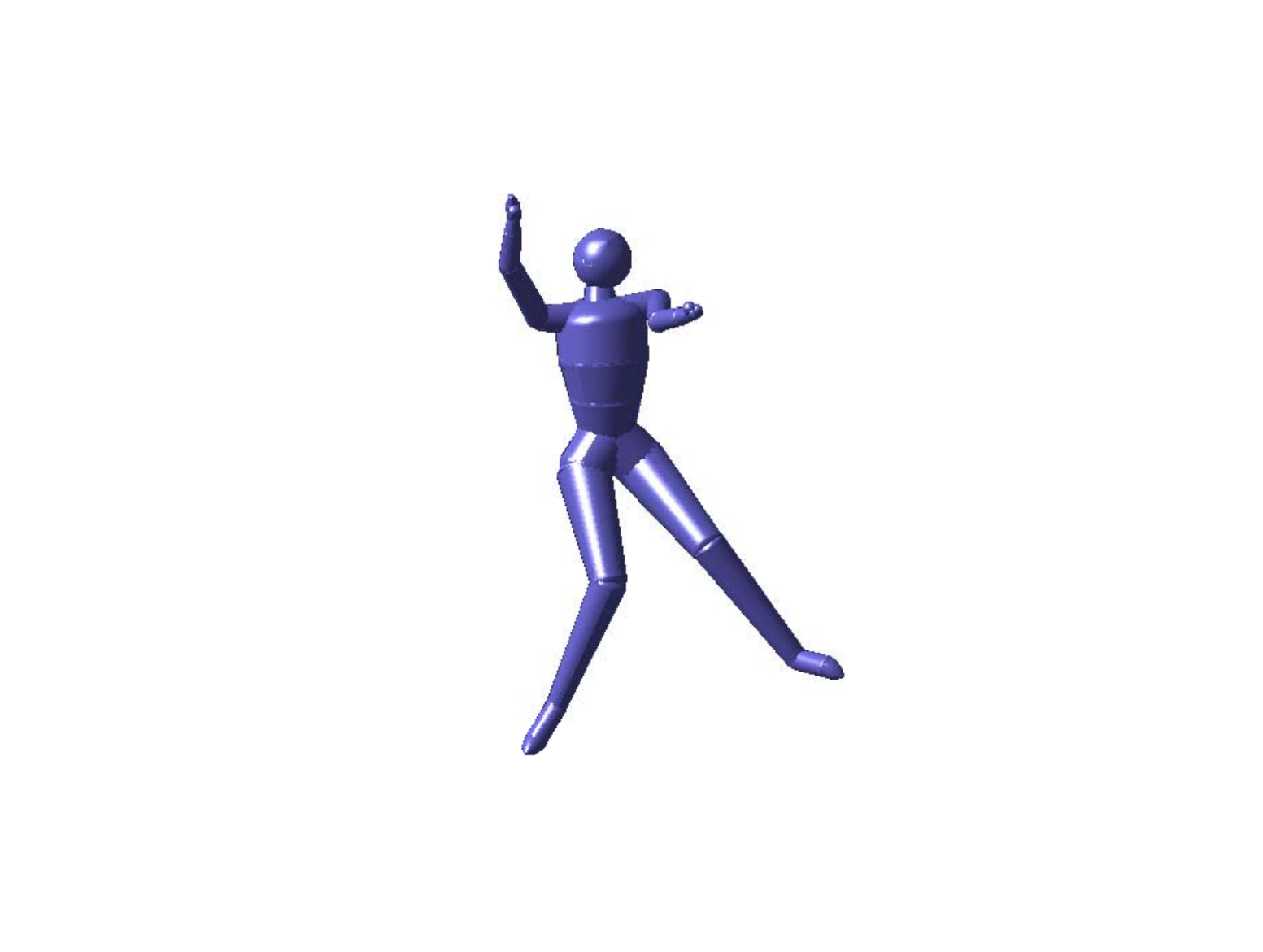} &
\includegraphics[scale=0.2]{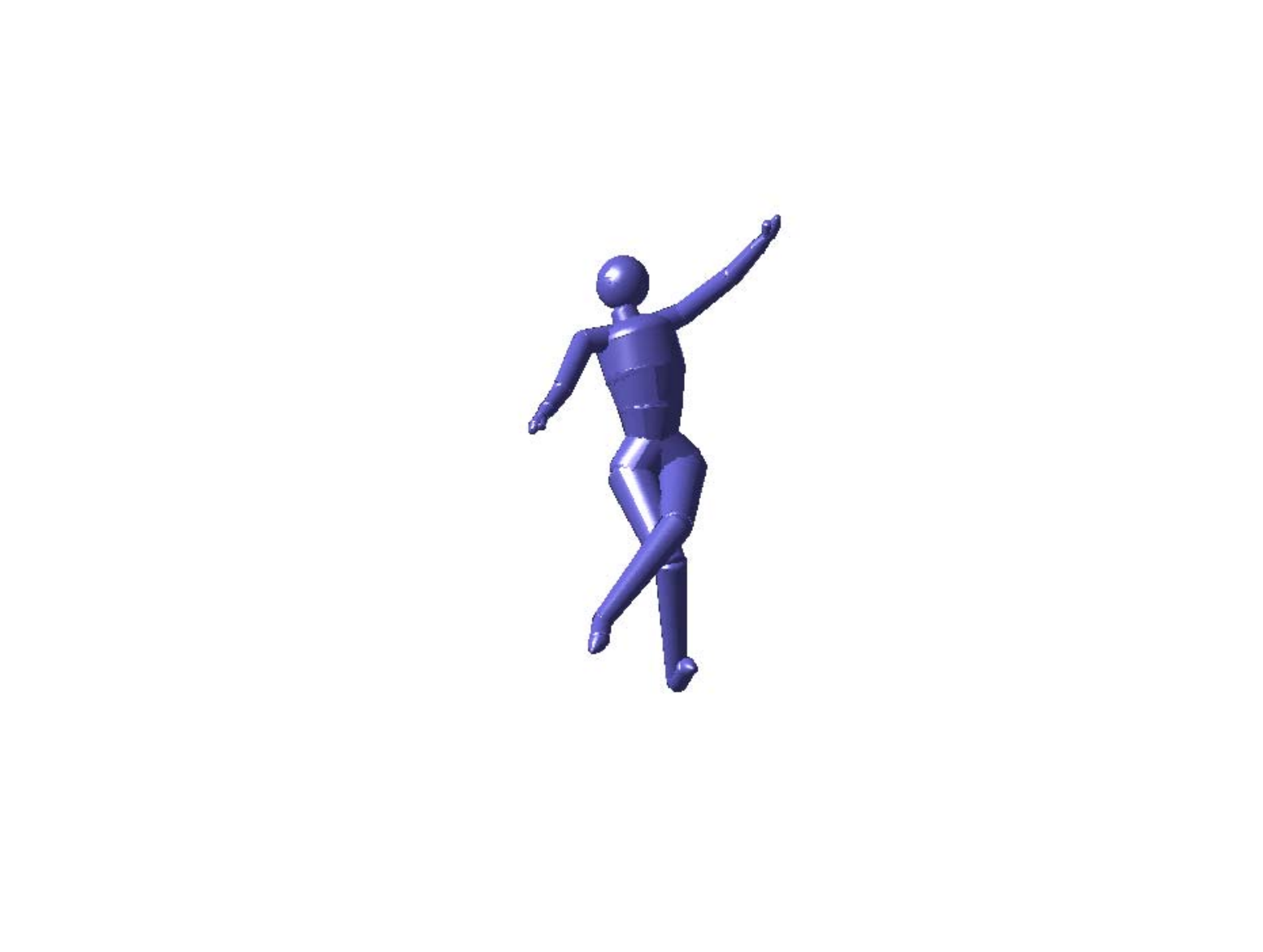} &
\includegraphics[scale=0.2]{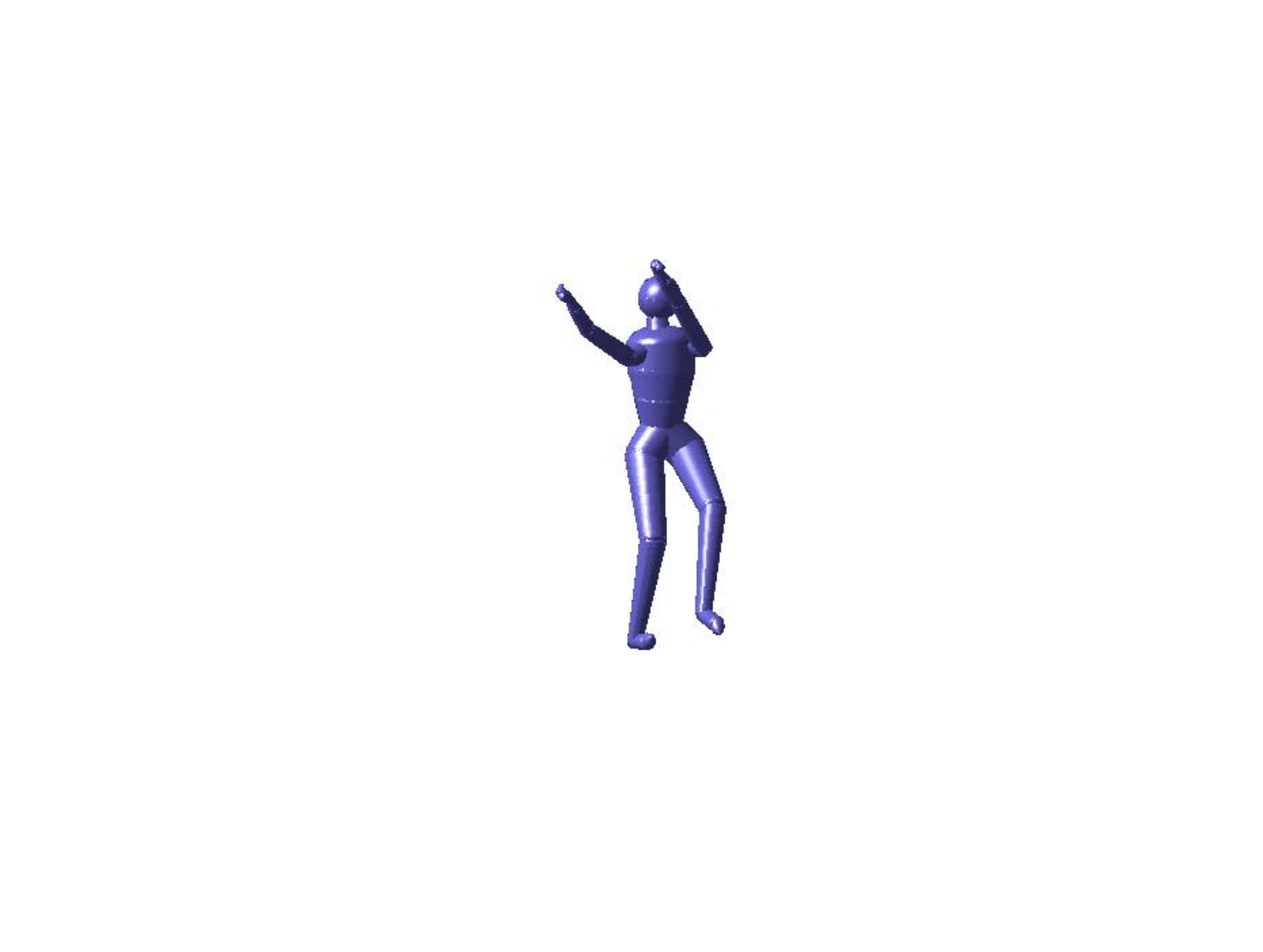} &
\includegraphics[scale=0.2]{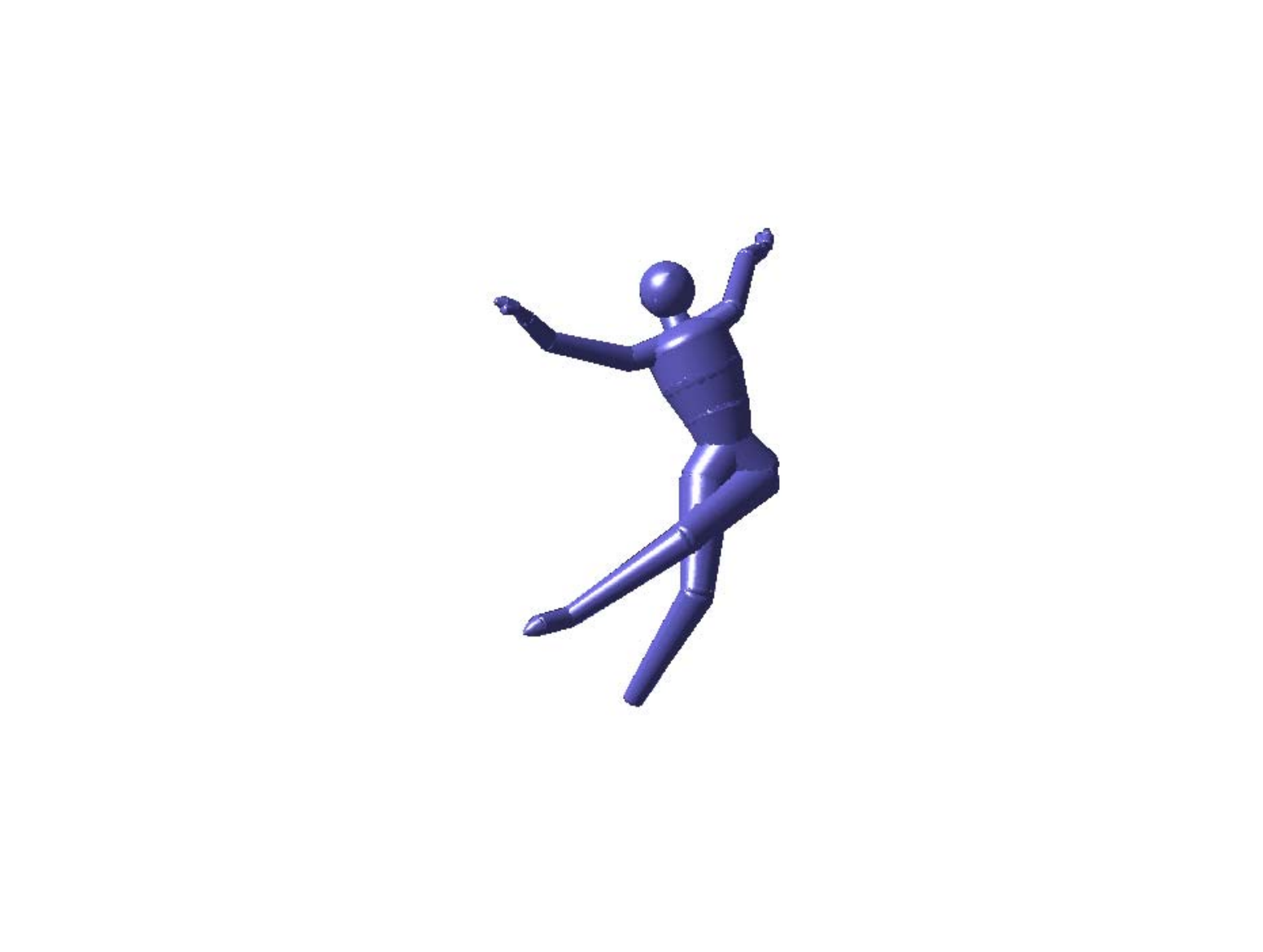}\\
\multicolumn{5}{c}{Poses synthesized from an RFF-approximated DPP}\\
\includegraphics[scale=0.2]{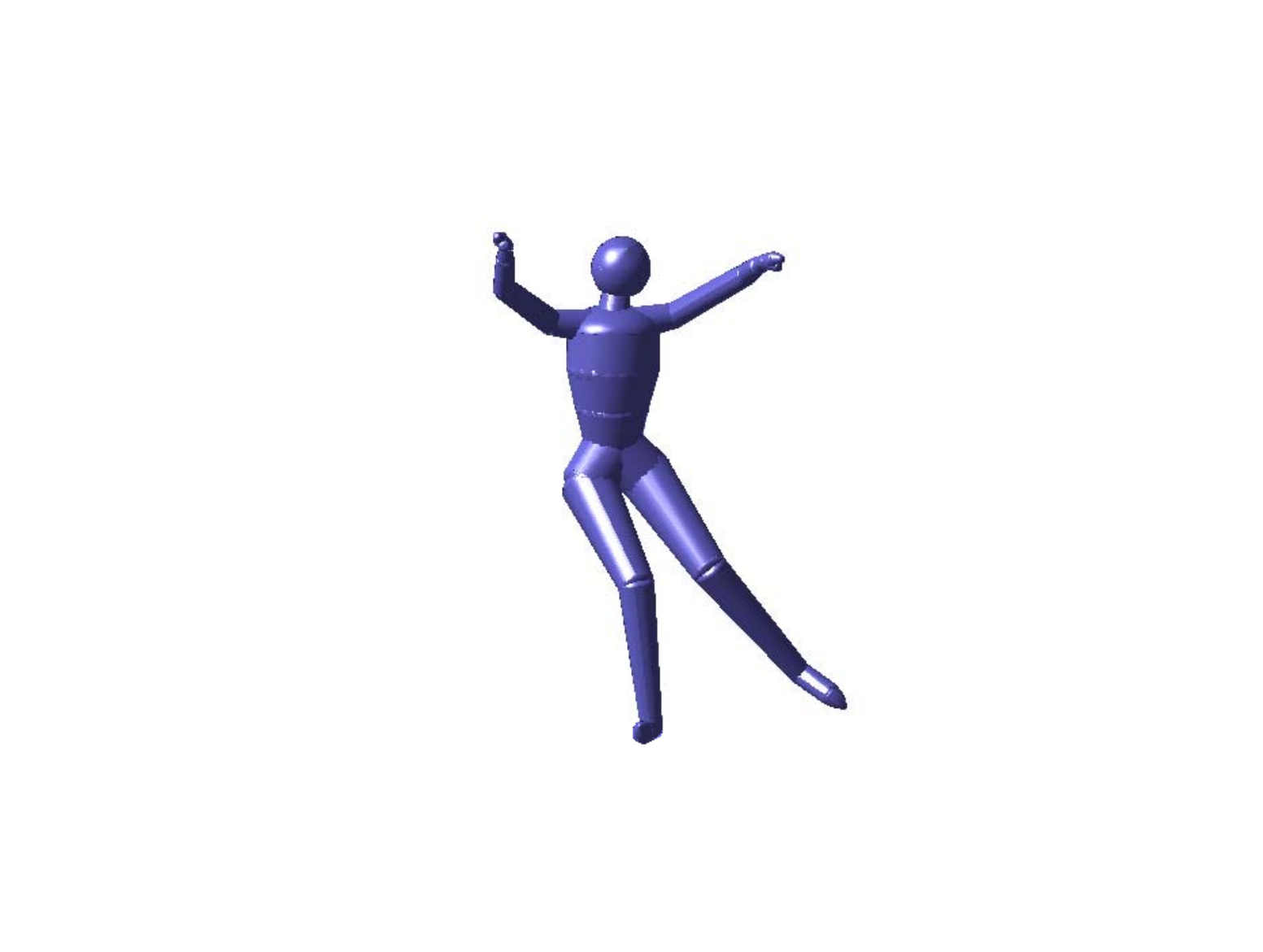} &
\includegraphics[scale=0.2]{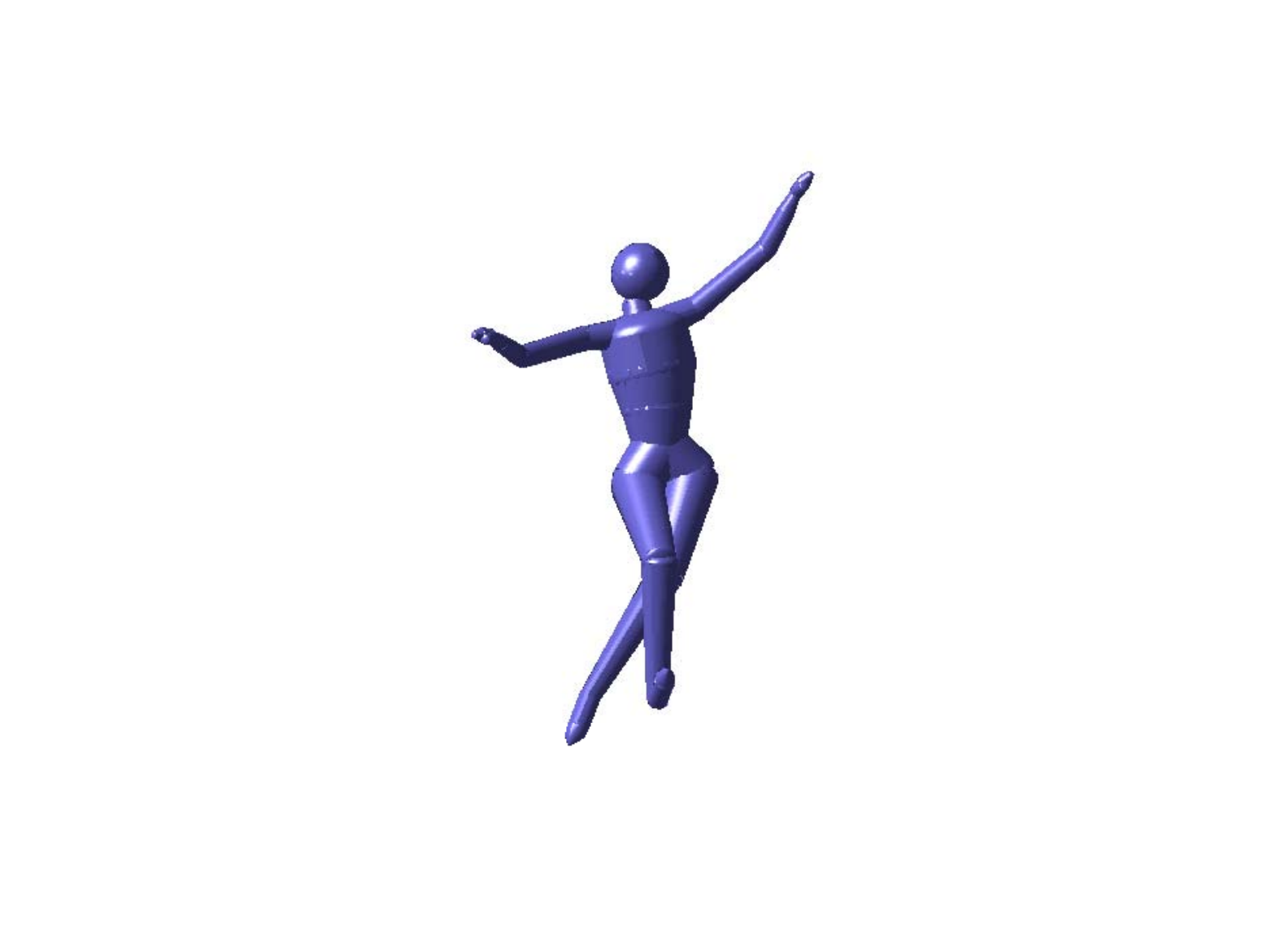} &
\includegraphics[scale=0.2]{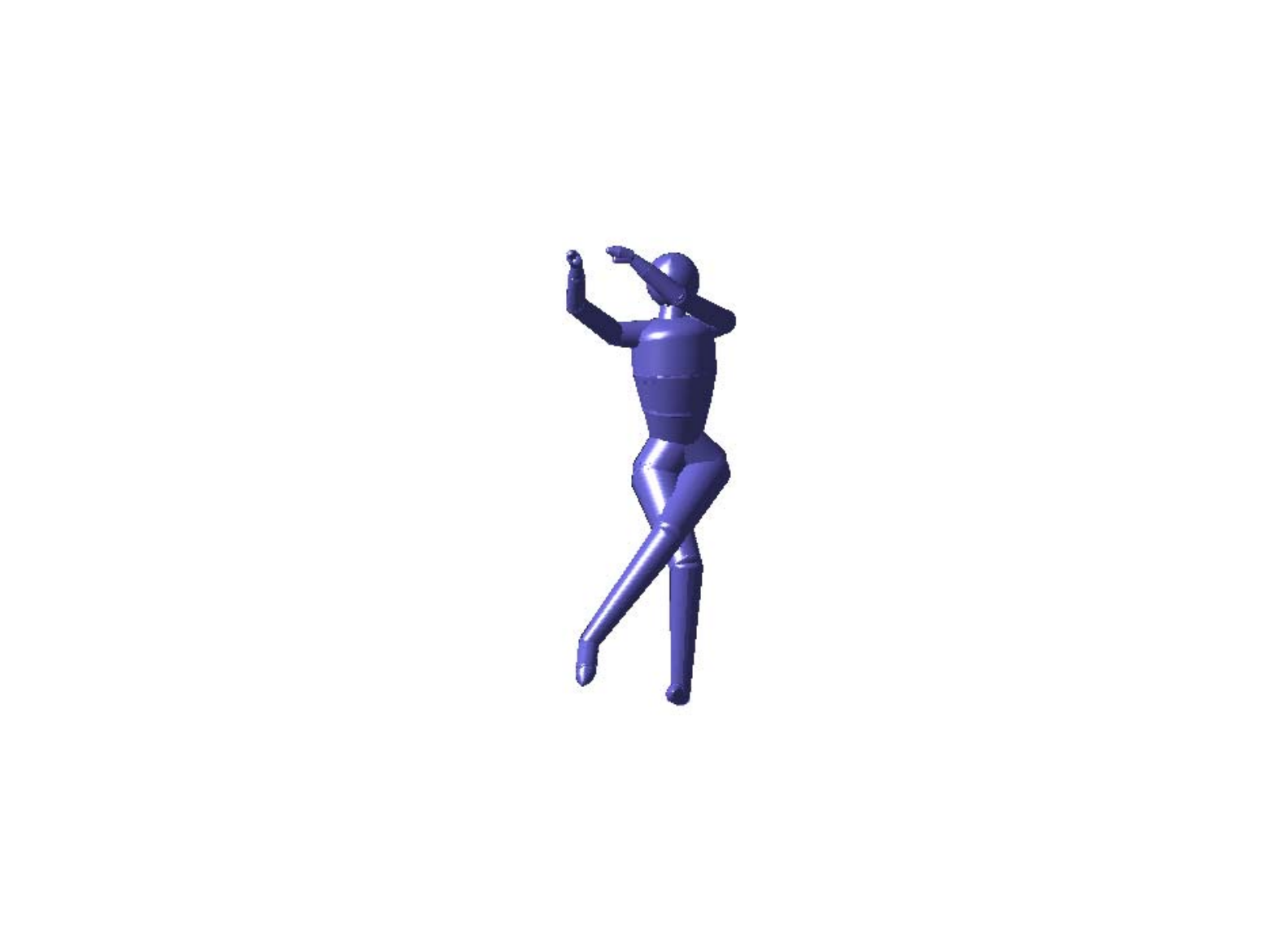} &
\includegraphics[scale=0.2]{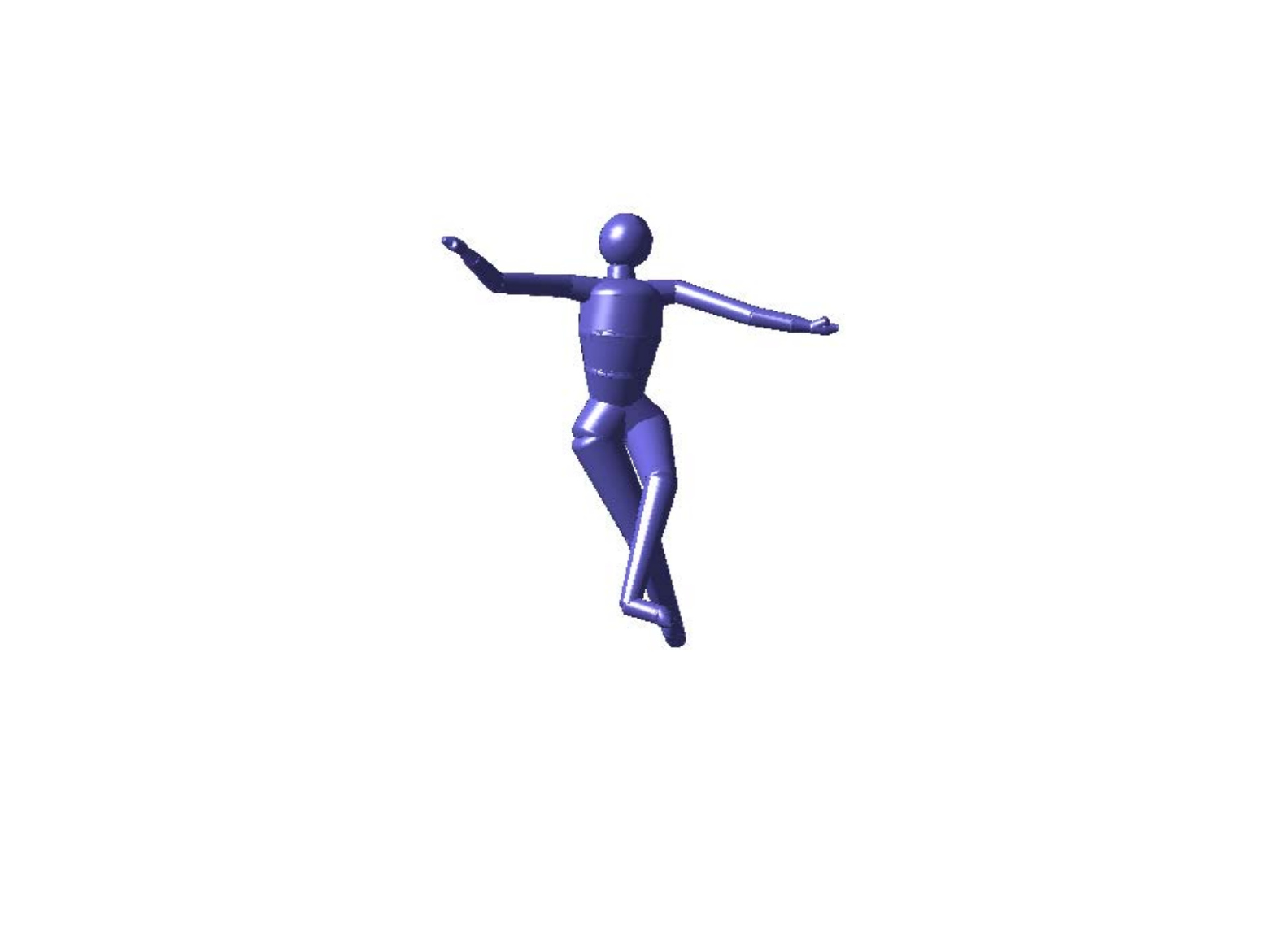} &
\includegraphics[scale=0.2]{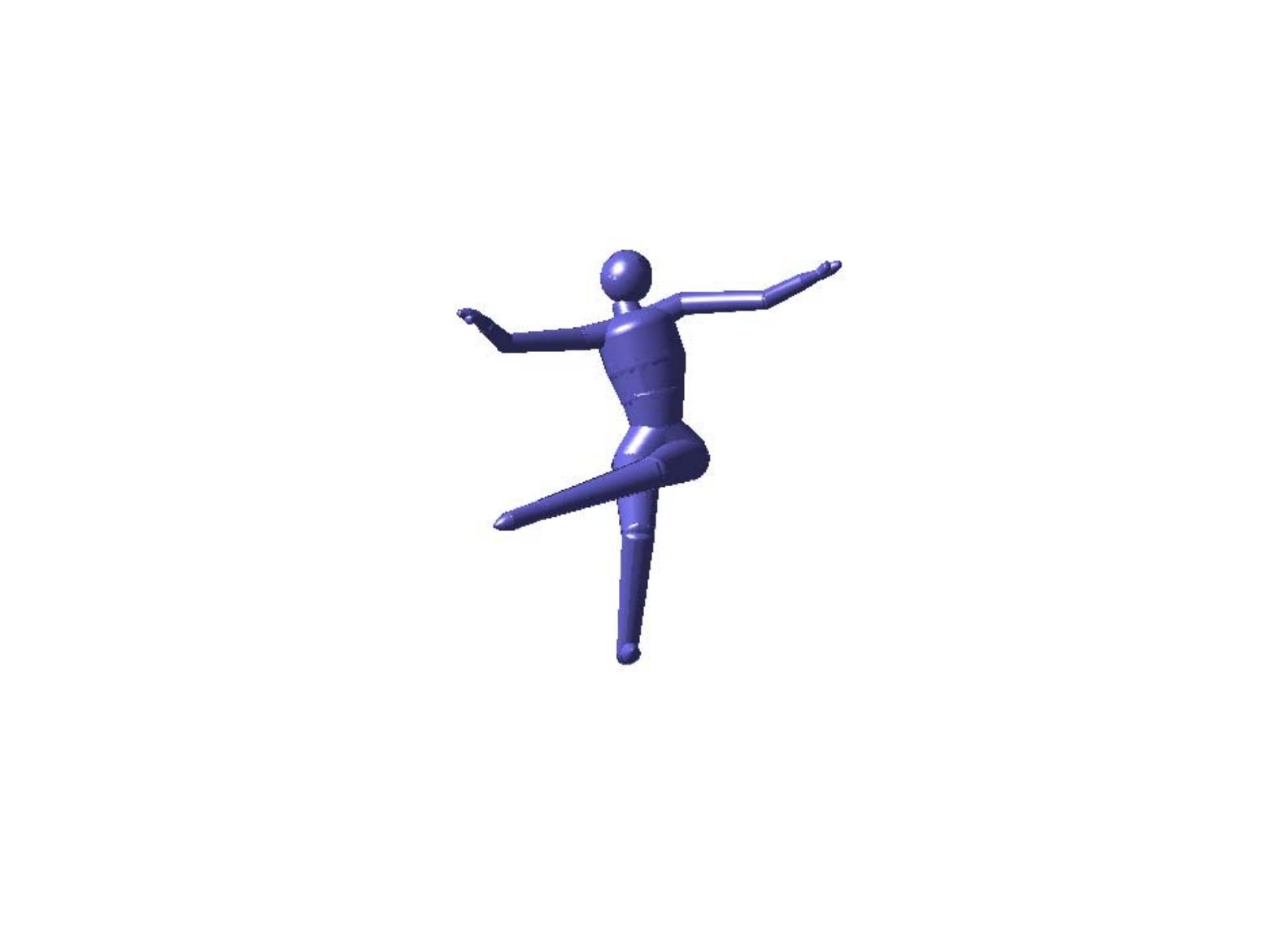} \\
\multicolumn{5}{c}{Poses synthesized from a Nystr{\"o}m-approximated DPP}
\end{tabular}
\end{center}
\caption{Synthesizing perturbed human poses relative to an original pose by sampling (1) i.i.d. from a multivarite Gaussian versus (2) drawing a set from an RFF- or Nystr{\"om}- approximated DPP with kernel based on MoCap data from the activity category.  The Gaussian covariance is likewise formed from the activity data.}
\label{fig:pose}
\end{figure}

Fig.~\ref{fig:pose} displays additional human poses that are drawn i.i.d. from a multivariate Gaussian, and compares to our DPP draws from both the RFF and Nystr{\"om} approximations.

\end{appendix}

\end{document}

%% file: sections/introCR.tex
\presec
\section{Introduction}
\label{sec:intro}
\postsec
Samples from a determinantal point process (DPP)~\cite{macchi1975coincidence} are sets of points that tend to be spread out. % as if repelled by each other.  
More specifically, given $\Omega\subseteq \mathbb{R}^d$ and a positive semidefinite kernel function $L:\Omega\times\Omega\mapsto\mathbb{R}$, the probability density of a point configuration $A \subset \Omega$ under a DPP with kernel $L$ is given by
\begin{equation}
\mathcal{P}_L(A)\propto \det(L_A)~,
\label{eq:DPP}
\end{equation}
where $L_A$ is the $|A|\times |A|$ matrix with entries $L(\bx,\by)$ for each $\bx,\by\in A$.
The tendency for repulsion is captured by the determinant since it depends on the volume spanned by the selected points in the associated Hilbert space of $L$.  Intuitively, points similar according to $L$ or
points that are nearly linearly dependent are less likely to be selected. 
%the simplex re of the probability distribution induces repulsion between points so that point configurations that are more spread out, as specified by the kernel $L$, are deemed to have higher probabilities.

Building on the foundational work in~\cite{borodin2005eynard} for the case where $\Omega$ is discrete and finite, DPPs have been  used in machine learning as a model for subset selection in which diverse sets are preferred~\cite{kulesza2012determinantal,affandi2012markov,Affandi:AISTATS2013,gillenwater2012discovering,kulesza2011k}. These methods build on the tractability of sampling based on the algorithm of Hough et al.~\cite{hough2006determinantal}, which relies on the eigendecomposition of the kernel matrix to recursively sample points based on their projections onto the subspace spanned by the %previously 
selected eigenvectors.

% TODO: Insert hard core ref
Repulsive point processes, like \emph{hard core processes}~\cite{matern1986spatial,daley2003introduction}, many based on thinned Poisson processes and Gibbs/Markov distributions, %\ebf{Insert references and *short* text here...Gibbs, Strauss, etc....maybe mention, maybe not.}, 
have a long history in the spatial statistics community, where considering continuous $\Omega$ is key.  Many naturally occurring phenomena exhibit diversity---trees tend to grow in the least occupied space~\citep{neeff2005}, ant hill locations are  over-dispersed relative to uniform placement~\citep{bernstein1979} and the spatial distribution of nerve fibers is indicative of neuropathy, with hard-core processes providing a critical  tool~\cite{waller2011second}.  Repulsive processes on continuous spaces have garnered interest in machine learning as well, especially relating to generative mixture modeling~\cite{zou2012priors,petralia2012repulsive}. 

The computationally attractive properties of DPPs make them appealing to consider in these applications.  
% However, the simplicity and tractability of DPP sampling in the discrete case does not extend to the continuous case for general kernels.  
On the surface, it seems that the eigendecomposition and projection algorithm of~\cite{hough2006determinantal} for discrete DPPs would naturally extend to the continuous case. % While this is true in a formal sense as $L$ becomes an operator instead of a matrix, the key steps such as the eigendecomposition of the kernel and projection of points on subspaces spanned by eigenfunctions are computationally infeasible except in a few very limited cases where approximations can be made~\cite{lavancier2012statistical}. 
While this is true in a formal sense as $L$ becomes an operator instead of a matrix, the key steps such as the eigendecomposition of the kernel and projection of points on subspaces spanned by \emph{eigenfunctions} are computationally infeasible except in a few very limited cases where approximations can be made~\cite{lavancier2012statistical}. The absence of a tractable DPP sampling algorithm for general kernels in continuous spaces has hindered progress in developing DPP-based models for repulsion.

In this paper, we propose an efficient algorithm to sample from DPPs in continuous spaces using low-rank approximations of the kernel function. We investigate two such schemes: Nystr{\"o}m and random Fourier features. Our approach utilizes a \emph{dual representation} of the DPP, a technique that has proven useful in the discrete $\Omega$ setting as well~\cite{kulesza2010structured}.  For $k$-DPPs, which only place positive probability on sets of cardinality $k$~\cite{kulesza2012determinantal}, we also devise a Gibbs sampler that iteratively samples points in the $k$-set conditioned on all $k-1$ other points.  The derivation relies on representing the conditional DPPs using the Schur complement of the kernel. % [BEN- CUT OR MODIFY THIS SENTENCE?].
%In this paper, we propose two efficient algorithms to sample from DPPs in continuous spaces. The first is based on low-rank approximations of the kernel function using the Nystr{\"o}m method or random Fourier features. The second approach devises a Gibbs sampler for the $k$-DPP that only places positive probability on sets of cardinality $k$~\cite{kulesza2012determinantal}.  
%We show how to compute a \emph{dual representation} of the DPP that has proven useful in the discrete $\Omega$ case as well~\cite{kulesza2010structured}. 
Our methods allow us to handle a broad range of typical kernels and continuous subspaces, provided certain simple integrals of the kernel function can be computed efficiently. Decomposing our kernel into \emph{quality} and \emph{similarity} terms as in~\cite{kulesza2012determinantal}, this includes, but is not limited to, all cases where the (i) spectral density of the quality and (ii) characteristic function of the similarity kernel can be computed efficiently.  %A list of standard cases is provided in the supplement. 
Our methods scale well with dimension, in particular with complexity growing linearly in $d$. %Note that, if desired, the approximations can be corrected for in a Metropolis-Hastings framework, for example.
%We will also show how the the low-rank approximations can be used as an initialization for the Gibbs sampling scheme that empirically improves the rate of convergence.
% For the two rank-$D$ approximation schemes of the kernel function $L$ we propose,  we can efficiently compute a $D\times D$ dual matrix and the $D$ eigenvectors.  
%Based on a Nystr{\"o}m or RFF approximation, we do not rely on knowing the eigendecomposition of $L$ to form this dual, and furthermore show that our sampling algorithm is tractable for a wide range of kernels.
% TODO: More text here
%[ MORE HERE.  THIS ISN'T AN ABSTRACT.  WHAT'S THE KEY INSIGHT?  WHY DOES A KERNEL APPROX LEAD TO AN ABILITY TO SAMPLE? HOW DOES IT WORK?]

In Sec.~\ref{sec:backgroundCR}, we review sampling algorithms for discrete DPPs and the challenges associated with sampling from continuous DPPs. We then propose continuous DPP sampling algorithms based on low-rank kernel approximations in Sec.~\ref{sec:methodCR} and Gibbs sampling in Sec.~\ref{sec:gibbsCR}. % We then propose continuous DPP sampling algorithms based on low-rank kernel approximations in Sec.~\ref{sec:methodCR} and show the generality of the methods. 
%An empirical analysis of the approximate samplers are provided in Sec.~\ref{sec:analysisCR}.
An empirical analysis of the two schemes is provided in Sec.~\ref{sec:analysisCR}. Finally, we apply our methods to repulsive mixture modeling and human pose synthesis in Sec.~\ref{sec:MixModelCR} and \ref{sec:MoCapCR}.%Finally, we apply our methods to repulsive mixture modeling and human pose synthesis in Sec.~\ref{sec:MixModel} and \ref{sec:MoCap}.

%% file: sections/backgroundCR.tex
\presec
\section{Sampling from a DPP}
\label{sec:backgroundCR}
\postsec
When $\Omega$ is discrete with cardinality $N$, an efficient algorithm for sampling from a DPP is given in~\cite{hough2006determinantal}. The algorithm, which is detailed in the supplement, uses an eigendecomposition of the kernel matrix $L = \sum_{n=1}^N\lambda_n v_n v_n^{\top}$ and recursively samples points $\bx_i$ as follows, resulting in a set $A\sim \mbox{DPP}(L)$ with $A=\{\bx_i\}$: %= \{\bx_i\}$ 
\begin{itemize}
\item[Phase 1] Select eigenvector $v_n$ with probability  $\frac{\lambda_n}{\lambda_n +1}$. Let $V$ be the selected eigenvectors ($k=|V|$).
\item[Phase 2] For $i=1,\dots,k$, sample points $\bx_i \in \Omega$ sequentially with probability  based on the projection of  $\bx_i$ onto the subspace spanned by $V$. Once $\bx_i$ is sampled, update $V$ by excluding the subspace spanned by the projection of $\bx_i$ onto $V$. 
\end{itemize}
When $\Omega$ is discrete, both steps are straightforward since the first phase involves eigendecomposing a kernel matrix and the second phase involves sampling from discrete probability distributions based on inner products between points and eigenvectors. Extending this algorithm to a continuous space was considered by~\cite{lavancier2012statistical}, but for a very limited set of kernels $L$ and spaces $\Omega$.  For general $L$ and $\Omega$, we face difficulties in both phases.
%\subsubsection*{Continuous space sampling}
Extending Phase 1 to a continuous space requires knowledge of the eigendecomposition of the kernel function. When $\Omega$ is a compact rectangle in $\mathbb{R}^d$, \cite{lavancier2012statistical} suggest approximating the eigendecomposition using an orthonormal Fourier basis.%When $\Omega$ is a compact rectangle in $\mathbb{R}^d$, \cite{lavancier2012statistical} suggest using an orthonormal Fourier basis for the eigendecomposition. 
%This approach is restricted to kernels $L$ with known spectral density and furthermore excludes kernels that are not translation invariant. 
%\ebf{Cut: Another case in which an explicit eigendecomposition is known is the Gaussian radial basis function (RBF) on $\Omega=\mathbb{R}$: \cite{fasshauer2012stable} derives the exact eigendecomposition with eigenfunctions proportional to Hermite functions, $H_n(\bx)$.}

Even if we are able to obtain the eigendecomposition of the kernel function (either directly or via approximations as considered in \cite{lavancier2012statistical} and Sec.~\ref{sec:methodCR}), we still need to implement Phase 2 of the sampling algorithm.
%Even if we are able to obtain the eigendecomposition of the kernel function (either directly or via approximations as considered in Sec.~\ref{sec:methodCR}), we still need to implement Phase 2 of the sampling algorithm. 
Whereas the discrete case only requires sampling from a discrete probability function, here we have to sample from a probability density. When $\Omega$ is compact, \cite{lavancier2012statistical} suggest using a rejection sampler with a uniform proposal on $\Omega$. The authors note that the acceptance rate of this rejection sampler decreases with the number of points sampled, making the method inefficient in sampling large sets from a DPP. In most other cases, implementing Phase 2 even via rejection sampling is infeasible since the target density is in general non-standard with unknown normalization.
%not in general a standard family of distributions with known normalization.
% and the required bound cannot be computed.  
Furthermore, a generic proposal distribution can yield extremely low acceptance rates.  
%\ebf{Cut: For example, in the Gaussian RBF case, one might imagine using a Gaussian proposal.  While the first Hermite eigenfunction behaves like a Gaussian, for later eigenfunctions, there is significant density away from zero.  In this case, it is not clear if there is a standard proposal distribution that can be used effectively in the rejection sampling algorithm.}

In summary, current algorithms can sample approximately from a continuous DPP only for translation-invariant kernels defined on a compact space. % (a hyper-rectangle).
%In summary, current algorithms can sample from a continuous DPP only for translation-invariant kernels defined on a compact space (most easily defined for a rectangle). 
In Sec.~\ref{sec:methodCR}, we propose a sampling algorithm that allows us to sample approximately
from DPPs for a wide range of kernels $L$ and spaces $\Omega$. %The first approach is based on low-approximations and relies heavily on a dual DPP representation and associated sampler, first introduced by~\cite{kulesza2010structured} for discrete $\Omega$. The second method employs Gibbs sampling scheme that makes use of the Schur complement representation of the DPP.%Our approach relies heavily on a dual DPP representation and associated sampler, first introduced by~\cite{kulesza2010structured} for discrete $\Omega$.  We review this approach below.

%% file: sections/methodCR.tex
\presec
\section{Sampling from a low-rank continuous DPP}
\label{sec:methodCR}
\postsec
%\subsubsection*{Dual DPP Sampling}
Again considering $\Omega$ discrete with cardinality $N$, the sampling algorithm of Sec.~\ref{sec:backgroundCR} has complexity dominated by the eigendecomposition, $O(N^3)$. If the kernel matrix $L$ is low-rank, i.e. $L=B^\top B,$ with $B$ a $D\times N$ matrix and $D \ll N$, \cite{kulesza2010structured} showed that the complexity of sampling can be reduced to $O(ND^2+D^3)$. The basic idea is to exploit the fact that $L$ and the \emph{dual kernel matrix} $C=BB^\top$, which is $D\times D$, share the same nonzero eigenvalues, and for each eigenvector $v_k$ of $L$, $B v_k$ is the corresponding eigenvector of $C$. See the supplement for algorithmic details.  

While the dependence on $N$ in the dual is sharply reduced, in continuous spaces, $N$ is infinite. In order to extend the algorithm, we must find efficient ways to compute $C$ for Phase 1 and manipulate eigenfunctions implicitly for the projections in Phase 2.
%, which is provided in the supplement, takes time $O(D^3 + ND^2)$ and space $O(ND+D^2)$.
Generically, consider sampling from a DPP on a continuous space $\Omega$ with kernel 
%\begin{align}
%	\vspace{-0.05in}
$L(\bx,\by)=\sum_{n=1}^{\infty}\lambda_n\phi_n(\bx)\overline{\phi_n}(\by), $%~~~~~~~~~~(\bx,\by)\in \Omega\times\Omega~,$
%\end{align}
where $\lambda_n$ and $\phi_n(\bx)$ are eigenvalues and eigenfunctions,  and $\overline{\phi_n}(\by)$ is the complex conjugate of $\phi_n(\by)$. 
Assume that we can approximate $L$ by a low-dimensional (generally complex-valued) mapping, $B(\bx) : \Omega \mapsto \mathbb{C}^D$:
\begin{align}
\Lapp(\bx,\by)=B(\bx)^*B(\by)~, \text{where}~ B(\bx)=[B_1(\bx),\ldots,B_D(\bx)]^{\top}.
\label{eq:approxL}
\end{align}
Here, $A^*$ denotes complex conjugate transpose of $A$. We consider two efficient low-rank approximation schemes in Sec.~\ref{sec:RFF} and \ref{sec:Nystrom}. 
Using such a low-rank representation, we propose an analog of the dual sampling algorithm for continuous spaces, described in Algorithm \ref{alg:dualCont}. A similar algorithm provides samples from a \emph{k-DPP}, which only gives positive probability to sets of a fixed cardinality $k$~\cite{kulesza2012determinantal}.  The only change required is to the for-loop in Phase 1 to select exactly $k$ eigenvectors using an efficient $O(Dk)$ recursion. %\ebf{insert short description here}.  
See the supplement for details.
\begin{algorithm}[tbh]
 \caption{Dual sampler for a low-rank continuous DPP}
  \label{alg:dualCont}
	\begin{minipage}[l]{2.6in}
		\begin{algorithmic}
 \STATE {\bfseries Input:} $\Lapp(\bx,\by)=B(\bx)^*B(\by)$,\\
  \STATE \hspace{0.4in} a rank-$D$ DPP kernel \\  
\STATE{\bfseries PHASE 1}
  \STATE Compute $C=\int_{\Omega}B(\bx)B(\bx)^* d\bx$  
  \STATE Compute eigendecomp. $C = \sum_{k=1}^D\lambda_k \bv_k \bv_k^*$\\ %$\{(\bv_k,\lambda_k)\}_{k=1}^D$ of $C$\\
  \STATE $J\leftarrow\emptyset$\\
  \FOR{$k=1,\ldots,D$}
  \STATE $J \leftarrow J\cup\{k\}$ with probability $\frac{\lambda_k}{\lambda_k+1}$\\
  \ENDFOR
  \STATE $V\leftarrow\{\frac{v_k}{\sqrt{v_k^{*}C v_k}}\}_{k\in J}$
	\end{algorithmic}
\end{minipage}
\begin{minipage}[r]{2.9in}
		\begin{algorithmic}
\STATE{\bfseries PHASE 2}
  \STATE $X\leftarrow\emptyset$\\
  \WHILE{$|V|>0$}
  \STATE Sample $\hat{\bx}$ from  $f(\bx)=\frac{1}{|V|}\sum_{\bv\in V}|\bv^*B(\bx)|^2$ \\
  \STATE $X\leftarrow X\cup\{\hat{\bx}\}$\\
  \STATE Let $\bv_0$ be a vector in $V$ such that $\bv_0^*B(\hat{\bx})\neq 0$\\
  \STATE Update $V\leftarrow\{\bv-\frac{\bv^*B(\hat{\bx})}{\bv_0^*B(\hat{\bx})}\bv_0~|~v\in V-\{v_0\} \}$\\
  \STATE Orthonormalize $V$ w.r.t. $\langle \bv_1,\bv_2\rangle=\bv_1^{*}C\bv_2$\\
  \ENDWHILE
  \STATE  {\bfseries Output:} $X$
\end{algorithmic}
\end{minipage}
\end{algorithm}

%Once we have this dual matrix, we can eigendecompose it and perform sampling in this dual representation, analogously to the dual DPP sampler of~\cite{kulesza2012determinantal} adapted to continuous $\Omega$.  See Algorithm \ref{alg:dualCont}.   

%Note that the non-zero eigenvalues for $\Lapp$ coincides with the eigenvalues of  $C$ and the corresponding eigenvectors,$\{\bv_i\}_{i=1}^D$ are related to the eigenfunctions,$\{\phi_i(\bx)\}_{i=1}^D$ of $\Lapp$ by  
%\begin{equation}
%\phi_i(\bx)=\frac{1}{\lambda_i}B(\bx)^*\bv_i~.
%\end{equation}

%Note that the approximation gives a huge advantage in that we do not need any knowledege of the original eigenvalues and eigenfunctions to derive this approximation. The only requirement we need at this point is knowing how to compute the integral that results in the dual matrix.    

In this dual view, we still have the same two-phase structure, and must address two key challenges:  
\vspace{-0.05in}
\begin{itemize}
	\item[Phase 1] Assuming a low-rank kernel function decomposition as in Eq.~\eqref{eq:approxL}, we need to able to compute the dual kernel matrix, 
	%$C \in \mathbb{R}^{D \times D}$, for continuous spaces is 
	given by an integral:
\begin{align}
C=\int_{\Omega}B(\bx)B(\bx)^* d\bx~. 
\label{eq:C}
\end{align} 
%	We show how we can do this efficiently for our proposed low-rank approximations.
	\item[Phase 2] In general, sampling directly from the density $f(\bx)$ is difficult; %(this corresponds to sampling proportional to the projection in the discrete case). 
	instead, we can compute the cumulative distribution function (CDF) and sample $\bx$ using the inverse CDF method~\cite{robert2004monte}:
\begin{align}
F(\hat{\bx}=(\hat{x}_1,\ldots,\hat{x}_d))=\prod_{l=1}^d\int_{-\infty}^{\hat{x}_l} f(\bx)1_{\{x_l\in\Omega\}}dx_l.~
\label{eq:F}
\end{align}
%and sample $\bx$ using the inverse CDF method~\cite{robert2004monte}. %\ebf{Insert }. %In particular, we randomly generate $u\sim$Unif[0,1] and use the bisection method to find the root of $F(\bx)-u$.
%We show the feasibility of this step as well. 
\end{itemize}
\vspace{-0.05in}

Assuming (i) the kernel function $\Lapp$ is finite-rank and (ii) the terms $C$ and $f(\bx)$ are computable, Algorithm~\ref{alg:dualCont} provides exact samples from a DPP with kernel $\Lapp$.  In what follows, approximations only arise from approximating general kernels $L$ with low-rank kernels $\Lapp$.  If given a finite-rank kernel $L$ to begin with, the sampling procedure is exact.

One could imagine approximating $L$ as in Eq.~\eqref{eq:approxL} by simply truncating the eigendecomposition (either directly or using numerical approximations).  However, this simple approximation for known decompositions does not necessarily yield a tractable sampler, because the products of eigenfunctions required in Eq.~\eqref{eq:C} might not be efficiently integrable. % as needed for Eq.~\eqref{eq:C}.  
%For example, in the Gaussian RBF case, the integral of Eq.~\eqref{eq:C} requires integration of terms $H_n(\bx)^2$, the squared Hermite function \ebf{need more if remove previous}, which do not have a standard form. 
For our approximation algorithm to work, not only do we need methods that approximate the kernel function well, but also that enable us to solve Eq.~\eqref{eq:C} and \eqref{eq:F} directly for many different kernel functions. We consider two such approaches that enable an efficient sampler for a wide range of kernels: Nystr{\"o}m and random Fourier features.  

%sampling from $f(\bx)$ using the inverse CDF method doesn't seem like a better alternative than sampling from $p_i(\bx)$ in Algorithm \ref{alg:DPPCont}. However, we will show that using RFF or Nystr{\"o}m yields a more manageable integral than that of $\int_{-\infty}^{x^*}\phi_k(\bx)\phi_j(\bx)d\bx$ required in Algorithm \ref{alg:DPPCont}. In fact, the only steps in Algorithm \ref{alg:dualCont} that may present some difficulties are computing $C$ and $F(x)$ and we will show that RFF and Nystr{\"o}m  method allows us to do this for many common kernels.

\pressec
\subsection{ Sampling from RFF-approximated DPP}
\label{sec:RFF}
\postssec
Random Fourier features (RFF) \citep{rahimi2007random} is an approach for approximating shift-invariant kernels, $k(\bx,\by)=k(\bx-\by)$, using  randomly selected frequencies. %, as an inner product between pairs of points in $\Omega$. %This is done by first sampling random features independently from the Fourier transform of the kernel function, $\bomega_1,\ldots,\bomega_D\sim\mathcal{F}(k(x-y))$ and letting %\bz(\bx)=\frac{1}{\sqrt{D}}\bigg[\exp\{i\bomega_1^{\top}\bx\},\ldots,\exp\{i\bomega_D^{\top}\bx\}\bigg]^{\top}$. 
%In particular, it introduced a randomized feature map, $\bz:\Omega\rightarrow\mathbb{R}^D$ so that
%\begin{equation}
%\tilde{k}(\bx-\by)=\bz(\bx)^{\top}\bz(\by)~,~~~~\bx,\by\in\Omega~.
%\end{equation} 
%
%This is done by first sampling random features
The frequencies are sampled independently from the Fourier transform of the kernel function, $\bomega_j\sim\mathcal{F}(k(\bx-\by))$,   and letting: %$\bz(\bx)=\frac{1}{\sqrt{D}}\bigg[\exp\{i\bomega_1^{\top}\bx\},\ldots,\exp\{i\bomega_D^{\top}\bx\}\bigg]^{\top}$. 
%
%The approximated kernel is then
\begin{align}
\tilde{k}(\bx-\by)=\frac{1}{D}\sum_{j=1}^D\exp\{i\bomega_j^{\top}(\bx-\by)\}~,~~~~\bx,\by\in\Omega~.
\label{eq:RFF}
\end{align} 

To apply RFFs, we factor $L$ into a quality function $q$ and similarity kernel $k$ (i.e., $q(\bx)=\sqrt{L(\bx,\bx)})$: 
\begin{align}
 L(\bx,\by)=q(\bx)k(\bx,\by)q(\by)~,~~~~~~~\bx,\by\in \Omega~~\text{where}~ k(\bx,\bx)=1.
\end{align}

The RFF approximation can be applied to cases where the similarity function % is shift-invariant %: $k(\bx,\by)=k(\bx-\by)$% with 
has a known characteristic function, e.g.,\ Gaussian, Laplacian and Cauchy. Using Eq.~\eqref{eq:RFF}, we can approximate the similarity kernel function to obtain a low-rank kernel and dual matrix:
\begin{align*}
\LRFF(\bx,\by)=\frac{1}{D}\sum_{j=1}^Dq(\bx)\exp\{i\bomega_j^{\top}(\bx-\by)\}q(\by),~\CRFF_{jk}=\frac{1}{D}\int_{\Omega}q^2(\bx)\exp\{i(\bomega_j-\bomega_k)^{\top}\bx\}d\bx.
\end{align*}
%
%The corresponding elements of dual matrix is given by
%\begin{equation}
%\CRFF_{jk}(\bx,\by)=\frac{1}{D}\int_{\Omega}q(\bx)\exp\{i(\bomega_j-\bomega_k)^{\top}\bx\}d\bx
%\label{eq:CRFF}
%\end{equation}

The CDF of the sampling distribution $f(\bx)$ in Algorithm \ref{alg:dualCont} is given by:
\begin{align}
\FRFF(\hat{\bx})=\frac{1}{|V|}\sum_{\bv\in V}\sum_{j=1}^D\sum_{k=1}^D v_jv_k^{*}\prod_{l=1}^d\int_{-\infty}^{\hat{x}_l}q^2(\bx)\exp\{i(\bomega_j-\bomega_k)^{\top}\bx\}1_{\{x_l\in\Omega\}}dx_l.
\label{eq:FRFF}
\end{align}
where $v_j$ denotes the $j$th element of vector $\bv$. Note that equations $\CRFF$ and $\FRFF$ can be computed for many different combinations of $\Omega$ and $q(\bx)$. In fact, this method works for any combination of (i) translation-invariant similarity kernel $k$ with known characteristic function and (ii) quality function $q$ with known spectral density. %For example, if $\Omega$ is a hypercube $\mathcal{A}^d$ in $\mathbb{R}^d$ and we assume uniform quality, $q(\bx)=1$ both equations are easy to evaluate. Furthermore, in cases where $\Omega=\mathbb{R}^d$ and $q(\bx)$ is either Gaussian, Laplacian or Cauchy, both equations can also be solved in terms of error functions, exponentials and exponential integrals respectively. 
The resulting kernel $L$ need not be translation invariant.
In the supplement, we illustrate this method by considering a common and important example where $\Omega=\mathbb{R}^d$, $q(\bx)$ is Gaussian, and $k(\bx,\by)$ is any kernel with known Fourier transform.% We will illustrate this method by considering a common and important example where $\Omega=\mathbb{R}^d$ and $q(\bx)$ is Gaussian. Further examples are provided in the supplement.

\pressec
\subsection{Sampling from a Nystr{\"o}m-approximated DPP}
\label{sec:Nystrom}
\postssec
Another approach to kernel approximation is the Nystr{\"o}m method~\cite{williams2001using}. 
%While this method has lately been used in many applications where a low-rank approximation to high-dimensional kernel matrix is desired, the method was originally developed to elicit approximate eigenvalues and eigenfunctions from kernel functions~\cite{williams2001using}. In particular, given a kernel $L(\bx,\by)=\sum_{n=1}^{\infty}\lambda_n\phi_n(\bx)\overline{\phi_n}(\by)$, the Nystr{\"o}m method solves
%\begin{align}
%\frac{1}{D}\sum_{k=1}^DL(\by,\bz_k)\tilde{\phi}_n(\bz_k)=\tilde{\lambda}_n\tilde{\phi}_n(\by)~,
%\end{align}
%where
In particular, given  $\bz_1,\ldots,\bz_D$ \emph{landmarks} sampled from $\Omega$, we can approximate the kernel function and dual matrix as, 
%\begin{equation}
%\tilde{L}(\bx,\by)=B(\bx)^{\top}B(\by)
%\end{equation}
%where $B(\bx)=\frac{1}{\sqrt{D}}\bigg[\sum_{j=1}L(\bz_j,\bz_1)^{-1/2}L(\bx,\bz_j),\ldots,\sum_{j=1}L(\bz_j,\bz_D)^{-1/2}L(\bx,\bz_j)]^{\top}$
%
%We can also perform Nystr{\"o}m approximation to the kernel to perform DPP sampling similar to the RFF-approximation. Using equation \eqref{eq:Nys}, we can approximate the the kernel with
\begin{align*}
\LNys(\bx,\by)=\sum_{j=1}^D\sum_{k=1}^DW_{jk}^2L(\bx,\bz_j)L(\bz_k,\by),~~\CNys_{jk} =\sum_{n=1}^D\sum_{m=1}^DW_{jn}W_{mk}\int_\Omega L(\bz_n,\bx)L(\bx,\bz_m)d\bx,
\end{align*}
where $W_{jk}=L(\bz_j,\bz_k)^{-1/2}$. %Then the elements of dual matrix $\CNys$ is given by
%\begin{equation}
%
%\label{eq:CNys}
%\end{equation}
%
Denoting $\bw_j(\bv)=\sum_{n=1}^DW_{jn}v_n$, the CDF of $f(\bx)$ in Alg. \ref{alg:dualCont} is:
\begin{align}
\FNys(\hat{\bx})=\frac{1}{|V|}\sum_{\bv\in V}\sum_{j=1}^D\sum_{k=1}^D \bw_j(\bv)\bw_k(\bv)\prod_{l=1}^d\int_{-\infty}^{\hat{x}_l}L(\bx,\bz_j)L(\bz_k,\bx)1_{\{x_l\in\Omega\}}dx_l.
\label{eq:FNys}
\end{align}
%Again, we consider factoring the kernel function as products of similarity and quality functions:
%\begin{equation}
% L(\bx,\by)=q(\bx)k(\bx,\by)q(\by)~,~~~~~~~\bx,\by\in \Omega
%\end{equation}

%Note that some combinations of $\Omega$, $k(\bx,\by)$ and $q(\bx)$ of which equations \ref{eq:CRFF} and \ref{eq:FRFF} can readily solve, can no longer be easily solved for equations \ref{eq:CNys} and \ref{eq:FNys}. In the case where $\Omega=\mathcal{A}^d$, and $q(\bx)=1$, equations \ref{eq:CNys} and \ref{eq:FNys} can still be computed for Gaussian, Laplace and Cauchy similarity kernels. When $\Omega=\mathbb{R}^d$, dealing with Cauchy kernels will be hard while combinations of Gaussian and Laplace qualities and similarities are still feasible. 
As with the RFF case, we consider a decomposition $L(\bx,\by) = q(\bx)k(\bx,\by)q(\by)$.  Here, there are no translation-invariant requirements, even for the similarity kernel $k$.  In the supplement, we provide the important example where $\Omega=\mathbb{R}^d$ and both $q(\bx)$ and $k(\bx,\by)$ are Gaussians and also when $k(\bx,\by)$ is polynomial, a case that cannot be handled by RFF since it is not translationally invariant. %property of the kernel.%In the supplement, we provide the important example where $\Omega=\mathbb{R}^d$ and both $q(\bx)$ and $k(\bx,\by)$ are Gaussians.

%\begin{figure}
%	\begin{center}%
%		\begin{tabular}
%		[c]{ccc}%
%		\includegraphics[scale=0.3]{figs/2DPlotRandom4} &
%		\includegraphics[scale=0.3]{figs/2DPlotNystrom4} &
%		\includegraphics[scale=0.3]{figs/2DPlotRFF4}\\
%		{\small (a)} & {\small (b)} & {\small (c)}
%		\end{tabular}
%	\end{center}
%	\vspace{-0.15in}
%\caption{\small \ebf{Remake figure and merge with figure in analysis section} (a) Random samples from a 2-D multivariate Gaussian.  For a DPP with non-diagonal Gaussian quality and similarity covariances $\Sigma$ and $\Gamma$,  draws from a (b) Nystr{\"o}m-approximated DPP and (c) RFF-approximated DPP.}
%\label{fig:samples}
%\end{figure}

%% file: sections/GibbsCR.tex
\presec
\section{Gibbs sampling}
\label{sec:gibbsCR}
\postsec
For $k$-DPPs, we can consider a Gibbs sampling scheme. In the supplement, we derive that the full conditional for the inclusion of point $\bx_k$ given the inclusion of the $k-1$ other points is a $1$-DPP with a modified kernel, which we know how to sample from. Let the kernel function be represented as before: $L(\bx,\by) = q(\bx)k(\bx,\by)q(\by)$. Denoting $J^{\backslash k}=\{\bx_j\}_{j\neq k}$ and $M^{\backslash k}=L_{J^{\backslash k}}^{-1}$ the full conditional can be simplified using Schur's determinantal equality \cite{schur1917potenzreihen}:
\begin{align}
%p(\bx_k|{\{\bx_j\}}_{j\neq k})\propto q(\bx_k)^2(1-\sum_{i,j\neq k}M^k_{ij}q(\bx_i)q(\bx_j)k(\bx_k,\bx_i)k(\bx_j,\bx_k)).
p(\bx_k|{\{\bx_j\}}_{j\neq k})\propto L(\bx_k,\bx_k)-\sum_{i,j\neq k}M^{\backslash k}_{ij}L(\bx_i,\bx_k)L(\bx_j,\bx_k).
\label{eq:fullCond}
\end{align}  
In general, sampling directly from this full conditional is difficult. However, for a wide range of kernel functions, including those which can be handled by the Nystr{\"o}m approximation in Sec.~\ref{sec:Nystrom}, the CDF can be computed analytically and $\bx_k$ can be sampled using the inverse CDF method:
\begin{align}
%F(\hat{\bx}_|{\{\bx_j\}}_{j\neq k})= \frac{\int_{-\infty}^{\hat{\bx}_l}q(\bx_k)^2(1-\sum_{i,j\neq k}M_{ij}q(\bx_i)q(\bx_j)k(\bx_k,\bx_i)k(\bx_j,\bx_k))1_{\{\bx_l\in \Omega\}}d\bx_l}{\int_\Omega q(\bx_i)^2(1-\sum_{i,j\neq k}M_{ij}q(\bx_i)q(\bx_j)k(\bx_k,\bx_i)k(\bx_j,\bx_k))d\bx}
F(\hat{\bx}_l|{\{\bx_j\}}_{j\neq k})= \frac{\int_{-\infty}^{\hat{\bx}_l}L(\bx_l,\bx_l)-\sum_{i,j\neq k}M^{\backslash k}_{ij}L(\bx_i,\bx_l)L(\bx_j,\bx_l)1_{\{\bx_l\in \Omega\}}d\bx_l}{\int_\Omega L(\bx,\bx)-\sum_{i,j\neq k}M^{\backslash k}_{ij}L(\bx_i,\bx)L(\bx_j,\bx)d\bx}
\label{eq:fullCondCDF}
\end{align}  
In the supplement, we illustrate this method by considering the case where $\Omega=\mathbb{R}^d$ and $q(\bx)$ and $k(\bx,\by)$ are Gaussians. We use this same Schur complement scheme for sampling from the full conditionals in the mixture model application of Sec.~\ref{sec:MixModelCR}. A key advantage of this scheme for several types of kernels is that the complexity of sampling scales linearly with the number of dimensions $d$ making it suitable in handling high-dimensional spaces.

%[Ben: Theoretical justification?]

As with any Gibbs sampling scheme, the mixing rate is dependent on the correlations between variables. In cases where the kernel introduces low repulsion we expect the Gibbs sampler to mix well, while in a high repulsion setting the sampler can mix slowly due to the strong dependencies between points and fact that we are only doing one-point-at-a-time moves.  We explore the dependence of convergence on repulsion strength in the supplementary materials. Regardless, this sampler provides a nice tool in the $k$-DPP setting.  Asymptotically, theory suggests that we get \emph{exact} (though correlated) samples from the $k$-DPP.  To extend this approach to standard DPPs, we can first sample $k$ (this assumes knowledge of the eigenvalues of $L$) and then apply the above method to get a sample. This is fairly inefficient if many samples are needed.  A more involved but potentially efficient approach is to consider a birth-death sampling scheme where the size of the set can grow/shrink by 1 at every step. 
%Fig.~\ref{fig:L1} ((a) and (b)) illustrate the convergence of the unnormalized probability and the second moment of Gibbs samples from a 1-D DPP with Gaussian quality and similarity kernels. While both quantities seem to converge rapidly in the low repulsion case, bad starting points can result in a convergence to a local mode in the high repulsion case. Thus an approximated-DPP samples obtained using methods in Sec.~\ref{sec:methodCR} can be used as initialization to ensure a fast mixing of the Gibbs samples.

%% file: sections/analysisCR.tex
\presec
\section{Empirical analysis}
\label{sec:analysisCR}
\postsec
To evaluate the performance of the RFF and Nystr{\"o}m approximations, we compute the total variational distance
%
%\begin{equation}
$\|\mathcal{P}_L-\mathcal{P}_{\tilde{L}}\|_1=\frac{1}{2}\sum_{X}|\mathcal{P}_L(X)-\mathcal{P}_{\tilde{L}}(X)|$,
%\end{equation}
%
where $\mathcal{P}_L(X)$ denotes the probability of set $X$ under a DPP with kernel $L$, as given by Eq.~\eqref{eq:DPP}. We restrict our analysis to the case where the quality function and similarity kernel are Gaussians with isotropic covariances $\Gamma=\diag(\rho^2,\ldots,\rho^2)$ and $\Sigma=\diag(\sigma^2,\ldots,\sigma^2)$, respectively, enabling our analysis based on the easily computed eigenvalues~\cite{fasshauer2012stable}. We also focus on sampling from $k$-DPPs for which the size of the set $X$ is always $k$. Details are in the supplement.% $\Gamma$ and $\Sigma$, respectively% One can show that the normalized density is $\mathcal{P}_L(X) =\frac{\det(L_X)}{\prod_{n=1}^\infty(1+\lambda_n(L))}$, which requires the eigenvalues of the kernel $L$. As such, for the experiments of this section, we restrict our analysis to the case where the quality function and similarity kernel are Gaussians with covariances $\Gamma$ and $\Sigma$, respectively. For convenience, we let $\Gamma=\diag(\rho^2,\ldots,\rho^2)$ and $\Sigma=\diag(\sigma^2,\ldots,\sigma^2)$. In this case, letting $n=(n_1,\dots,n_d)$ with $n_j \in \mathbb{Z}_+$, the eigenvalues (indexed by $n$) are~\cite{fasshauer2012stable}:%\ben{need to define $n_j$ below (is n a multindex?)}:
%\begin{equation}
%\phi_n(x)=\left(\frac{\rho^2}{\pi}\right)^{\frac{d}{4}}\exp\left\{-\frac{\|x\|^2}{2\rho^2}\right\}\prod_{j=1}^d\sqrt{\frac{\beta}{2^{n-1}\Gamma(n)}}\exp\bigg(-\frac{(\beta^2-1)x_j^2}{2\rho^2}\bigg)H_{n_j-1}\left(frac{\beta x_j}{\sqrt{\rho^2}}\right)
%\end{equation}
%and 
%\begin{equation}
%\lambda_n=\prod_{j=1}^d \sqrt{\frac{2\pi}{\rho^2(\beta^2+1)+4(\frac{\rho^4}{\sigma^2})}}\bigg(\frac{4}{(\frac{\sigma^2}{\rho^2})(\beta^2+1)+4}\bigg)^{n_j-1}~,
%\lambda_n=\prod_{j=1}^d \sqrt{\frac{2\pi}{\rho^2(c_1+c_2)}}\bigg(\frac{c_2}{c_1+c_2}\bigg)^{n_j-1} \hspace{0.25in} c_1 = (\beta^2+1) \hspace{0.25in} c_2 = 4\frac{\rho^2}{\sigma^2}~.
%\end{equation}
%where $\beta=(1+\frac{2\rho^2}{\sigma^2})^{\frac{1}{4}}$ and $H_n$ denotes the $n$th Hermite polyomial.
%Our analyses also focus on sampling from $k$-DPPs for which the size of the set $X$ is always $k$.

\begin{figure}
	\begin{center}
		\begin{tabular}{cccc}
			\includegraphics[scale=0.18]{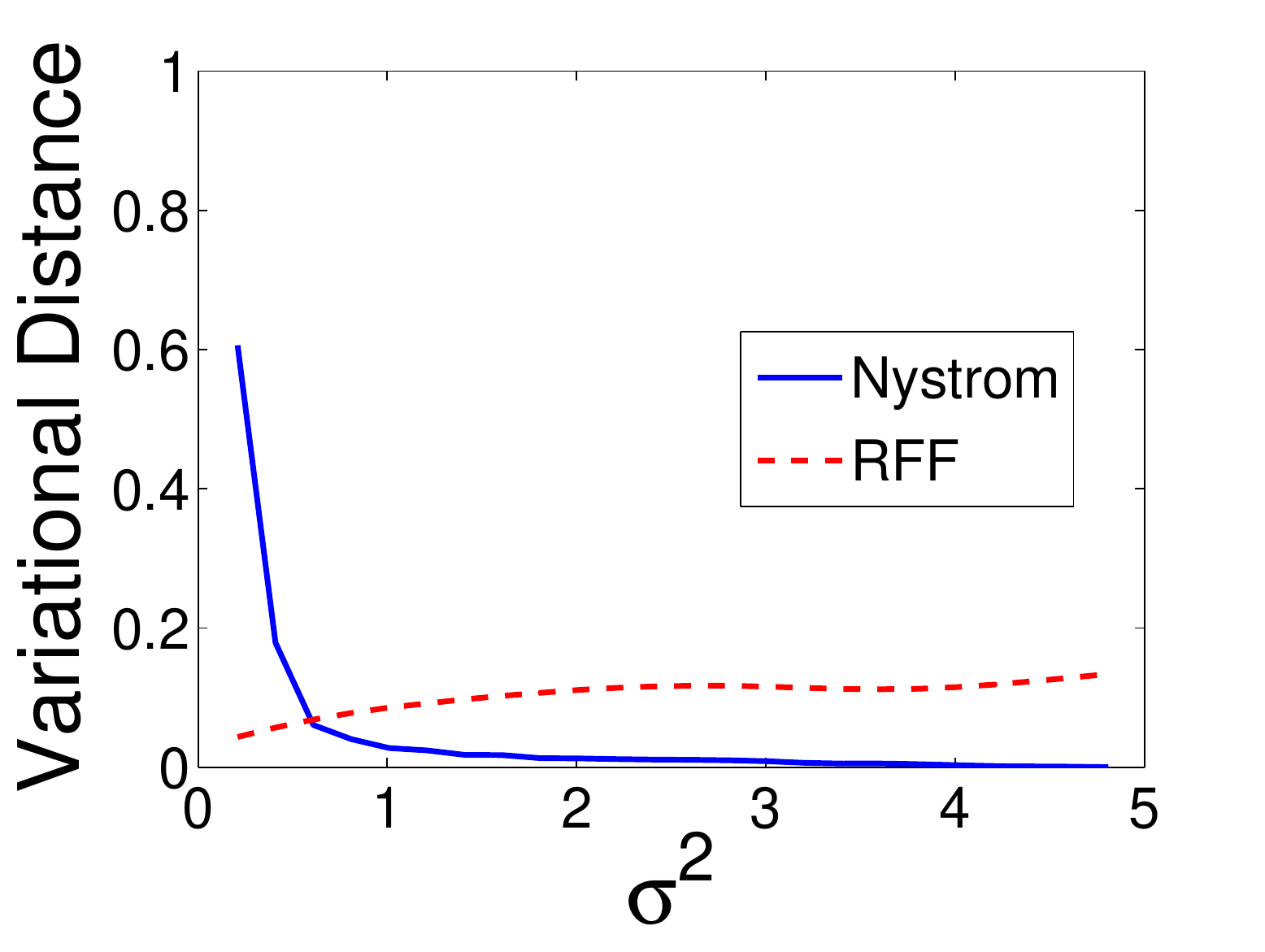} &
			\includegraphics[scale=0.18]{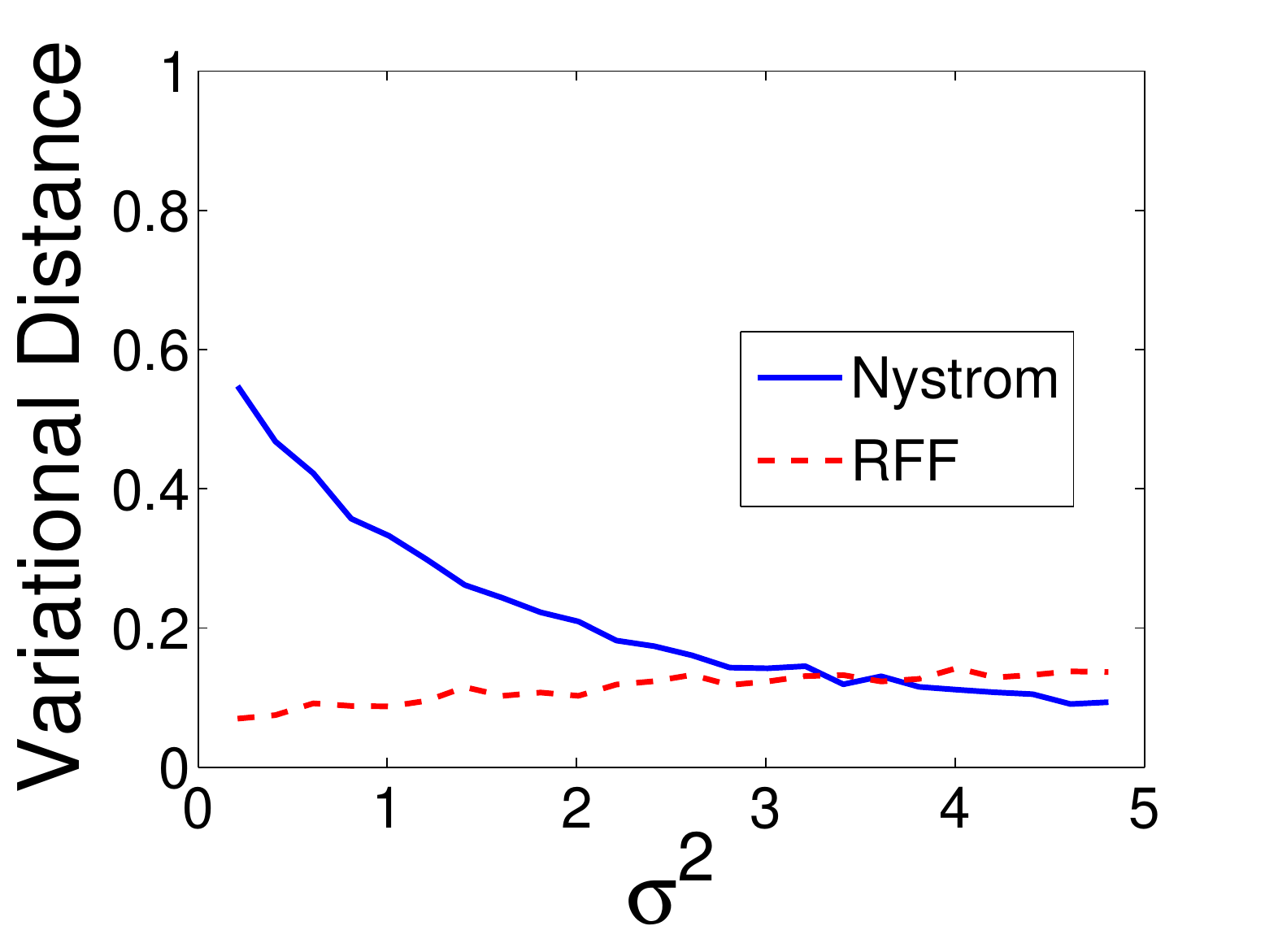} &
			\includegraphics[scale=0.18]{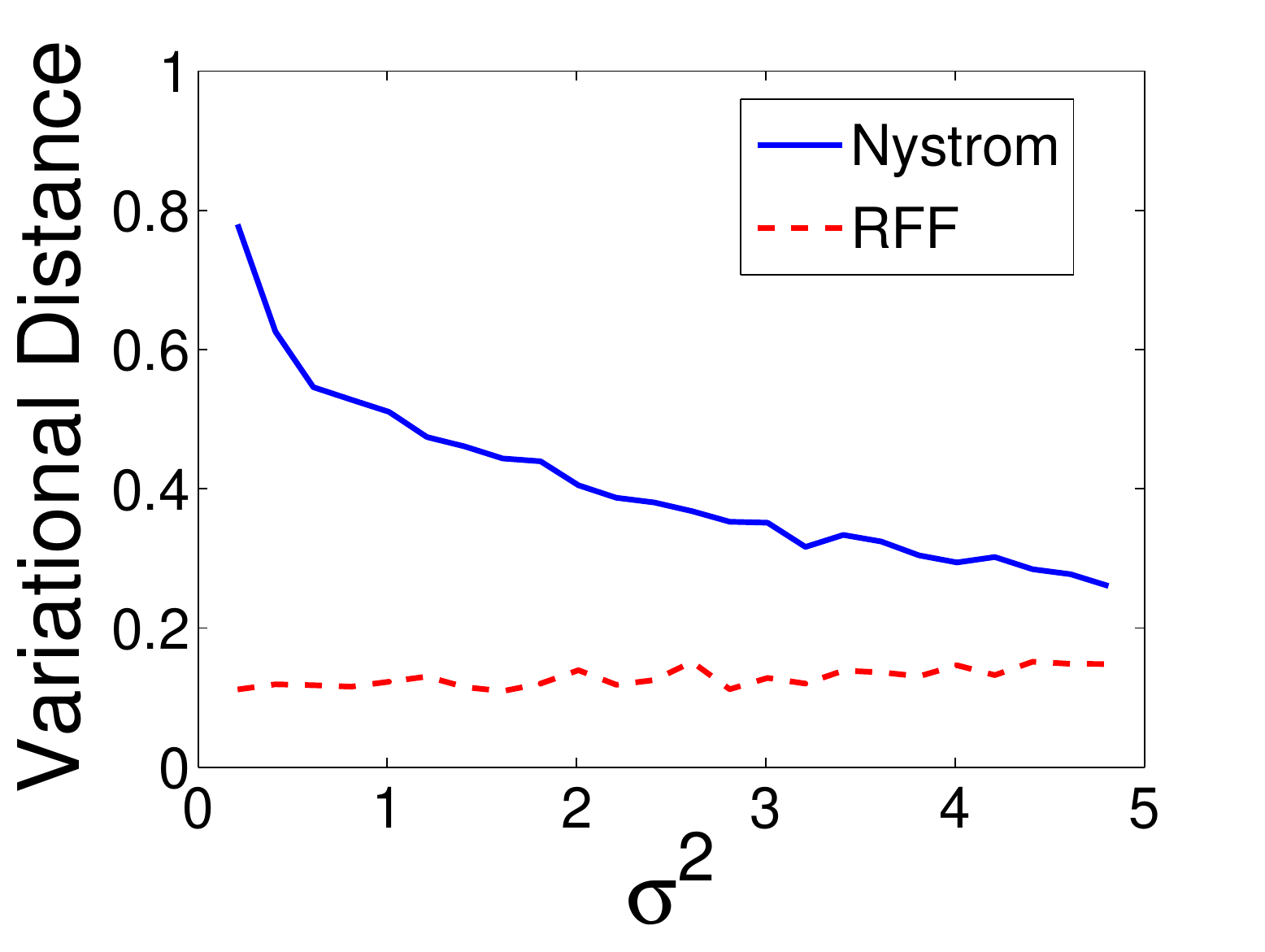}  &
			\includegraphics[scale=0.18]{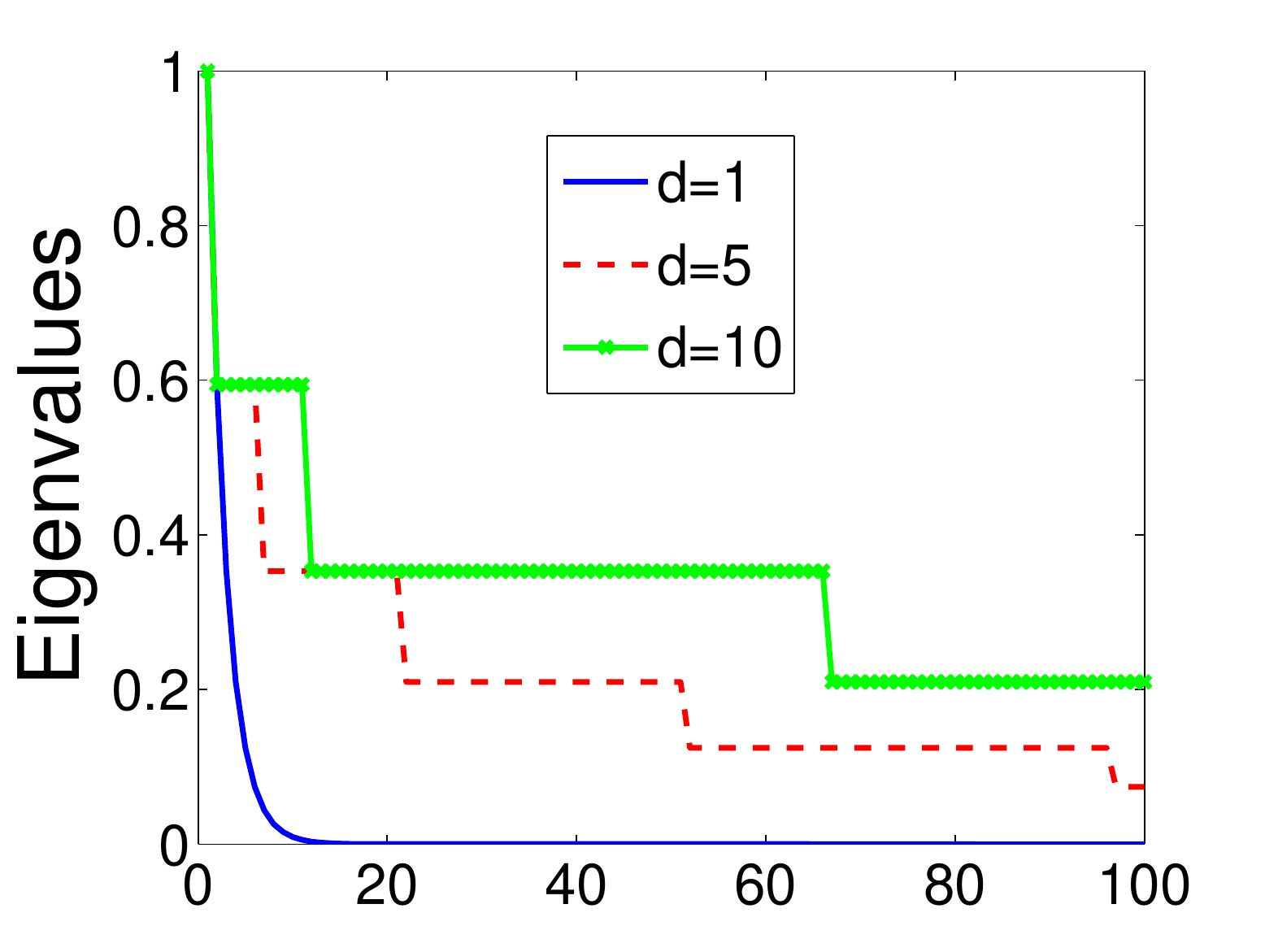} 
                                \vspace{-0.05in}\\
			{\small (a)} & {\small (b)} & {\small (c)} &{\small (d)}
		\end{tabular}
			\precap
\caption{\small Estimates of total variational distance for Nystr{\"o}m and RFF approximation methods to a DPP with Gaussian quality and similarity with covariances $\Gamma=\diag(\rho^2,\ldots,\rho^2)$ and $\Sigma=\diag(\sigma^2,\ldots,\sigma^2)$, respectively. (a)-(c) For dimensions $d$=1, 5 and 10, each plot considers $\rho^2=1$ and varies $\sigma^2$. (d) Eigenvalues for the Gaussian kernels with $\sigma^2=\rho^2=1$ and varying dimension $d$.}
%\caption{\small
% Log of unnormalized probability of the Gibbs samples in (a) low repulsion kernel and (b) high repulsion kernel. %\small 
%Estimates of total variational distance for Nystr{\"o}m and RFF approximation methods to a DPP with Gaussian quality and similarity with covariances $\Gamma=\diag(\rho^2,\ldots,\rho^2)$ and $\Sigma=\diag(\sigma^2,\ldots,\sigma^2)$, respectively. (c)-(e) For dimensions $d$=1, 5 and 10, each plot considers $\rho^2=1$ and varies $\sigma^2$.}% (d) Eigenvalues for the Gaussian kernels with $\sigma^2=\rho^2=1$ and varying dimension $d$.}

\label{fig:L1}
\postcap
	\end{center}
\end{figure}

Fig.~\ref{fig:L1} displays estimates of the total variational distance for the RFF and Nystr{\"o}m approximations when $\rho^2=1$, varying $\sigma^2$ (the repulsion strength) and the dimension $d$. Note that the RFF method performs slightly worse as $\sigma^2$ increases and is rather invariant to $d$ while the Nystr{\"o}m method performs much better for increasing $\sigma^2$ but worse for increasing $d$. % Furthermore, the Nystr{\"o}m approximation appears to perform progressively worse with increasing $d$ while the RFF method seems to perform comparably well across dimensions. 

While this phenomenon seems perplexing at first, a study of the eigenvalues of the Gaussian kernel across dimensions sheds light on the rationale (see Fig.~\ref{fig:L1}). 
Note that for fixed $\sigma^2$ and $\rho^2$, the decay of eigenvalues is slower in higher dimensions. It has been previously demonstrated that the Nystr{\"o}m method performs favorably in kernel learning tasks compared to RFF in cases where there is a large eigengap in the kernel matrix~\cite{yang2012nystrom}. The plot of the eigenvalues seems to indicate the same phenomenon here. Furthermore, this result is consistent with the comparison of RFF to Nystr{\"om} in approximating DPPs in the discrete $\Omega$ case provided in~\cite{Affandi:AISTATS2013}.

This behavior can also be explained by looking at the theory behind these two approximations. %For the RFF, Rahimi and Recht~\cite{rahimi2007random} proved that $\Exp[\tilde{k}(\bx-\by)]=k(\bx-\by)$ with
%
%\begin{equation}
%\mathcal{P}\left(\sup_{\bx,\by\in \Omega}|\tilde{k}(\bx-\by)-k(\bx-\by)|\ge\epsilon\right)\le C_{\Omega,\epsilon}\exp\left\{\frac{-D\epsilon^2}{4(d+2)}\right\}~,
%\label{eq:RFFbound}
%\end{equation}
%
% TODO: discuss bias wording in this paragraph
%for some constant $C_{\Omega,\epsilon}$ that depends on $\Omega$ and $\epsilon$. 
For the RFF, while the kernel approximation is guaranteed to be an unbiased estimate of the true kernel element-wise, the variance is fairly high~\cite{rahimi2007random}. %in Eq.~\eqref{eq:RFFbound} 
In our case, we note that the RFF estimates of minors are biased because of non-linearity in matrix entries, overestimating probabilities for point configurations that are more spread out, %This effect can be seen in the RFF-approximated DPP sample of Fig.~\ref{fig:L1}(a), %\ebf{correct fig ref} 
%where we see that the sampled points tend to be overly dispersed.
which leads to samples that are overly-dispersed.
% TODO: what does "other points" mean here?  Also, large sigma^2 --> small sigma^2
For the Nystr{\"o}m method, on the other hand, the quality of the approximation depends on how well the landmarks cover $\Omega$. In our experiments the landmarks are sampled i.i.d. from $q(\bx)$.  
%\ebf{Do we need more here on landmark sampling?  Randomly sampled how?  From Gaussian?} 
When either the similarity bandwidth $\sigma^2$ is small or the dimension $d$ is high, the effective distance between points increases, thereby decreasing the accuracy of the approximation. Theoretical bounds for the Nystr{\"om} DPP approximation in the case when $\Omega$ is finite are provided in~\cite{Affandi:AISTATS2013}. We believe the same result holds for continuous $\Omega$ by extending the eigenvalues and spectral norm of the kernel matrix to operator eigenvalues and operator norms, respectively. 

In summary, for moderate values of $\sigma^2$ it is generally good to use the Nystr{\"o}m approximation for low-dimensional settings and RFF for high-dimensional settings. %Alternatively, we can use a convex combination of the two approximation methods to achieve a compromise between them, specifically by considering $\rho \CRFF + (1-\rho)\CNys$. Sampling can be performed as efficiently as for each individually.  Fig.~\ref{fig:L1} shows that this convex combination between the RFF and Nystr{\"o}m approximations that minimizes variational distance does better than either method individually.  %as determined by cross validation, does at least as well as the two methods separately.  
%Therefore, this convex combination provides a nice default procedure for approximating DPPs in the continuous $\Omega$ case
%$\ebf{need to explain what it means to have a convex combo, because there are other things you can do, in particular that mixed C and mixed F can be computed efficiently at least when everything is Gaussian. also, i don't see the figure with convex combo}.

%% file: sections/applicationsCR.tex
\presec
%\section{Applications}
%\postsec
\section{Repulsive priors for mixture models}
\label{sec:MixModelCR}
\postsec
Mixture models are used in a wide range of applications from clustering to density estimation. A common issue with such models, especially in density estimation tasks, is the introduction of redundant, overlapping components that increase the complexity and reduce interpretability of the resulting model. This phenomenon is especially prominent when the number of samples is small. In a Bayesian setting, a common fix to this problem is to consider a sparse Dirichlet prior on the mixture weights, which penalizes the addition of non-zero-weight components.  However, such approaches run the risk of inaccuracies in the parameter estimates~\cite{petralia2012repulsive}. Instead,~\cite{petralia2012repulsive} show that sampling the location parameters using repulsive priors leads to better separated clusters while maintaining the accuracy of the density estimate. They propose a class of repulsive priors that rely on explicitly defining a distance metric and the manner in which small distances are penalized.  The resulting posterior computations can be fairly complex.

The theoretical properties of DPPs make them an appealing choice as a repulsive prior. In fact,~\cite{zou2012priors} considered using DPPs as repulsive priors in latent variable models. However, in the absence of a feasible continuous DPP sampling algorithm, their method was restricted to performing MAP inference. Here we propose a fully generative probabilistic mixture model using a DPP prior for the location parameters, with a $K$-component model using a $K$-DPP.

In the common case of mixtures of Gaussians (MoG), our posterior computations can be performed using Gibbs sampling with nearly the same simplicity of the standard case where the location parameters $\mu_k$ are assumed to be i.i.d..  In particular, with the exception of updating the location parameters $\{\mu_1,\ldots,\mu_K\}$, our sampling steps are identical to standard MoG Gibbs updates in the uncollapsed setting. For the location parameters, instead of sampling each $\mu_k$ independently from its conditional posterior, our full conditional depends upon the other locations $\mu_{\backslash k}$ as well. Details are in the supplement, where we show that this full conditional has an interpretation as a single draw from a tilted $1$-DPP.  As such, we can employ the Gibbs sampling scheme of Sec.~\ref{sec:gibbsCR}.  %Since we are just sampling one item from the DPP, there is a closed form and we do not rely on approximations.  
%sample each location parameter $\mu_j$ from a DPP using the likelihood of $\mu_j$ as the quality function, $q(\mu_j)$ and a Gaussian kernel for the similarity function, conditional on the other location parameters. Thus, conditional on the other centroids, the sampling distribution for $\mu_j$ will be augmented such that it will be repelled away from them (see Figure \ref{fig:CondDist})   This algorithm is detailed in the supplementary material.
%An impressive aspect of our DPP formulation is that it maintains a Gibbs sampling strategy with nearly the same simplicity of the standard i.i.d. sampler.

We assess the clustering and density estimation performance of the DPP-based model on both synthetic and real datasets. In each case, we run 10,000 Gibbs iterations, discard 5,000 as burn-in and thin the chain by 10.  Hyperparameter settings are in the supplement. We randomly permute the labels in each iteration to ensure balanced label switching. Draws are post-processed following the algorithm of~\cite{stephens2000dealing} to address the label switching issue. 

\begin{figure}
	\vspace*{-0.05in}
	\begin{center}
		\begin{tabular}{ccccc}
			%\hspace{-0.15in}\includegraphics[scale=0.2]{figs/synthwellhist} & \hspace{-0.05in}
			%\includegraphics[scale=0.2]{figs/synthpoorhist} & \hspace{-0.05in}
			%\multirow{2}{*}{\includegraphics[scale=0.2]{figs/galaxyhist}} & \hspace{-0.05in}
			%\multirow{2}{*}{\includegraphics[scale=0.2]{figs/enzymehist}} &\hspace{-0.05in}
			%\multirow{2}{*}{\includegraphics[scale=0.2]{figs/acidityhist}}\vspace{-0.02in}\\
			%\hspace{-0.15in}\includegraphics[scale=0.2]{figs/synthwellact} & \hspace{-0.05in}
			%\includegraphics[scale=0.2]{figs/synthpooract} &  & & \vspace{-0.02in}\\
			\hspace{-0.15in}\includegraphics[scale=0.15]{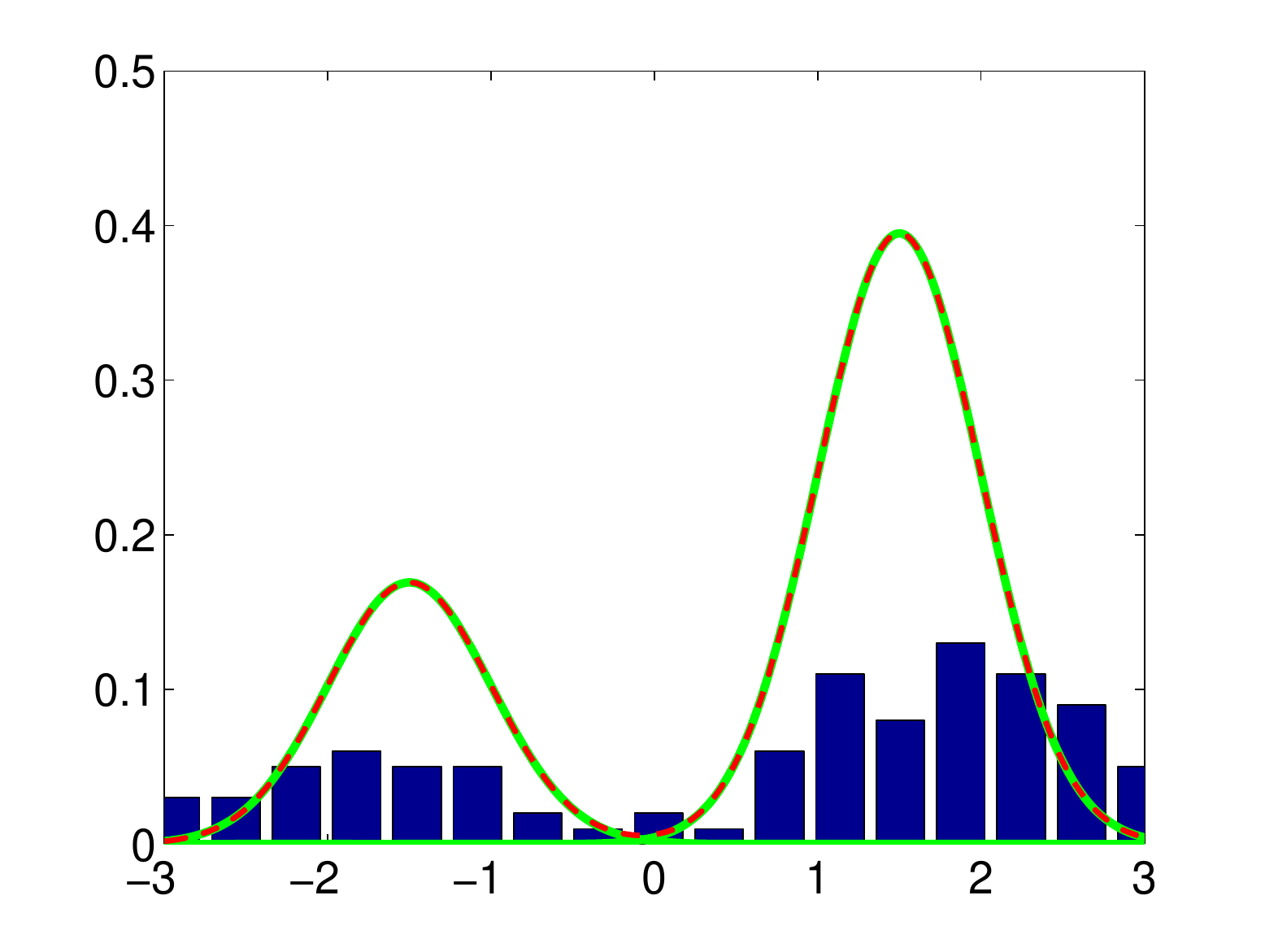} & \hspace{-0.05in} 
			\includegraphics[scale=0.15]{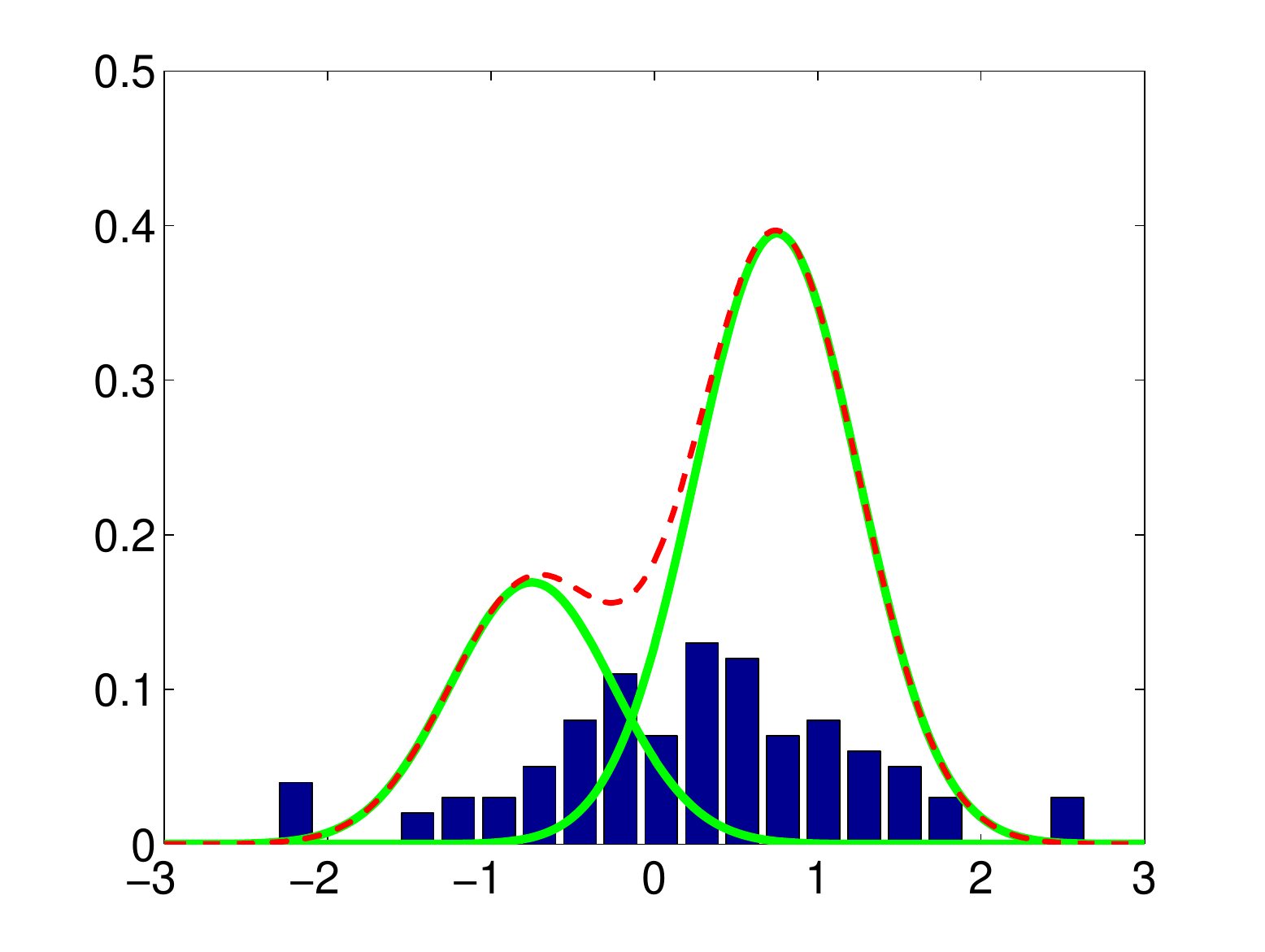} & \hspace{-0.08in}
			\includegraphics[scale=0.15]{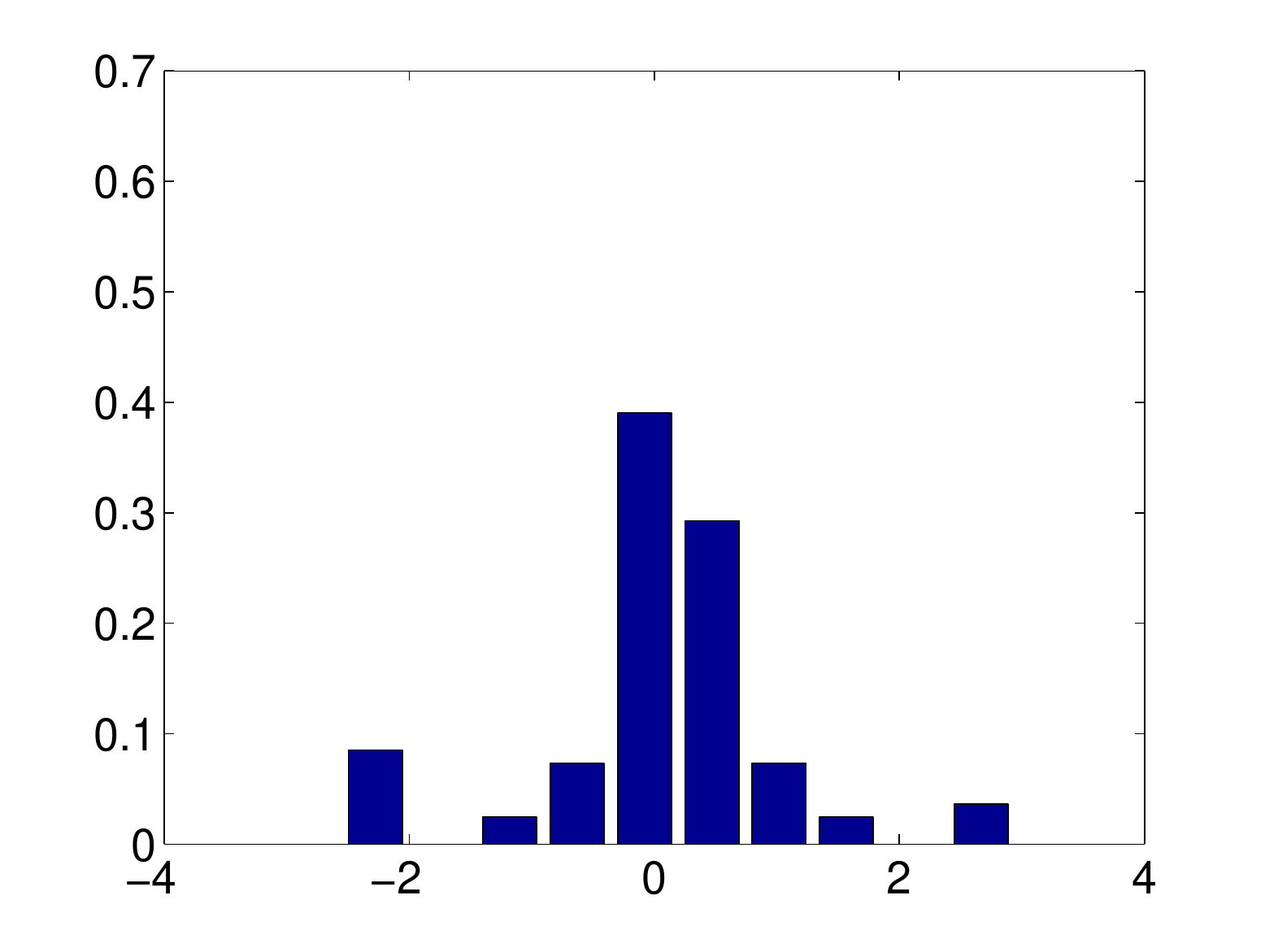} & \hspace{-0.08in}
			\includegraphics[scale=0.15]{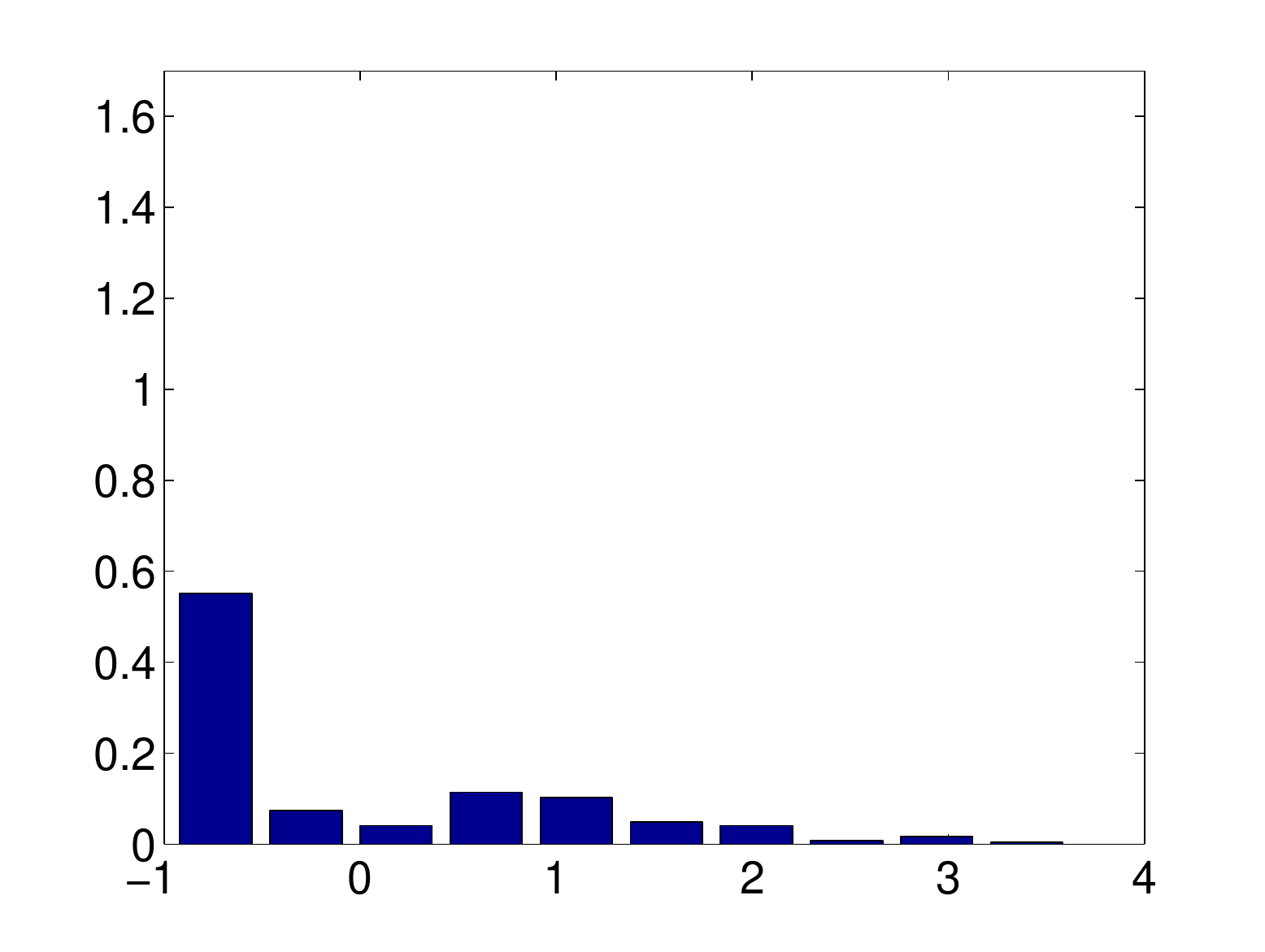} &\hspace{-0.08in}
			\includegraphics[scale=0.15]{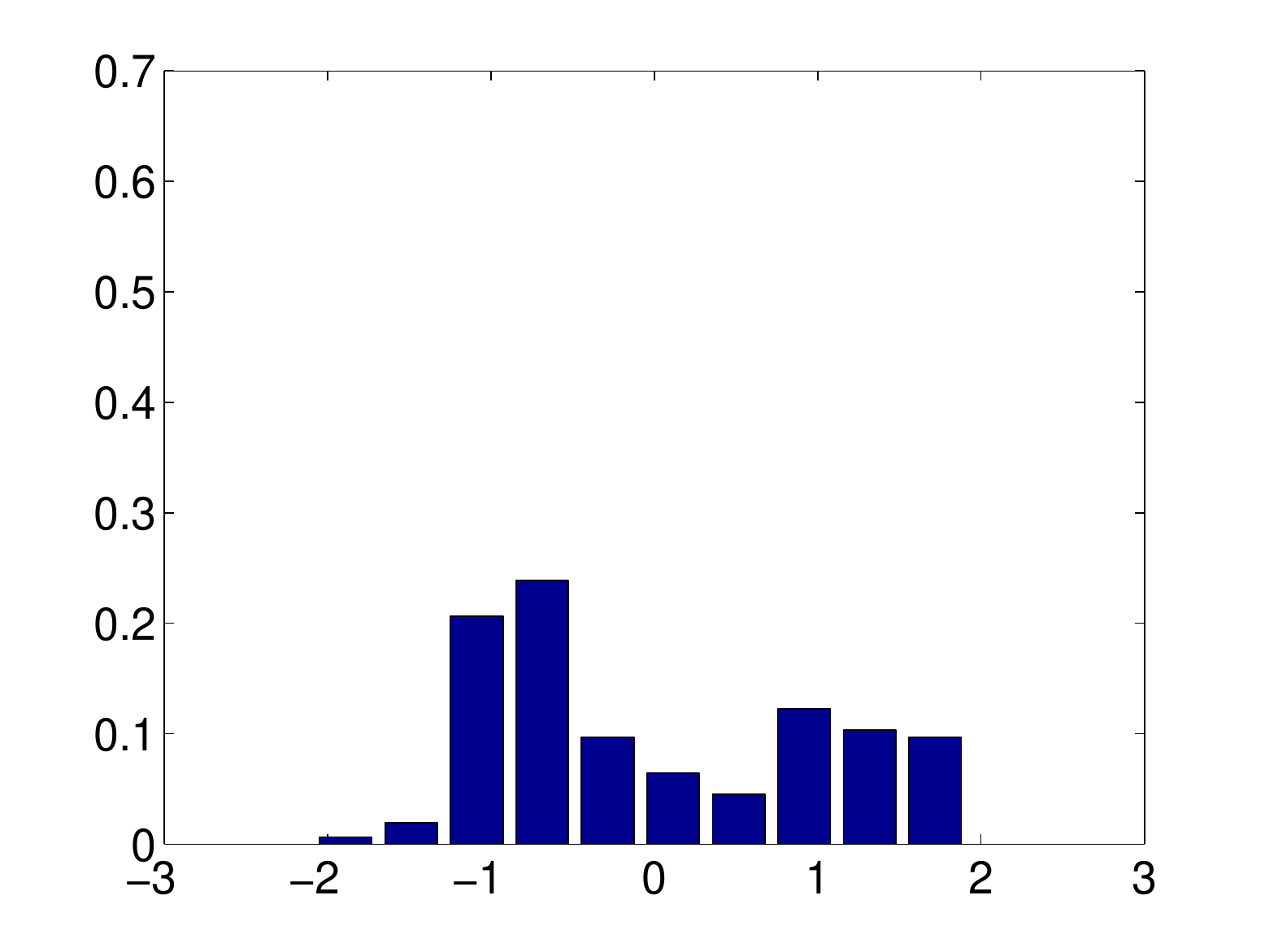}\vspace{-0.02in}\\
			\hspace{-0.15in}\includegraphics[scale=0.15]{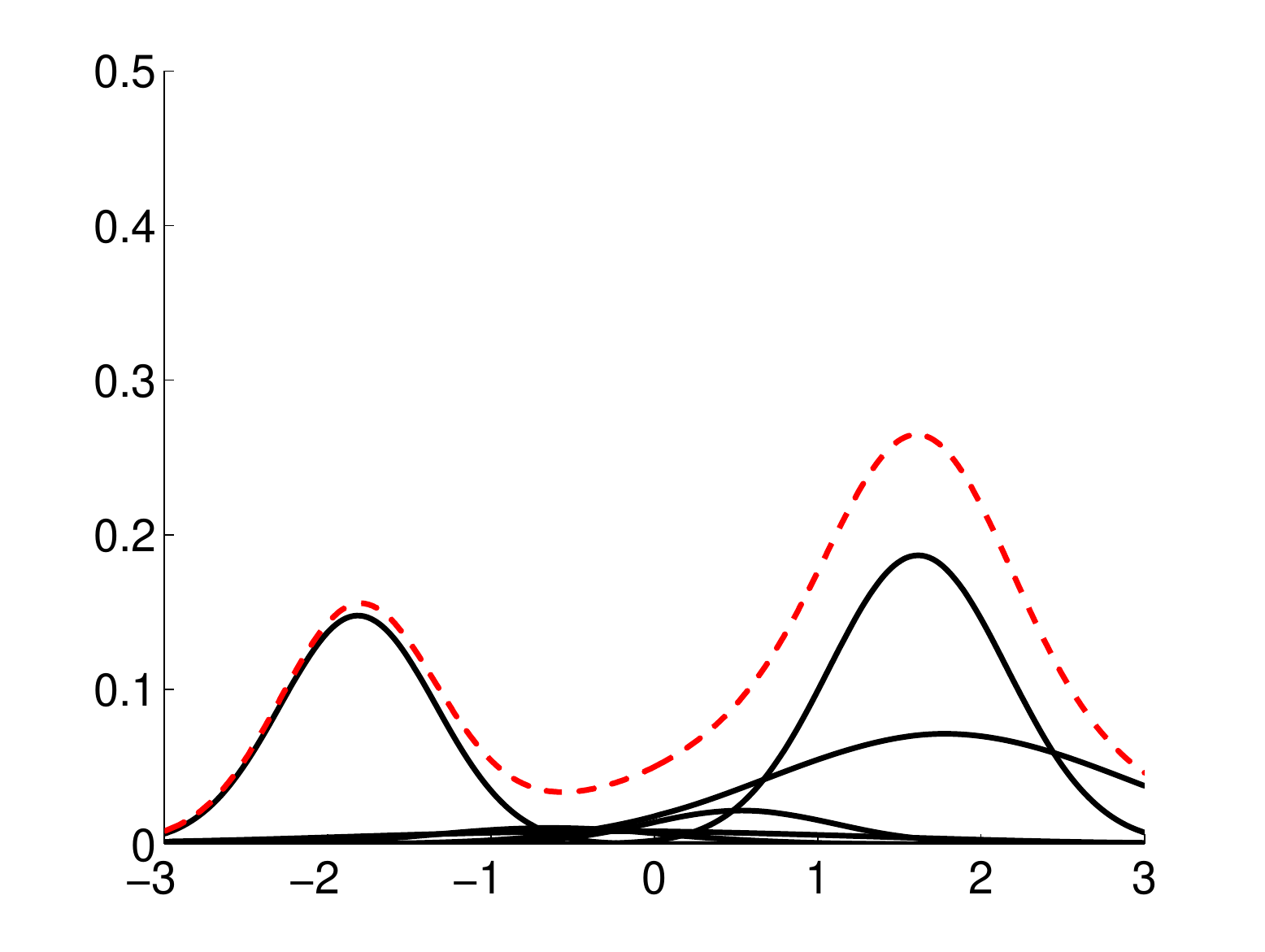} &\hspace{-0.08in}
			\includegraphics[scale=0.15]{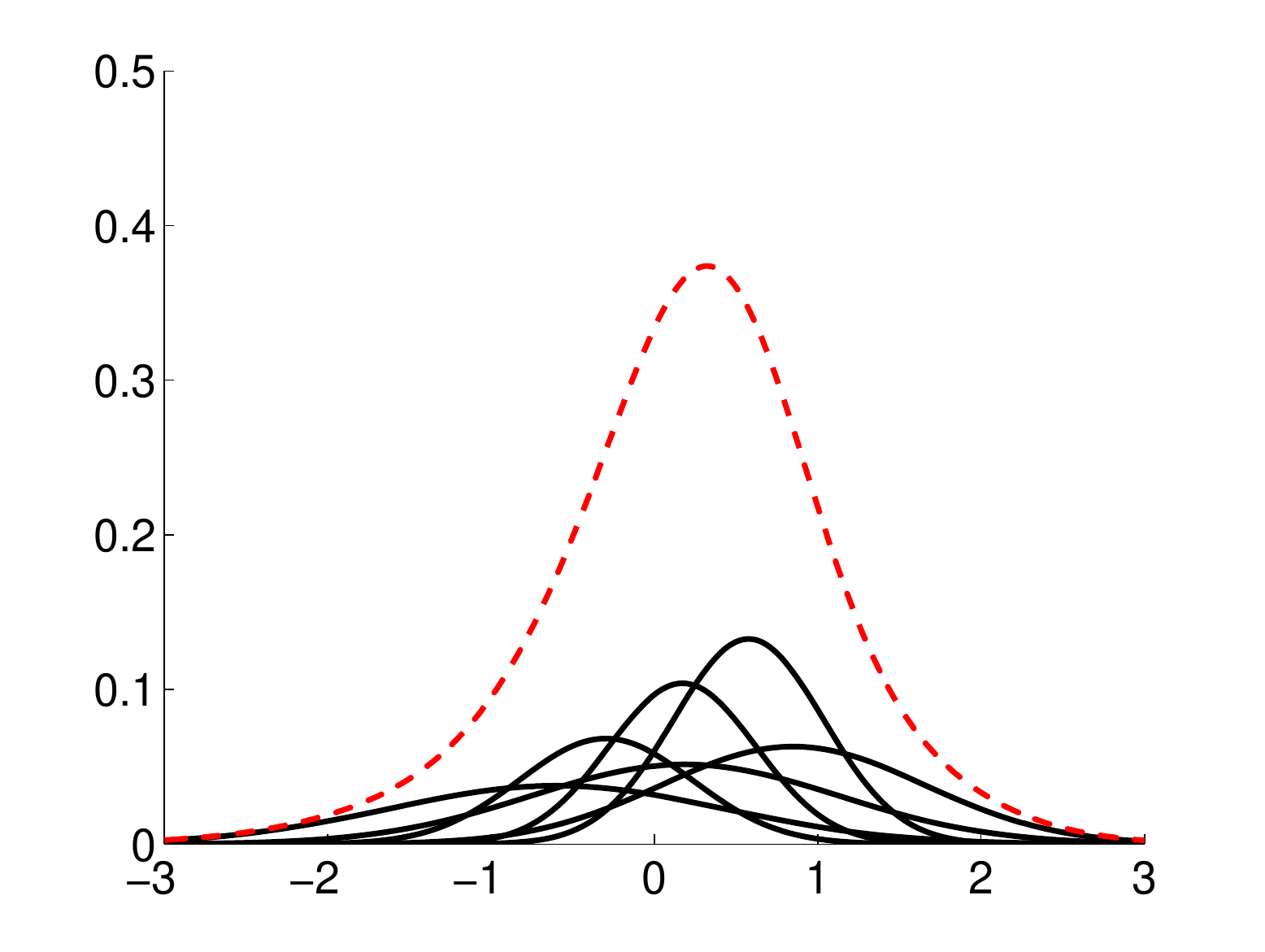} &\hspace{-0.08in}
			\includegraphics[scale=0.15]{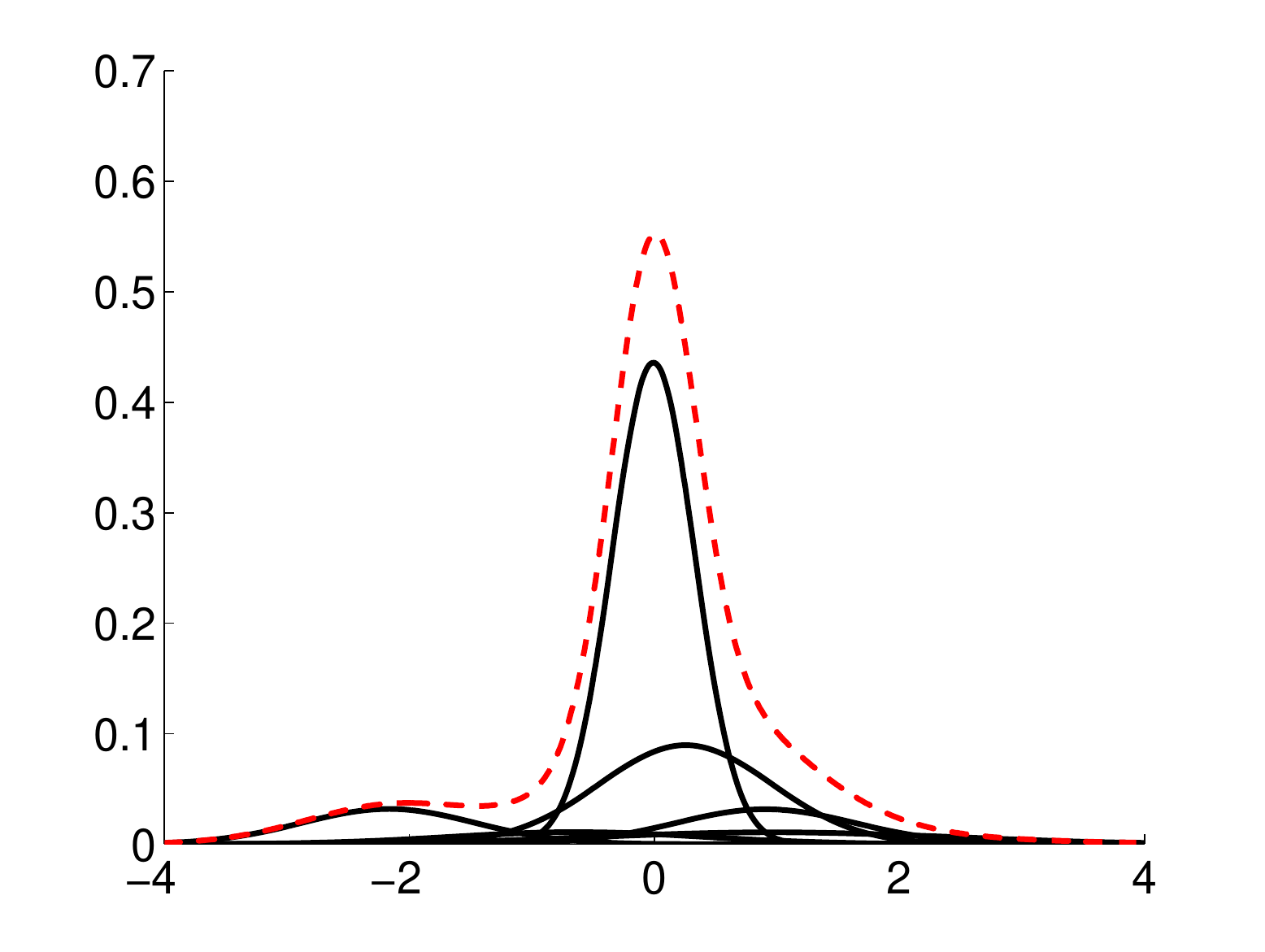} &\hspace{-0.08in}
			\includegraphics[scale=0.15]{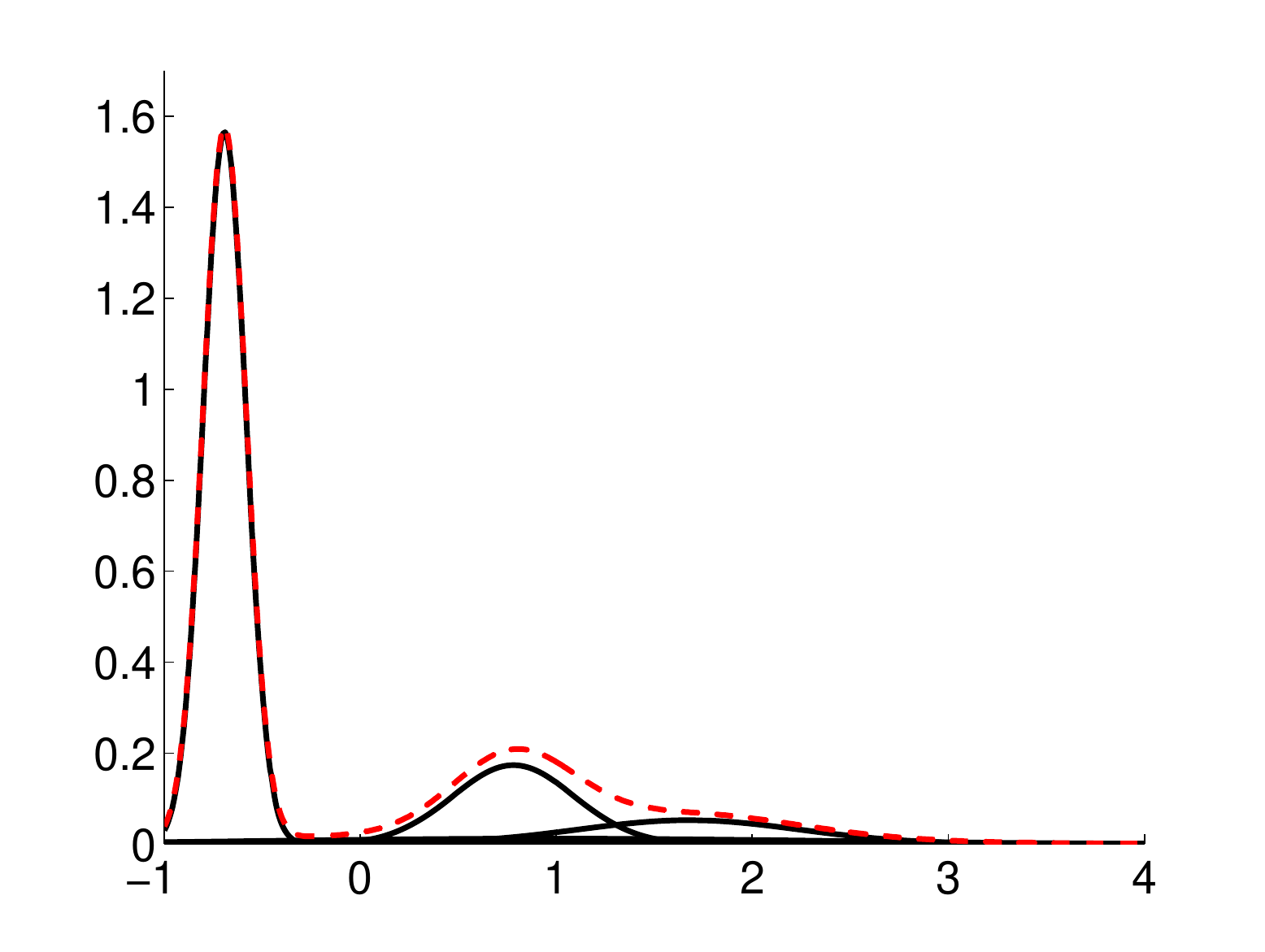} &\hspace{-0.08in}
			\includegraphics[scale=0.15]{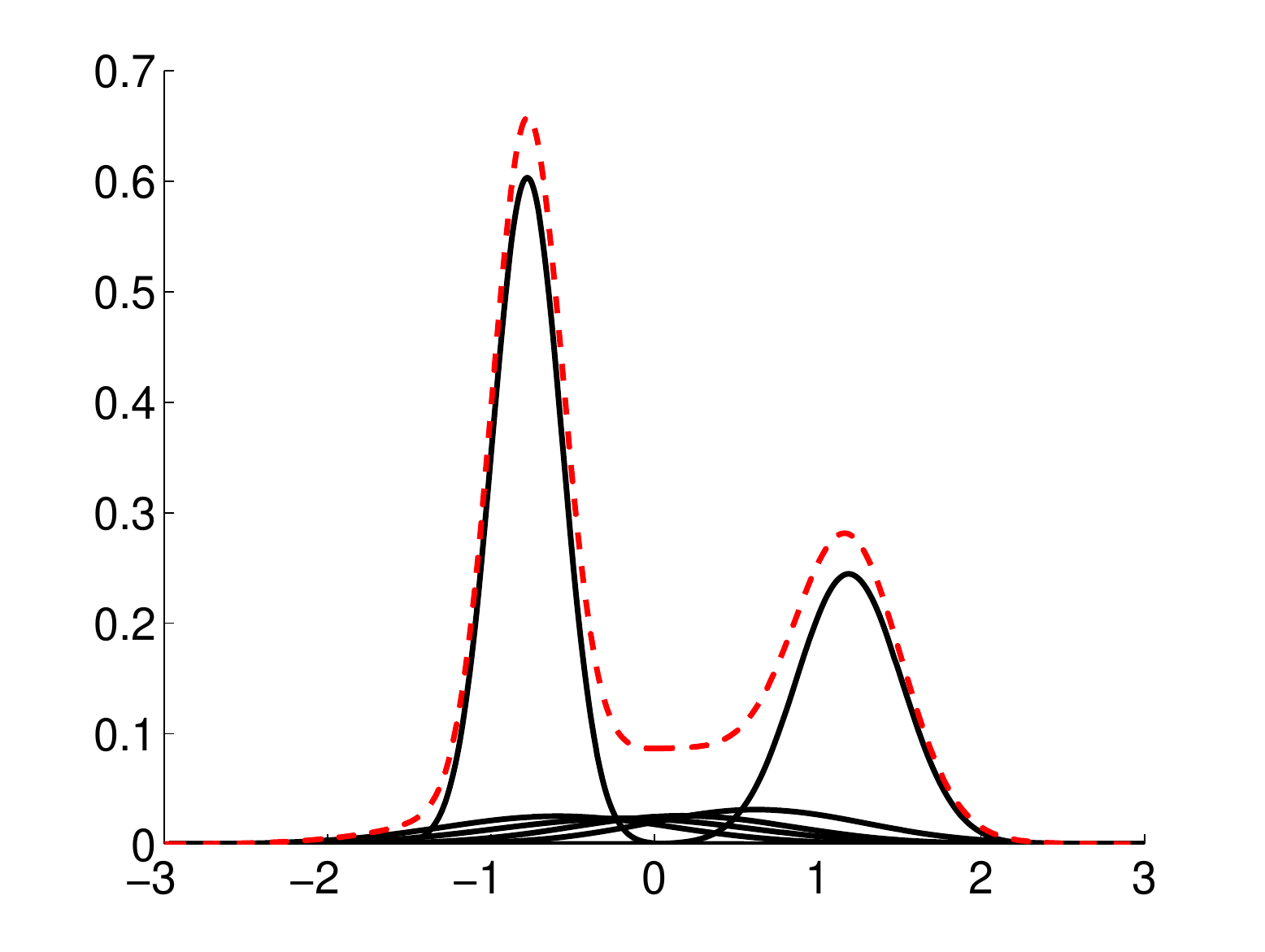}\vspace{-0.02in} \\
			\hspace{-0.15in}\includegraphics[scale=0.15]{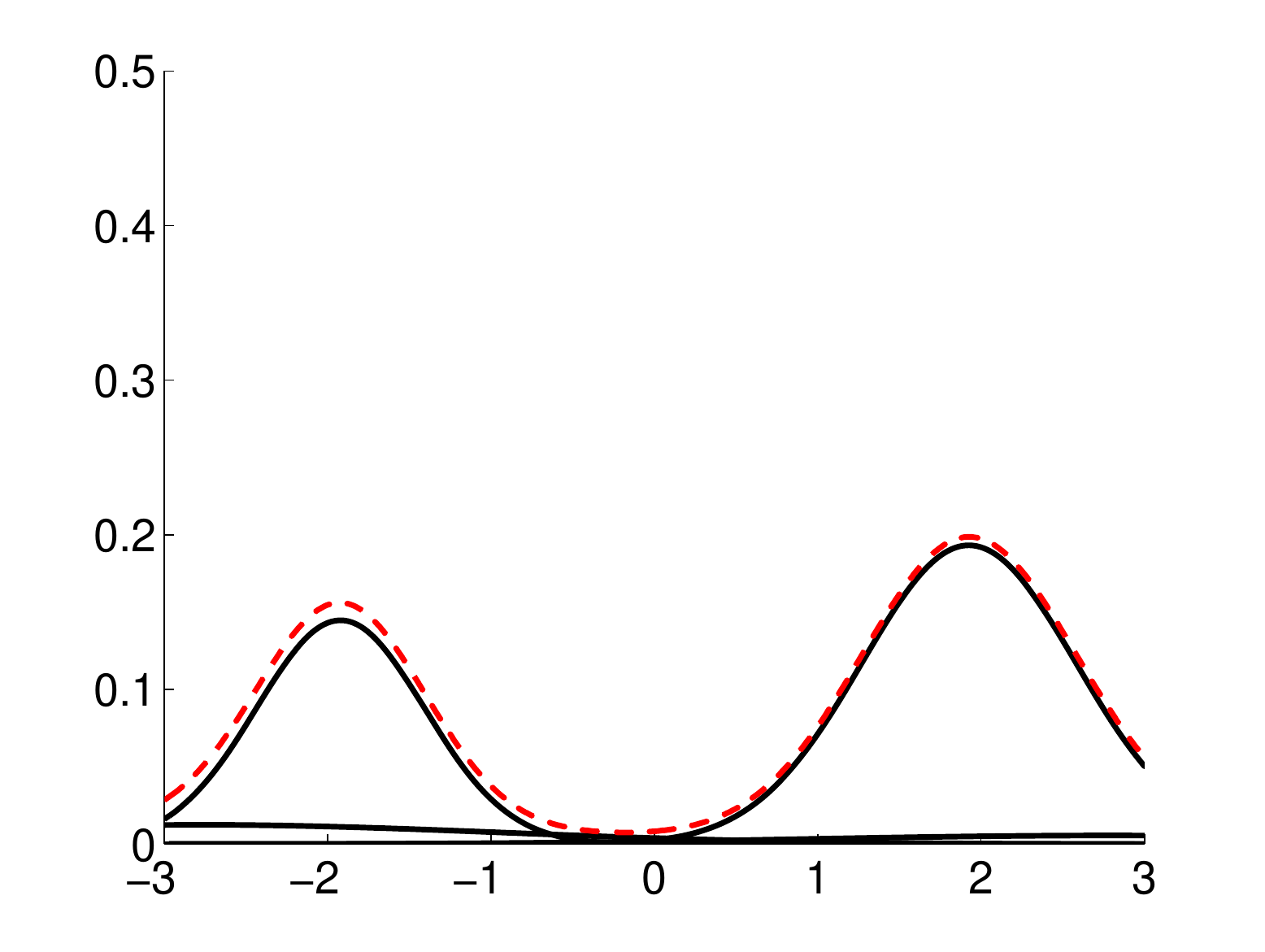} &\hspace{-0.08in}
			\includegraphics[scale=0.15]{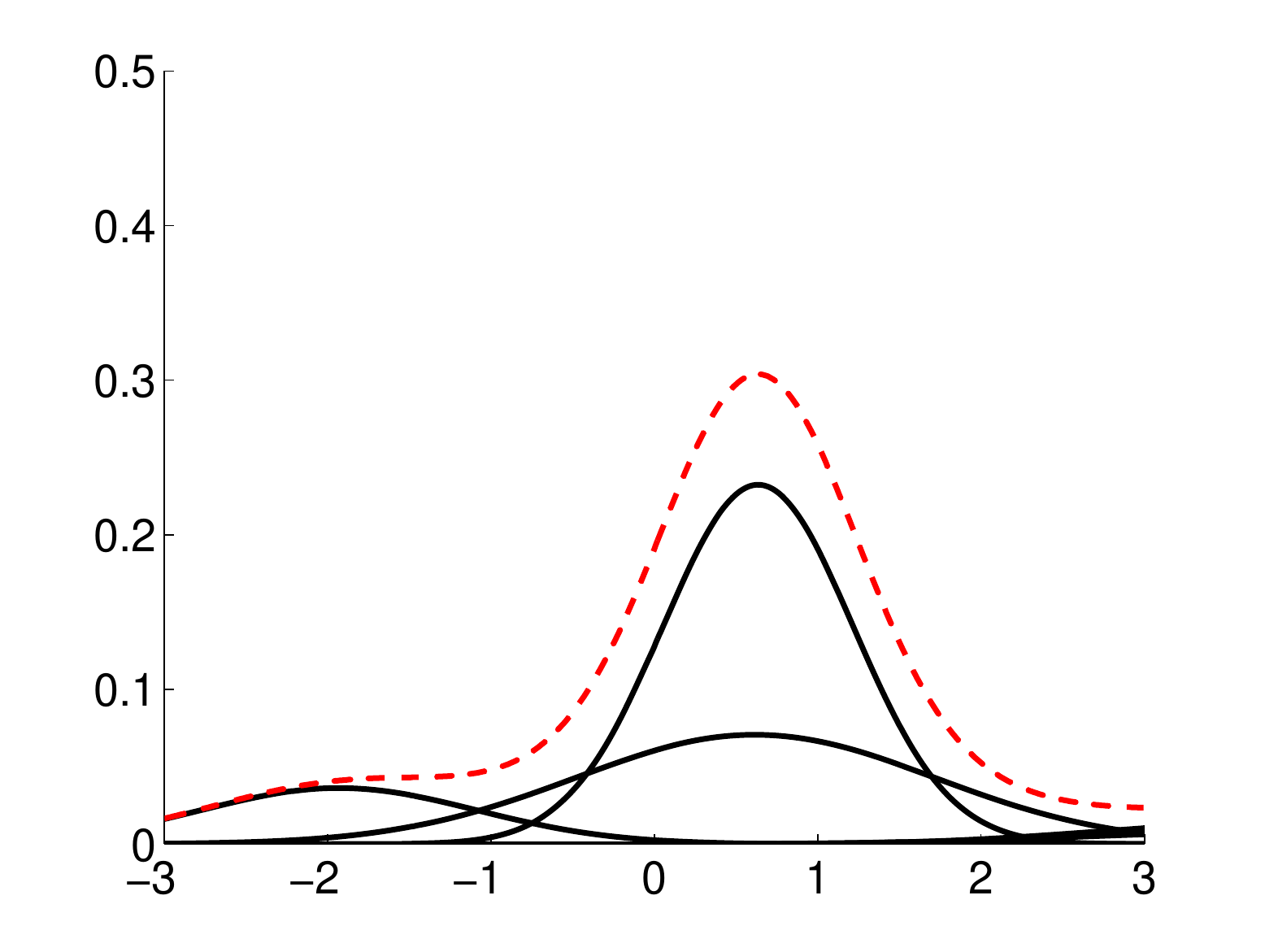} &\hspace{-0.08in}
			\includegraphics[scale=0.15]{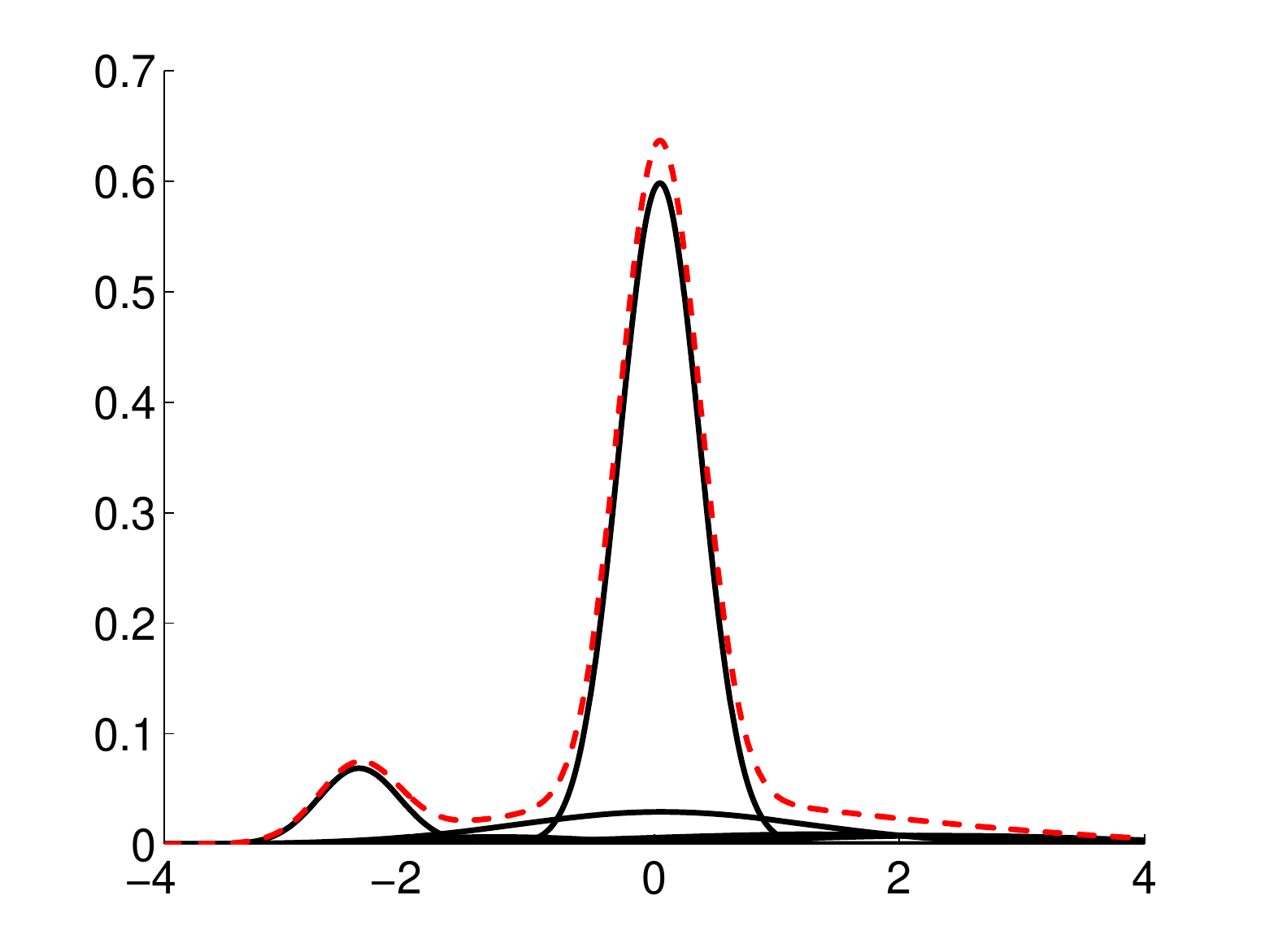} &\hspace{-0.08in}
			\includegraphics[scale=0.15]{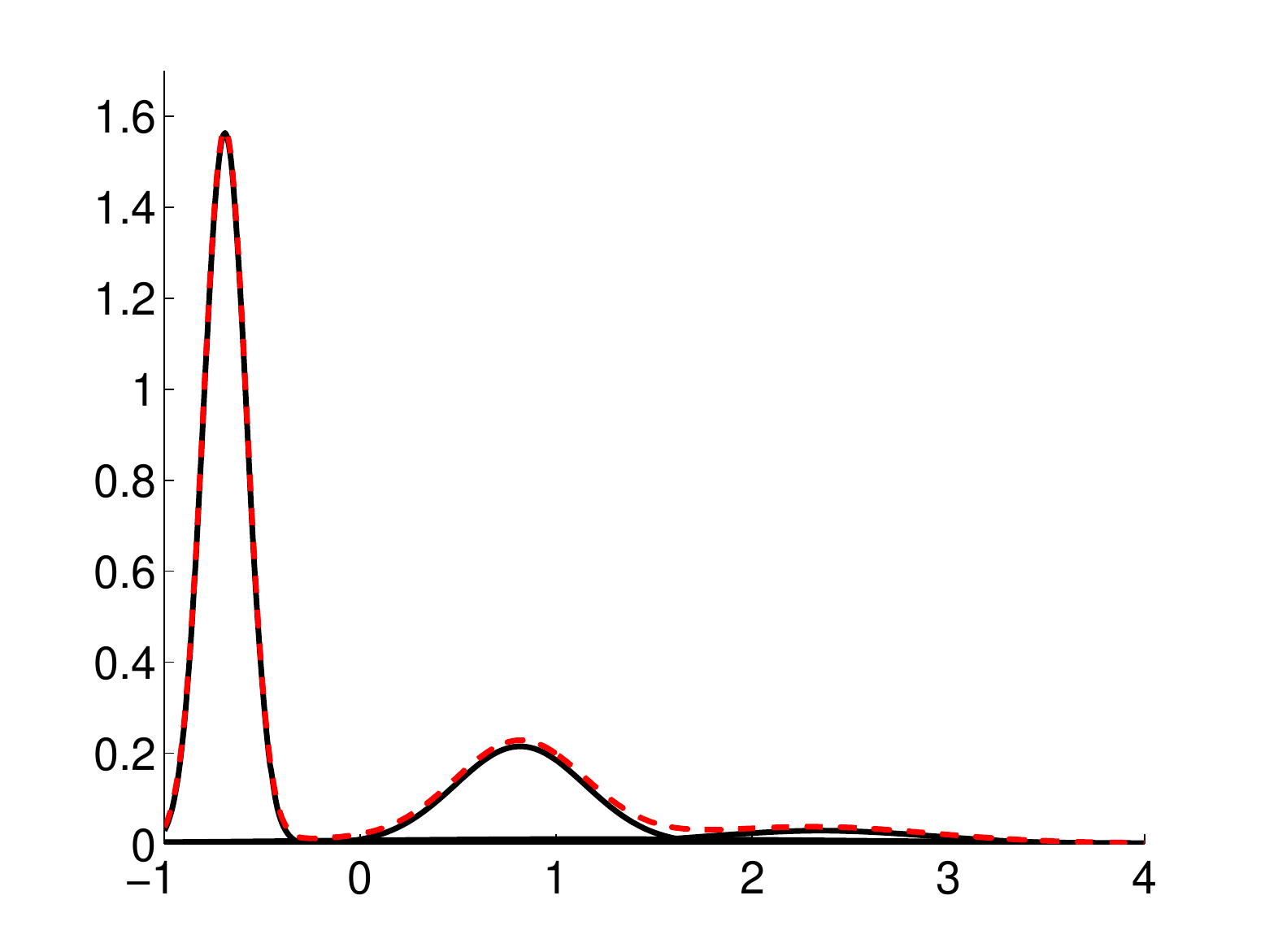} &\hspace{-0.08in}
			\includegraphics[scale=0.15]{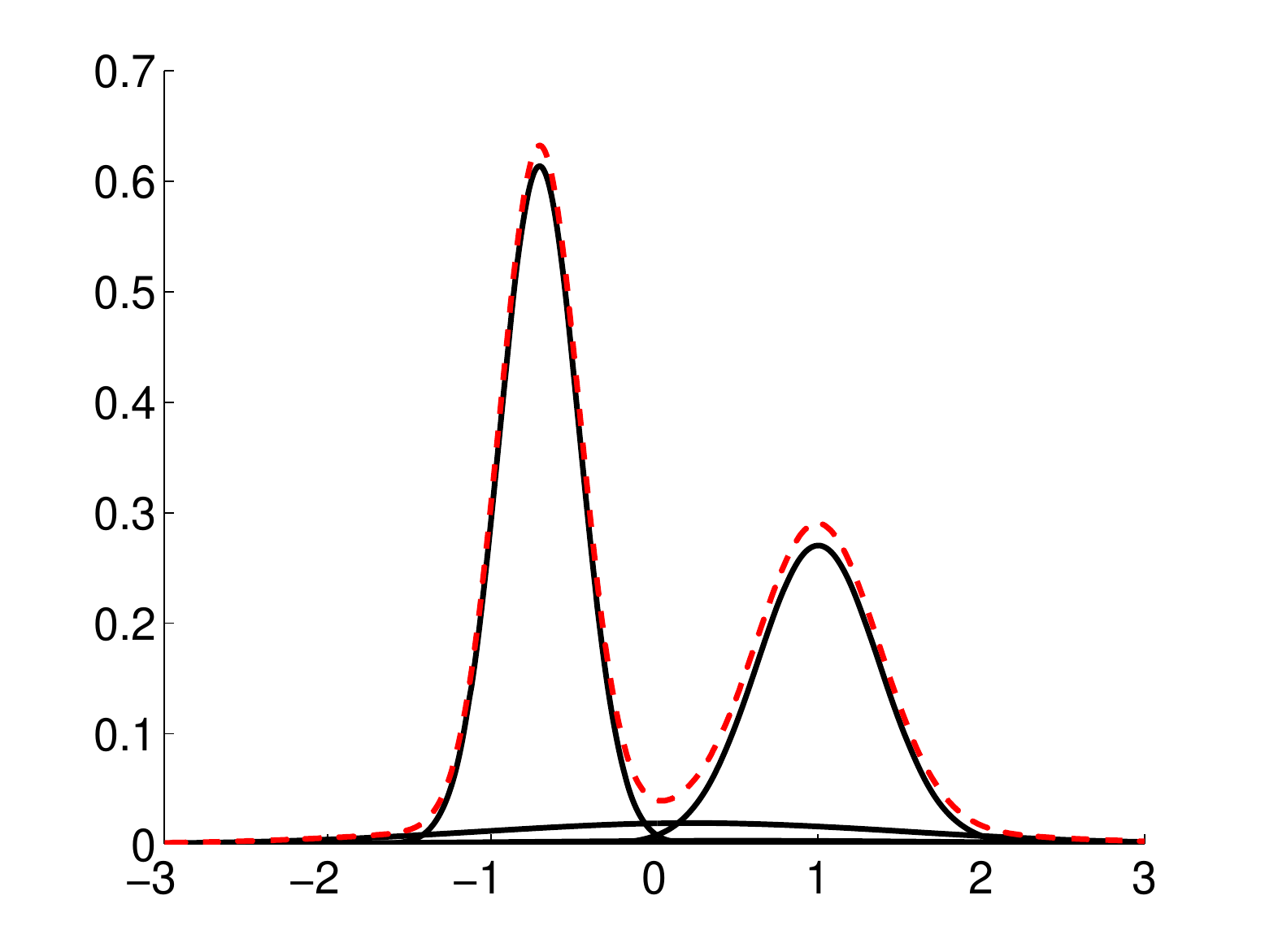}\vspace{-0.02in}\\
			\textbf{Well-Sep} &\hspace{-0.08in} \textbf{Poor-Sep} &\hspace{-0.08in} \textbf{Galaxy} &\hspace{-0.08in} \textbf{Enzyme} &\hspace{-0.08in} \textbf{Acidity}
\end{tabular}
\precap
\caption{\small For each synthetic and real dataset: (top) histogram of data overlaid with actual Gaussian mixture generating the synthetic data, and posterior mean mixture model for (middle) \texttt{IID} and (bottom) \texttt{DPP}. Red dashed lines indicate resulting density estimate.}
\label{fig:mog}
\postcap
\end{center}
\end{figure}

\prepar
\paragraph{Synthetic data}
To assess the role of the prior in a density estimation task, we generated a small sample of 100 observations from a mixture of two Gaussians. 
%We generated 100 observations from a mixture of two Gaussians.
We consider two cases, the first with well-separated components and the second with poorly-separated components. We compare a mixture model with locations sampled i.i.d. (\texttt{IID}) to our DPP repulsive prior (\texttt{DPP}). In both cases, we set an upper bound of six mixture components. In Fig.~\ref{fig:mog}, we see that both \texttt{IID} and \texttt{DPP} provide very similar density estimates. However, \texttt{IID} uses many large-mass components to describe the density.  As a measure of simplicity of the resulting density description, we compute the average entropy of the posterior mixture membership distribution, which is a reasonable metric given the similarity of the overall densities.  Lower entropy indicates a more concise representation in an information-theoretic sense. We also assess the accuracy of the density estimate by computing both (i) Hamming distance error relative to true cluster labels and (ii) held-out log-likelihood on 100 observations.  The results are summarized in Table~\ref{table:synth}.  We see that \texttt{DPP} results in (i) significantly lower entropy, (ii) lower overall clustering error, and (iii) statistically indistinguishable held-out log-likelihood.  These results signify that we have a sparser representation with well-separated (interpretable) clusters while maintaining the accuracy of the density estimate. 
\begin{table}[t]
%	\vspace*{-0.2in}
\caption{\small For \texttt{IID} and \texttt{DPP} on synthetic datasets: mean (stdev) for  mixture membership entropy, cluster assignment error rate and held-out log-likelihood of 100 observations under the posterior mean density estimate.}
\label{table:synth}
\begin{center}
\begin{tabular}{lcccccc}
\multicolumn{1}{l}{DATASET}&  \multicolumn{2}{c}{ENTROPY} & \multicolumn{2}{c}{CLUSTERING ERROR} &\multicolumn{2}{c}{HELDOUT LOG-LIKE.} 
\\ \hline
 & \texttt{IID} & \texttt{DPP} & \texttt{IID} & \texttt{DPP} & \texttt{IID} & \texttt{DPP}\\
Well-separated         & 1.11 (0.3) & 0.88 (0.2) & 0.19 (0.1) & 0.19 (0.1) & -169 (6) & -171(8)\\
Poorly-separated       & 1.46 (0.2) & 0.92 (0.3) & 0.47 (0.1) & 0.39 (0.1) & -211(10) & -207(9)\\
\end{tabular}
\end{center}
\vspace*{-0.2in}
\end{table}

\prepar
\paragraph{Real data}
We also tested our DPP model on three real density estimation tasks considered in \cite{richardson1997bayesian}: 82 measurements of velocity of galaxies diverging from our own (\emph{galaxy}), acidity measurement of 155 lakes in Wisconsin (\emph{acidity}), and the distribution of enzymatic activity in the blood of 245 individuals (\emph{enzyme}). We once again judge the complexity of the density estimates using the posterior mixture membership entropy as a proxy. To assess the accuracy of the density estimates, we performed 5-fold cross validation to estimate the predictive held-out log-likelihood. As with the synthetic data, we find that \texttt{DPP} visually results in better separated clusters (Fig.~\ref{fig:mog}). The \texttt{DPP} entropy measure is also significantly lower for data that are not well separated (\emph{acidity} and \emph{galaxy}) while the differences in predictive log-likelihood estimates are not statistically significant (Table \ref{table:real}).

Finally, we consider a classification task based on the \emph{iris} dataset: 150 observations from three iris species with four length measurements.  For this dataset, there has been significant debate on the optimal number of clusters. While there are three species in the data, it is known that two have very low separation. Based on loss minimization, ~\cite{sugar2003finding,wang2010consistent} concluded that the optimal number of clusters was two. Table~\ref{table:real} compares the classification error using \texttt{DPP} and \texttt{IID} when we assume for evaluation the real data has three or two classes (by collapsing two low-separation classes) , but consider a model with a maximum of six components. While both methods perform similarly for three classes, \texttt{DPP} has significantly lower classification error under the assumption of two classes, since 
%In the 3-cluster setting, 
\texttt{DPP} places large posterior mass on only two mixture components.  
This result hints at the possibility of using the DPP mixture model as a model selection method.  %method to elicit a good estimates of number of clusters. 
\begin{table}[t]
%	\vspace*{-0.2in}
\caption{\small For \texttt{IID} and \texttt{DPP}, mean (stdev) of (\emph{left}) mixture membership entropy and held-out log-likelihood for three density estimation tasks and (\emph{right}) classification error under 2 vs. 2 of true classes for the \emph{iris} data.}
\vspace{-0.05in}
\label{table:real}
\begin{center}
\hspace{-0.22in}	\begin{tabular}{cc}
\begin{tabular}{lcccc}
\multicolumn{1}{l}{DATA}&  \multicolumn{2}{c}{ENTROPY} &\multicolumn{2}{c}{HELDOUT LL.} 
\\ \hline
 & \texttt{IID} & \texttt{DPP} & \texttt{IID} & \texttt{DPP}\\
Galaxy        & 0.89 (0.2) & 0.74 (0.2) & -20(2) & -21(2)\\
Acidity       & 1.32 (0.1) & 0.98 (0.1) & -49 (2) & -48(3)\\
Enzyme        & 1.01 (0.1) & 0.96 (0.1) & -55(2) & -55(3)\\
\end{tabular} & 
\hspace{-0.1in}\begin{tabular}{lcc}
\multicolumn{1}{l}{DATA}&  \multicolumn{2}{c}{CLASS ERROR} 
\\ \hline
 & \texttt{IID} & \texttt{DPP}\\
Iris (3 cls) & 0.43 (0.02) & 0.43 (0.02)\\
Iris (2 cls) & 0.23 (0.03) & 0.15 (0.03)
\end{tabular}
\end{tabular}
\end{center}
\vspace*{-0.2in}
\end{table}

%\begin{table}[t]
%\caption{Mean and standard deviation of cluster assignment error rate under different number of true clusters for the \emph{iris} data}
%\label{table:iris}
%\begin{center}
%\begin{tabular}{lcc}
%\multicolumn{1}{l}{\bf DATA}&  \multicolumn{2}{c}{\bf CLASSIFICATION ERROR} 
%\\ \hline
% & Regular Prior & DPP Prior\\
%Iris (3 clusters) & 0.43 (0.02) & 0.43 (0.02)\\
%Iris (2 clusters) & 0.23 (0.03) & 0.15 (0.03)
%\end{tabular}
%\end{center}
%\end{table}
%
%\subsection{Generating Diverse MoCap Poses for Inverse Reinforcement Learning}
\presec
\section{Generating diverse sample perturbations}
\label{sec:MoCapCR}
\postsec
We consider another possible application of continuous-space sampling. % and present some preliminary results.  
In many applications of inverse reinforcement learning or inverse optimal control, the learner is presented with
control trajectories executed by an expert and tries to estimate a reward function that would approximately reproduce such policies~\cite{abbeel2004alv}.
In order to estimate the reward function, the learner needs to compare the rewards of a large set of trajectories (or all, if possible), which becomes intractable in high-dimensional spaces with complex non-linear dynamics.  A typical approximation is to use a set of perturbed expert trajectories as a comparison set, where a good set of trajectories should cover as large a part of the space as possible.

\begin{figure}
		\begin{minipage}[l]{0.6in}
			\hspace{-.3in}\begin{tabular}{c}
			\includegraphics[scale=0.25,trim=120 60 100 60,clip]{figs/Ori1Pose}\\%\vspace{-0.1in}\\
			\textbf{Original}
		\end{tabular}
		\end{minipage} 
		\begin{minipage}[c]{3.5in}
			\begin{tabular}{ccccc}
				\includegraphics[scale=0.2,trim=60 60 60 60,clip]{figs/RFF1Pose2} & \hspace{-0.5in}
				\includegraphics[scale=0.2,trim=60 60 60 60,clip]{figs/RFF1Pose5} & \hspace{-0.5in}
				\includegraphics[scale=0.2,trim=60 60 60 60,clip]{figs/RFF1Pose8} & \hspace{-0.5in}
				\includegraphics[scale=0.2,trim=60 60 60 60,clip]{figs/RFF1Pose9} & \hspace{-0.5in}
				\includegraphics[scale=0.2,trim=60 60 60 60,clip]{figs/RFF1Pose10} \vspace{-0.1in}\\
				%& & \textbf{RFF DPP Samples} & &\vspace{-0.1in}\\
				\includegraphics[scale=0.2,trim=60 60 60 60,clip]{figs/Nys1Pose3} & \hspace{-0.5in}
				\includegraphics[scale=0.2,trim=60 60 60 60,clip]{figs/Nys1Pose5} & \hspace{-0.5in}
				\includegraphics[scale=0.2,trim=60 60 60 60,clip]{figs/Nys1Pose6} & \hspace{-0.5in}
				\includegraphics[scale=0.2,trim=60 60 60 60,clip]{figs/Nys1Pose9} & \hspace{-0.5in}
				\includegraphics[scale=0.2,trim=60 60 60 60,clip]{figs/Nys1Pose10} \vspace{-0.1in}\\
				& & \textbf{DPP Samples} & &
			\end{tabular}
		\end{minipage}
		\begin{minipage}[r]{1in}
			\includegraphics[scale=0.2]{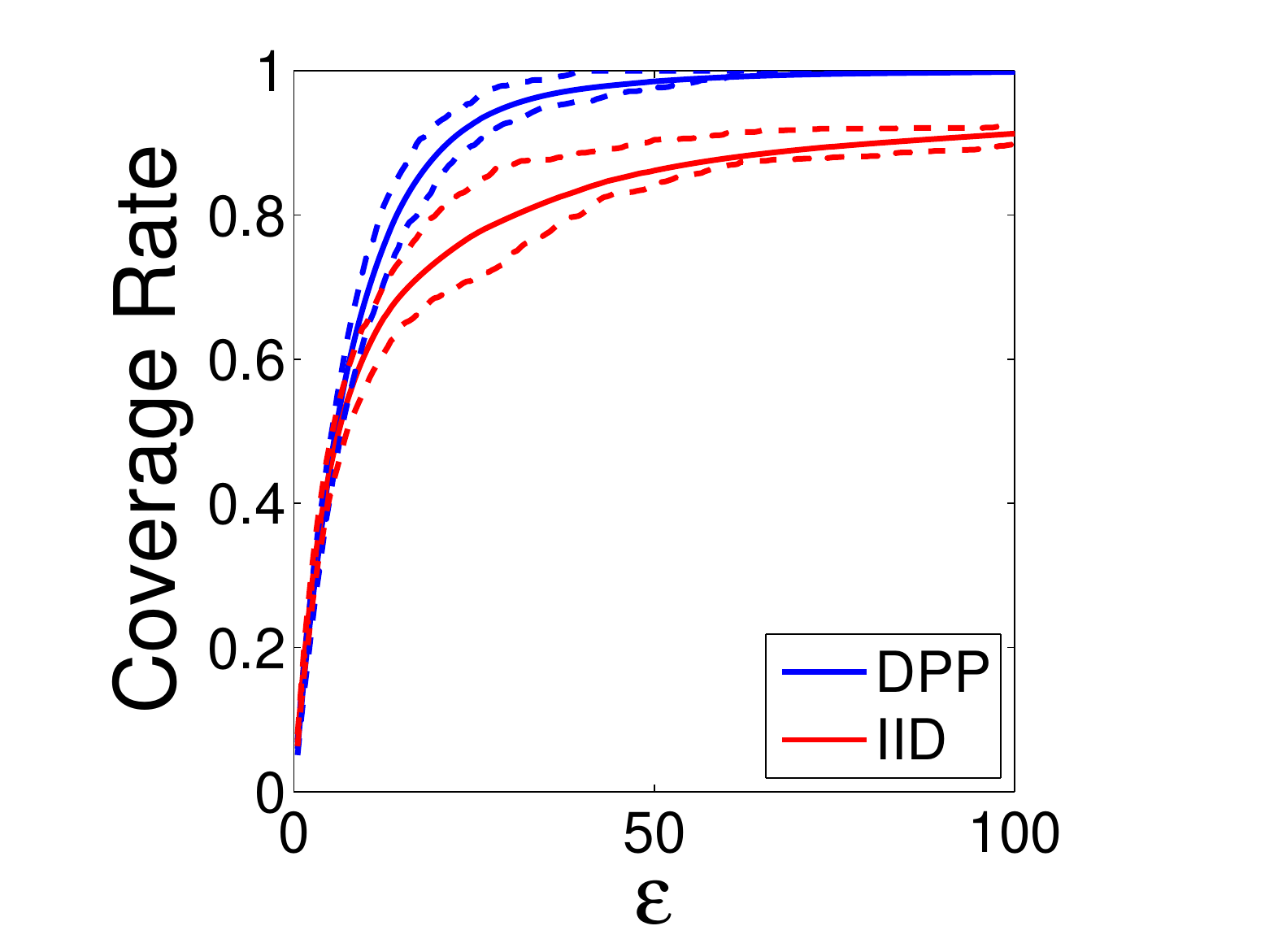}
		\end{minipage}
		\precap
\caption{\small \emph{Left:} Diverse set of human poses relative to an original pose by sampling from an RFF (top) and Nystr{\"om} (bottom) approximations with kernel based on MoCap of the activity  \emph{dance}. \emph{Right:} Fraction of data having a DPP/i.i.d. sample within an $\epsilon$ neighborhood.}
\label{fig:MoCapSynth}
\vspace{-0.1in}
\end{figure}

We propose using DPPs to sample a large-coverage set of trajectories, in particular focusing on a human motion application where we assume a set of motion capture (MoCap) training data taken from the CMU database~\cite{CMUmocap}. Here, our dimension $d$ is 62, corresponding to a set of joint angle measurements.  For a given activity, such as \emph{dancing}, we aim to select a reference pose and synthesize a set of diverse, perturbed poses.  To achieve this, we build a kernel with Gaussian quality and similarity using covariances estimated from the training data associated with the activity.  The Gaussian quality is centered about the selected reference pose and we synthesize new poses by sampling from our continuous DPP using the low-rank approximation scheme.  In Fig.~\ref{fig:MoCapSynth}, we show an example of such DPP-synthesized poses.  For the activity \emph{dance}, to quantitatively assess our performance in covering the activity space, we compute a \emph{coverage rate} metric based on a random sample of 50 poses from a DPP.  For each training MoCap frame, we compute whether the frame has a neighbor in the DPP sample within an $\epsilon$ neighborhood.  We compare our coverage to that of i.i.d. sampling from a multivariate Gaussian chosen to have variance matching our DPP sample.  %Despite favoring the i.i.d. case by inflating the variance to match the diverse DPP sample, the DPP poses still provide better coverage.
Despite favoring the i.i.d. case by inflating the variance to match the diverse DPP sample, the DPP poses still provide better average coverage over 100 runs. See Fig.~\ref{fig:MoCapSynth} (right) for an assessment of the coverage metric.  A visualization of the samples is in the supplement. Note that the i.i.d. case requires on average $\epsilon=253$ to cover all data whereas the DPP only requires $\epsilon=82$.  By $\epsilon=40$, we cover over 90\% of the data on average.  Capturing the rare poses is extremely challenging with i.i.d. sampling, but the diversity encouraged by the DPP overcomes this issue.

%\begin{figure}
%\begin{minipage}[l]{4.75in}
%	\begin{tabular}{cccc}
%\hspace{-0.3in}\includegraphics[scale=0.23]{figs/MoCapPCA1mod}&\hspace{-0.4in}
%\includegraphics[scale=0.23]{figs/MoCapPCA2mod}&\hspace{-0.4in}
%\includegraphics[scale=0.23]{figs/MoCapPCA3mod}&\hspace{-0.4in}
%\includegraphics[scale=0.23]{figs/CovRate100}\vspace{-0.05in}\\
%{\small (a)} & {\small (b)} & {\small (c)} & {\small (d)} 
%\end{tabular}
%\end{minipage}
%\begin{minipage}[r]{1.25in}
%	\begin{tabular}{l|c}
%		& $\epsilon$ required\\
%		&  for full coverage\\
%		\hline
%	DPP & 82\\
%	IID & 253
%\end{tabular}
%\end{minipage}
%\precap
%\caption{\small (a)-(c) DPP (blue) and i.i.d. multivariate Gaussian (red) samples projected onto the top 4 principal components of the \emph{dance} data. (d) Fraction of data having a DPP/i.i.d. sample within an $\epsilon$ neighborhood, with $\epsilon$ for 100\% listed in the table (right). The dashed line indicates the 25th and 75th quantile over 100 runs.}% \end{minipage}each data point to have a sample neighbor.}
%\label{fig:MoCapPCA}
%\vspace{-0.05in}
%\end{figure}

%% file: sections/conclusionCR.tex
\presec
\section{Conclusion}
\label{sec:conclusionCR}
\postsec
Motivated by the recent successes of DPP-based subset modeling in finite-set applications and the growing interest in repulsive processes on continuous spaces, we considered methods by which continuous-DPP sampling can be straightforwardly and efficiently approximated for a wide range of kernels.  Our low-rank approach harnessed approximations provided by Nystr{\"om} and random Fourier feature methods and then utilized a continuous dual DPP representation.  The resulting approximate sampler garners the same efficiencies that led to the success of the DPP in the discrete case.  One can use this method as a proposal distribution and correct for the approximations via Metropolis-Hastings, for example. For $k$-DPPs, we devised an exact Gibbs sampler that utilized the Schur complement representation. %We also show how these two methods can be combined by using the low-rank approximations to generate initial distributions for the Gibbs sampling scheme.%In particular, our approach harnessed (i) low-rank kernel approximations provided by Nystr{\"om} and random Fourier feature (RFF) methods and (ii) a continuous dual DPP representation.  %We also showed that one can combine the Nystr{\"om} and RFF approximations to provide performance as good as or better than each individually.  
Finally, we demonstrated that continuous-DPP sampling is useful both for repulsive mixture modeling (which utilizes the Gibbs sampling scheme) and in synthesizing diverse human poses (which we demonstrated with the low-rank approximation method).  %Future work includes exploring alternative low-rank approximations.
As we saw in the MoCap example, we can handle high-dimensional spaces $d$, with our computations scaling just linearly with $d$. 
We believe this work opens up opportunities to use DPPs as parts of many models.
%The successes here open up opportunities to use DPPs as parts of many models and can be added to the Bayesian arsenal.

\textbf{Acknowledgements:} RHA and EBF were supported in part by AFOSR Grant FA9550-12-1-0453 and DARPA Grant FA9550-12-1-0406 negotiated by AFOSR. BT was partially supported by NSF CAREER Grant 1054215 and by STARnet, a Semiconductor Research Corporation program sponsored by MARCO and DARPA.

%% file: notation.tex
%%%%%%%%%%%%%%%%%%%%%%%%%%%%%%%%%%%%%%%%%%%%%%%%%%%%%%%%%%%%%%%%%%
% \condcommand{numargs}{\name}{def}

\newcommand{\condcommand}[3]{
  \ifdefined #2
    \PackageWarning{condcommand}{#2 is being redefined}
    \renewcommand{#2}[#1]{#3}
  \else
    \newcommand{#2}[#1]{#3}
  \fi
}

%%%%%%%%%%%%%%%%%%%%%%%%%%%%%%%%%%%%%%%%%%%%%%%%%%%%%%%%%%%%%%%%%%

\condcommand{0}{\Lapp}{\tilde{L}}
\condcommand{0}{\Lad}{\hat{L}}
\condcommand{0}{\Lerr}{E}
\condcommand{0}{\l}{\lambda}
\condcommand{0}{\lapp}{\tilde{\lambda}}
\condcommand{0}{\lerr}{\xi}
\condcommand{0}{\lmax}{\hat{\lambda}}
\condcommand{0}{\Papp}{\tilde{\mathcal{P}}}
%% utils
\condcommand{1}{\todo}{\textcolor{red}{TODO: #1}}
\condcommand{1}{\out}{}
\mathchardef\mhyphen="2D

%% mathbb

\condcommand{0}{\E}{\mathop{\mathbb{E}}}
\condcommand{0}{\I}{\mathbb{I}}
\condcommand{0}{\ind}{\boldsymbol{1}}
\condcommand{0}{\reals}{\mathbb{R}}
\condcommand{0}{\ints}{\mathbb{Z}}

%% mathcal

\condcommand{0}{\A}{\mathcal{A}}
\condcommand{0}{\B}{\mathcal{B}}
\condcommand{0}{\C}{\mathcal{C}}
\condcommand{0}{\K}{\mathcal{K}}
\condcommand{0}{\L}{\mathcal{L}}
\condcommand{0}{\N}{\mathcal{N}}
\condcommand{0}{\P}{\mathcal{P}}
\condcommand{0}{\T}{\mathcal{T}}
\condcommand{0}{\W}{\mathcal{W}}
\condcommand{0}{\X}{\mathcal{X}}
\condcommand{0}{\Y}{\mathcal{Y}}

%% bold

\condcommand{0}{\bpsi}{\boldsymbol{\psi}}
\condcommand{0}{\bzero}{\boldsymbol{0}}
\condcommand{0}{\bone}{\boldsymbol{1}}
\condcommand{0}{\bepsilon}{\boldsymbol{\epsilon}}
\condcommand{0}{\bmu}{\boldsymbol{\mu}}
\condcommand{0}{\bpi}{\boldsymbol{\pi}}
\condcommand{0}{\btau}{\boldsymbol{\tau}}
\condcommand{0}{\btheta}{\boldsymbol{\theta}}
\condcommand{0}{\a}{\boldsymbol{a}}
\condcommand{0}{\b}{\boldsymbol{b}}
\condcommand{0}{\e}{\boldsymbol{e}}
\condcommand{0}{\f}{\boldsymbol{f}}
\condcommand{0}{\g}{\boldsymbol{g}}
\condcommand{0}{\r}{\boldsymbol{r}}
\condcommand{0}{\v}{\boldsymbol{v}}
\condcommand{0}{\w}{\boldsymbol{w}}
\condcommand{0}{\p}{\boldsymbol{p}}
\condcommand{0}{\x}{\boldsymbol{x}}
\condcommand{0}{\y}{\boldsymbol{y}}
\condcommand{0}{\m}{\boldsymbol{m}}
\condcommand{0}{\z}{\boldsymbol{z}}
\condcommand{0}{\bY}{\boldsymbol{Y}}
\condcommand{0}{\bA}{\boldsymbol{A}}
\condcommand{0}{\bJ}{\boldsymbol{J}}
\condcommand{0}{\bD}{\boldsymbol{D}}

%% aliases

\condcommand{0}{\cv}{\boldsymbol{\hat v}}
\condcommand{0}{\cV}{\hat V}
\condcommand{0}{\M}{M}
\condcommand{0}{\word}{w}
\condcommand{0}{\allwords}{W}

\condcommand{0}{\cW}{\overline{W}}

%% special

\condcommand{0}{\argmax}{\mathop{\arg\max}}
\condcommand{0}{\argmin}{\mathop{\arg\min}}
\condcommand{0}{\edge}{\mathrm{edge}}
\condcommand{0}{\local}{\mathrm{LOCAL}}
\condcommand{0}{\map}{\mathrm{MAP}}
\condcommand{0}{\mbr}{\mathrm{MBR}}
\condcommand{0}{\marg}{\mathrm{MARG}}
\condcommand{0}{\node}{\mathrm{node}}
\condcommand{0}{\proj}{\mathrm{Proj}}
\condcommand{0}{\st}{\mathrm{s.t.}}
\condcommand{0}{\trans}{\top}
\condcommand{0}{\dist}{\mathrm{dist}}
\condcommand{0}{\rank}{\mathrm{rank}}
\condcommand{0}{\diag}{\mathrm{diag}}
\condcommand{0}{\tr}{\mathrm{tr}}
\condcommand{0}{\length}{\mathrm{length}}
\condcommand{0}{\factor}{\alpha}
\condcommand{0}{\ftr}{\phi}
\condcommand{0}{\var}{\mathrm{Var}}
\condcommand{0}{\vol}{\mathrm{Vol}}
\condcommand{0}{\inc}{\mathrm{in}}
\condcommand{0}{\exc}{\mathrm{out}}
\condcommand{0}{\distinct}{\mathrm{distinct}}
\condcommand{0}{\mrf}{\mathrm{MRF}}
\condcommand{0}{\dpp}{\mathrm{DPP}}
\condcommand{0}{\mmr}{\mathrm{MMR}}
\condcommand{0}{\words}{\mathrm{words}}
\condcommand{0}{\tf}{\mathrm{tf}}
\condcommand{0}{\idf}{\mathrm{idf}}
\condcommand{0}{\tfidf}{\mathrm{tfidf}}
\condcommand{0}{\dsim}{\overset{\scriptstyle D}{\sim}}
\condcommand{0}{\comp}{\mathrm{comp}}
\condcommand{0}{\size}{\mathrm{size}}
\condcommand{0}{\cossim}{\mathrm{cos\mhyphen sim}}
\condcommand{0}{\sgn}{\mathrm{sgn}}
\condcommand{0}{\disc}{\mathrm{disc}}
\condcommand{0}{\rouge}{\mbox{\scriptsize{ROUGE-1F}}}

%% theorems

\newtheorem{theorem}{Theorem}
\newtheorem{lemma}{Lemma}
\newtheorem{corollary}{Corollary}
\newtheorem{proposition}{Proposition}
\newtheorem{definition}{Definition}
\newtheorem{conjecture}{Conjecture}

%% references

\condcommand{1}{\chaplabel}{\label{chap:#1}}
\condcommand{1}{\chapref}{Chapter~\ref{chap:#1}}
\condcommand{2}{\chapsref}{Chapters~\ref{chap:#1} and~\ref{chap:#2}}

\condcommand{1}{\seclabel}{\label{sec:#1}}
\condcommand{1}{\secref}{Section~\ref{sec:#1}}

\condcommand{1}{\figlabel}{\label{fig:#1}}
\condcommand{1}{\figref}{Figure~\ref{fig:#1}}
\condcommand{2}{\figsref}{Figures~\ref{fig:#1} and~\ref{fig:#2}}

\condcommand{1}{\tablabel}{\label{tab:#1}}
\condcommand{1}{\tabref}{Table~\ref{tab:#1}}

\condcommand{1}{\eqlabel}{\label{eq:#1}}
\condcommand{1}{\eqref}{Equation~(\ref{eq:#1})}
\condcommand{2}{\eqsref}{Equations~(\ref{eq:#1}) and~(\ref{eq:#2})}

\condcommand{1}{\proplabel}{\label{prop:#1}}
\condcommand{1}{\propref}{Proposition~\ref{prop:#1}}

\condcommand{1}{\conjlabel}{\label{conj:#1}}
\condcommand{1}{\conjref}{Conjecture~\ref{conj:#1}}

\condcommand{1}{\deflabel}{\label{def:#1}}
\condcommand{1}{\defref}{Definition~\ref{def:#1}}

\condcommand{1}{\lemlabel}{\label{lem:#1}}
\condcommand{1}{\lemref}{Lemma~\ref{lem:#1}}

\condcommand{1}{\thmlabel}{\label{thm:#1}}
\condcommand{1}{\thmref}{Theorem~\ref{thm:#1}}

\condcommand{1}{\alglabel}{\label{alg:#1}}
\condcommand{1}{\algref}{Algorithm~\ref{alg:#1}}